\documentclass[10pt]{article} % For LaTeX2e
\usepackage[preprint]{tmlr}

\usepackage{amsmath,amsfonts,bm}

\def\eqref#1{equation~\ref{#1}}
\def\1{\bm{1}}

\def\va{{\bm{a}}}

\def\vc{{\bm{c}}}
\def\vd{{\bm{d}}}
\def\ve{{\bm{e}}}
\def\vf{{\bm{f}}}

\def\vo{{\bm{o}}}
\def\vp{{\bm{p}}}
\def\vq{{\bm{q}}}

\def\vs{{\bm{s}}}
\def\vt{{\bm{t}}}
\def\vu{{\bm{u}}}
\def\vv{{\bm{v}}}

\def\vx{{\bm{x}}}

\def\mC{{\bm{C}}}
\def\mD{{\bm{D}}}

\def\mK{{\bm{K}}}

\def\mM{{\bm{M}}}
\def\mN{{\bm{N}}}

\def\mP{{\bm{P}}}

\def\mR{{\bm{R}}}

\def\mT{{\bm{T}}}

\def\mZ{{\bm{Z}}}

\DeclareMathAlphabet{\mathsfit}{\encodingdefault}{\sfdefault}{m}{sl}
\SetMathAlphabet{\mathsfit}{bold}{\encodingdefault}{\sfdefault}{bx}{n}
\newcommand{\tens}[1]{\bm{\mathsfit{#1}}}

\def\tB{{\tens{B}}}

\def\tI{{\tens{I}}}

\def\tM{{\tens{M}}}

\def\tP{{\tens{P}}}

\def\sB{{\mathbb{B}}}
\def\sC{{\mathbb{C}}}
\def\sD{{\mathbb{D}}}
\def\sG{{\mathbb{G}}}

\def\sO{{\mathbb{O}}}
\def\sP{{\mathbb{P}}}

\def\sR{{\mathbb{R}}}
\def\sS{{\mathbb{S}}}
\def\sT{{\mathbb{T}}}

\def\sV{{\mathbb{V}}}

\def\sZ{{\mathbb{Z}}}

\usepackage{amsmath, amssymb, amsfonts, mathtools}
\usepackage{algorithm}
\usepackage{algpseudocode}
\usepackage{booktabs}
\usepackage{graphicx}
\usepackage{subcaption}
\usepackage{multirow}
\usepackage{xcolor}
\usepackage{enumitem}
\usepackage{bbm}
\usepackage{hyperref}
\usepackage{url}
\usepackage{listings}
\usepackage{pifont}
\newcommand{\cmark}{\ding{51}}\newcommand{\xmark}{\ding{55}}

\title{DivAS: Interactive 3D Segmentation by Depth-Weighted Voxel Aggregation}

\author{\name Ayush Pande \email ayushp@cse.iitk.ac.in \\
      \addr Department of Computer Science and Engineering\\
      Indian Institute of Technology Kanpur
      \AND
      \name Mayank Vatsa \email mvatsa@iitj.ac.in \\
      \addr Department of Computer Science and Engineering\\
      Indian Institute of Technology Jodhpur
      }

\def\month{MM}  % Insert correct month for camera-ready version
\def\year{YYYY} % Insert correct year for camera-ready version
\def\openreview{\url{https://openreview.net/forum?id=XXXX}} % Insert correct link to OpenReview for camera-ready version

\begin{document}

\maketitle

% \section{Submission of papers to TMLR}

\begin{abstract}
Interactive 3D segmentation of a reconstructed scene should not require a representation-specific optimization loop. We observe that the recipe for lifting 2D foundation-model masks into 3D, namely prompting a few views, refining the resulting masks with rendered depth, and fusing the multi-view evidence into a voxel grid, is shared across scene representations. What remains representation-specific is only the depth signal returned by the renderer and the occupancy prior that gates fusion. We present \textbf{DivAS} (Depth-interactive Voxel Aggregation Segmentation), an optimization-free, training-free framework that realizes this recipe as a single interaction-and-fusion skeleton with lightweight, representation-specific adapters, instantiated on both Gaussian Splatting (GS) and NeRF backbones.

On standard forward-facing and unbounded benchmarks, the GS instantiation attains segmentation quality competitive with state-of-the-art optimization-based methods, and the best on LLFF, while being the only one to reach this quality within the consumer-hardware memory envelope at standard resolution. Both instantiations run end-to-end around $2\times$ faster than feature-field baselines, with a per-update fusion-kernel cost below $70$ ms. Because segmentation evidence is gathered from a small, bounded set of anchor views, user effort and computation remain independent of the training-set size. The same skeleton applied to a NeRF backbone matches or exceeds the performance of optimization-based NeRF baselines, confirming that the recipe transfers across fundamentally different 3D representations.

\end{abstract}

% ================================================================
%  intro.tex  --  DivAS / DivAS-GS  (TMLR)
%  Drop-in section:  \input{intro}  inside main.tex, immediately
%  after the abstract.
% ================================================================

\section{Introduction}
\label{sec:intro}

The ability to segment and annotate 3D scenes efficiently underpins a growing list of computer-vision and graphics workflows: content creation, asset re-use, scene editing, robotics, and autonomous
perception. Recent years have witnessed remarkable progress in two parallel directions: \emph{neural implicit and explicit representations} for 3D scene modeling and \emph{foundation models} for general-purpose image understanding. Yet the practitioner who wishes to combine the two is forced to choose between several representation-specific pipelines, each with its own training loop,
hyperparameter set, and code base.

Neural Radiance Fields (NeRF)~\citep{mildenhall2020nerf} represent a 3D scene as a continuous volumetric function optimized via differentiable volume rendering. Variants such as Instant-NGP ~\citep{muller2022instantngp} reduce training to minutes, and Mip NeRF~360$^\circ$ ~\citep{barron2022mip} extends the formulation to unbounded outdoor environments. NeRF density fields encode rich geometric cues: volumetric density, integrated depth, and view-dependent appearance that make them attractive substrates for scene understanding. More recently, 2D Gaussian Splatting (2DGS)~\citep{huang20242d} and 3D Gaussian Splatting (3DGS)~\citep{kerbl20233d} have shifted the field towards an explicit, real-time scene primitive: a discrete set of oriented surfels carrying opacity and view-dependent color, which can be splatted into the image plane at hundreds of frames per second.
The two representations differ fundamentally in geometry: NeRF is volumetric and continuous, GS is sparse and explicit; NeRF returns depth integrated along a ray, GS returns per-pixel depth statistics (median, expected) over a discrete sample list.

In parallel, the Segment Anything Model (SAM)~\citep{kirillov2023segment}, trained on billions of masks, has established a foundation for zero-shot, prompt-based 2D segmentation. SAM's success raises a practical question: \emph{Can we achieve comparable zero-shot segmentation performance in 3D scenes, regardless of whether the scene is stored as a NeRF or as a 2DGS surfel set?}

\paragraph{The bifurcated state of 3D segmentation.}
The community's response to this question has been representation-specific. NeRF segmentation methods take diverse routes: SA3D~\citep{cen2023segment} runs an iterative inverse-rendering loop with self-prompted view propagation, SANeRF-HQ~\citep{liu2024sanerf} propagates a multi-view user-prompts across the training views via a learned refinement network, and ISRF~\citep{goel2023interactive} distills DINO~\citep{caron2021emerging} features into a feature field for query-based selection. GS segmentation methods are equally varied: SAGA~\citep{cen2025segment} distills SAM features per Gaussian and clusters similar primitives at query time, Gaussian Grouping~\citep{ye2024gaussian} associates 
  a learnable identity feature with each primitive trained against video object segmentation pseudo-labels, and SA3D-GS~\cite{cen2024segment3dradiancefields} replays the SA3D inverse-rendering loop on top of Gaussian primitives. Despite this surface diversity, every one of these methods shares a backbone commitment and an end goal: to turn 2D user prompts into per-primitive (or per-sample) 3D evidence in a single chosen representation. Two further costs follow from this duplication:
(i) every method commits to a single backbone, so a practitioner who trains a 2DGS scene cannot reuse the segmentation tooling developed for NeRF, and vice versa; and (ii) almost all of these methods are \emph{optimization-based}, running a per-scene inverse-rendering loop whose cost scales with the
training-set size $N_\text{train}$ and routinely takes tens of minutes to an hour per segmentation session, depending on the scene complexity.

\paragraph{Insight: one interaction-and-fusion recipe, two representation-specific adapters.}
We argue that a single recipe, depth-weighted voxel voting, can replace all of the backbone-specific machinery above. The recipe factors into five stages: (1) sample a set of anchor views over the camera distribution, (2) prompt SAM on a centroid view of the target, (3) refine the resulting 2D mask using a depth-derived weighting, (4) aggregate the multi-view evidence into a pre-computed voxel-occupancy grid, and (5) display the running 3D mask back to the user for the next click. These five stages, namely the anchor-view sampling strategy, centroid-view exploration, SAM prompting workflow, depth-guided mask refinement, voxel-space evidence aggregation, and interactive fusion loop, are shared by the NeRF and 2DGS instantiations. We therefore describe DivAS as a \emph{shared interaction-and-fusion framework with representation-specific adapters} rather than as a fully representation-independent method. The framework is common at the interaction-and-fusion level, while only two adapter layers are specialized per representation: (i) the extraction of depth and occupancy information from the underlying renderer, and (ii) the projection of the final voxel segmentation back to the native scene representation. In NeRF these adapters operate on volumetric densities, whereas in 2DGS they operate on anisotropic surfels. The shared skeleton stays unchanged even though the underlying rendering primitives differ substantially, which is what lets the same interaction pipeline transfer across fundamentally different 3D scene representations.

\paragraph{The cost story: per-view latency vs.\ per-session optimization.}
  The efficiency advantage of optimization-free segmentation is not fewer user actions, since a depth-weighted voxel-vote method still solicits as many anchor and centroid prompts as the user finds useful, but rather \emph{when} the cost is paid and how interactive the loop feels. Optimization-based baselines such as SAGA and SA3D amortize their cost into a single, long, monolithic per-scene optimization pass that runs after the user has finished annotating and can take tens of minutes to over an hour, depending on scene complexity. DivAS instead pays its cost \emph{incrementally, per annotated view}: SAM inference on the centroid view for 2DGS and rendering time in NeRF plus a single CUDA voxel-fusion kernel call (below $70$\,ms across all benchmark scenes) that re-fuses the entire view buffer accumulated so far. Once a view's annotations are committed, the updated 3D mask is immediately back-projected onto the next anchor view as visual feedback, so the user observes the segmentation evolve \emph{across successive view annotations} rather than \emph{after} a long batch optimization.

\paragraph{Bounded anchor view set.}
A second, structural advantage of optimization-free voxel voting is that the set of anchor views from which segmentation evidence is collected is decoupled from the training-set size. SAGA and SA3D operate by inverse rendering across all training cameras; their cost is therefore $O(N_\text{train})$ and scales with denser captures. DivAS instead works from a fixed-size set of anchor views, obtained for unbounded $360^\circ$ scenes by Fibonacci-sphere distribution and then ranked by geometry~\citep{gonzalez2010measurement}, and for forward-facing LLFF by manual selection. The anchor count is therefore a \emph{bounded view-processing budget} that does not grow with $N_\text{train}$. The bounded anchor budget suffices for a $60$-image LLFF capture and a $300$-image Mip-NeRF~360$^\circ$ capture, and the per-anchor centroid views that the user generates by clicking into each anchor are likewise bounded regardless of how the scene was captured.

\paragraph{Scope and applicability.}
DivAS targets static scenes that have already been reconstructed as a NeRF or 2DGS representation. The contribution is an interactive segmentation framework that operates \emph{on top of} the reconstructed scene rather than a new capture-time or reconstruction-time method. Following the standard evaluation protocols used by prior interactive segmentation work, we report results on the forward-facing LLFF~\citep{mildenhall2019local} benchmark and the inward-facing Mip-NeRF~$360^\circ$~\citep{barron2022mip} benchmark. The segmentation skeleton itself, depth-weighted SAM refinement, voxel-occupancy fusion, and back-projected GUI feedback, makes no assumption beyond a renderer that exposes a depth statistic and an opacity volume, so it is in principle applicable to any reconstructed scene from which those two signals can be read. The Fibonacci-sphere anchor-view scheduler, in contrast, is designed for inward-facing captures where the scene can be reasonably approximated by a viewing sphere, and is the only component of the pipeline that carries a layout assumption. For more general camera layouts (outdoor street captures, drive-throughs, irregular trajectories), the Fibonacci scheduler may not propose the most informative anchors. In those settings the GUI exposes manual anchor selection, which trades a small amount of additional user interaction for unrestricted camera placement. This is a tradeoff between automation and user effort rather than a hard limitation of the segmentation framework.

\paragraph{Consumer-hardware accessibility.}
Peak VRAM, and its dependence on input resolution, is an often-overlooked dimension of segmentation cost. Optimization-based GS methods such as SAGA and Gaussian Grouping keep the trained primitives, gradient buffers, and a per-Gaussian learned feature resident throughout their per-scene optimization, so at the standard $\times 4$ down-sampling their peak VRAM exceeds the consumer-GPU envelope and effectively restricts them to data-center hardware. SAGA fits consumer GPUs only by dropping to the more aggressive $\times 8$ down-sampling, which halves linear resolution and degrades fine geometry. DivAS instead runs at the standard $\times 4$ within the consumer envelope on both NeRF and 2DGS backbones, because it keeps no learnable parameters or feature volumes co-resident with the scene at fusion time, so peak VRAM scales with image resolution and active-voxel count rather than with the trained model.

\paragraph{DivAS: a unified, optimization-free framework.}
Guided by these observations, we present \textbf{DivAS} - a unified framework for interactive 3D segmentation that introduces no additional training or optimization beyond the underlying scene representation. Given a pre-trained NeRF or 2DGS scene, DivAS performs segmentation through direct geometric reasoning and multi-view voxel voting rather than a per-scene inverse-rendering optimization loop. The framework exposes a common interaction-and-fusion skeleton on top of two backbone adapters:
\begin{itemize}[leftmargin=1.5em,itemsep=2pt]
\item The \textbf{NeRF adapter} populates a voxel-occupancy grid by ray-marching the trained density field, producing a grid analogous to the density grid that NeRF itself maintains for empty-space skipping and uses the alpha-composited expected depth as the per-pixel geometric prior for multi-view voxel voting.
\item The \textbf{2DGS adapter} populates the same voxel-occupancy grid by scattering trained surfel opacities directly into voxels. 2DGS has no continuous density field to ray-march and uses the rasterized median depth as the geometric prior, together with lightweight safeguards for the anisotropic floaters and transmissive materials introduced by 2DGS-specific geometry.
\end{itemize}

\noindent The shared skeleton Fibonacci-sphere anchor selection, geometric view ranking, centroid-zoom view generation, depth-weighted SAM mask refinement, and a single CUDA voxel-fusion kernel with
thick-structure and thin-structure paths are identical between the two adapters. The user's interactive experience (click, watch the 3D mask update, refine) is also identical: a single GUI drives both backbones.

\paragraph{Contributions.}
The contributions of this paper are:

\begin{itemize}[leftmargin=1.5em,itemsep=2pt]
\item \textbf{A shared interaction-and-fusion framework with representation-specific adapters.} We isolate the lifting recipe that NeRF and Gaussian-Splatting segmentation methods share into a single, training-free interactive pipeline, where only two lightweight adapters are specialized per backbone, one for renderer depth and occupancy extraction and one for back-projection to the native representation, preserving SAM's zero-shot generalization without per-scene optimization.

\item \textbf{Depth-guided voxel fusion.} Unlike feature-driven affinity or optimization-based inverse rendering, DivAS reweights every SAM mask by the rendered scene depth and aggregates the multi-view evidence into a probabilistic voxel-occupancy grid through a single CUDA kernel that enforces depth, density, and spatial consistency per voxel, lifting masks only where they are geometrically coherent and at a per-call latency below $70$\,ms across all benchmark scenes.

\item \textbf{Thin-structure recovery.} A footprint-based thin-structure path complements the center-ray thick path, aggregating mask and depth evidence over each voxel's full projected extent to recover fine structures such as Trex ribs, fern fronds, and wires that single-pixel center-ray voting drops, which we find is especially critical for the surface-only 2DGS representation.

\item \textbf{Efficient 2DGS voxel-to-surfel mapping.} For the explicit 2DGS backbone, we adapt the tile-sorted rasterization machinery of 2DGS~\citep{huang20242d} into a coverage-based gather that back-projects the fused voxel grid to per-surfel labels. Beyond the borrowed tile scaffold, the mapping contributes a per-voxel view-coverage gate, a depth-free tile sort, a tight one-sigma coverage test, and an anisotropy (needle) filter. Because each surfel is scored by the foreground coverage over its own projected footprint, the mapping thresholds every surfel on its individual area rather than imposing the single strict global threshold used by the baselines, producing sharp, topologically accurate surfel masks at interactive rates without per-primitive feature training.
\end{itemize}

\noindent Together, these choices keep the per-session compute bounded to a small set of anchor and centroid views that scales with object complexity rather than with the training-set size $N_\text{train}$, and hold peak VRAM under $15$\,GB on both backbones, within the consumer-hardware envelope that the data-center footprints of SAGA and Gaussian Grouping exceed (Section~\ref{sec:exp-quant}).

\noindent The remainder of the paper is structured as follows.
Section~\ref{sec:related} reviews prior work in NeRF segmentation, GS segmentation, and 2D foundation-model lifting. Section~\ref{sec:method} presents the shared skeleton and its 2DGS instantiation in detail, with the NeRF instantiation provided in the appendix as evidence that the recipe transfers. Section~\ref{sec:method-gs-fusion} details the shared CUDA voxel-fusion kernel and its thin-structure path. Section~\ref{sec:experiments} reports quantitative and qualitative results on both benchmarks, comparisons against SA3D, SA3D-GS, SAGA, and ablation studies.

\section{Related Work}
\label{sec:related}

\subsection{3D Scene Representations}
Neural Radiance Fields (NeRF)~\citep{mildenhall2020nerf} represent scenes implicitly and synthesize novel views through differentiable volume rendering, with later variants improving efficiency, training speed, and unbounded-scene support~\citep{muller2022instantngp,chen2022tensorf,garbin2021fastnerf,fridovich2022plenoxels,barron2021mip}. We build on Instant-NGP~\citep{muller2022instantngp} for its real-time rendering, which keeps interactive segmentation responsive. 3D Gaussian Splatting (3DGS)~\citep{kerbl20233d} instead represents the scene as an explicit set of anisotropic primitives rasterized through a tile-based differentiable pipeline, trading NeRF's continuous density field for a sparse, surface-attached primitive cloud with real-time high-resolution rendering. 2D Gaussian Splatting (2DGS)~\citep{huang20242d} specializes these primitives to flat surfels aligned with the surface, improving multi-view depth consistency. Because 2DGS surfels live only on the visible surface, the rasterizer returns a single planar median depth per pixel rather than a volume-rendered integral, and the per-pixel splat count $n_\text{samp}$ is a natural proxy for local depth uncertainty. These two signals, planar depth and splat count, are the contact points exploited by the 2DGS instantiation of our framework.

\subsection{2D Segmentation Foundation Models}
2D segmentation has progressed from CNN-based models~\citep{badrinarayanan2017segnet,chen2017deeplab} to transformer-based architectures that leverage global context~\citep{dosovitskiy2020image,xie2021segformer,liu2021swin,strudel2021segmenter}. The Segment Anything Model (SAM)~\citep{kirillov2023segment} established a foundation for zero-shot, prompt-based segmentation, with subsequent efficiency-oriented refinements~\citep{zhang2023faster,xiong2024efficientsam}. Transferring SAM's 2D capabilities into 3D remains an open question, and our work bridges this gap by integrating SAM's zero-shot priors with a representation-agnostic framework that operates over either a NeRF density field or 2DGS surfels.

\subsection{3D Segmentation in NeRFs}
Early works to segment NeRFs, such as Semantic-NeRF~\citep{zhi2021place}, trained a separate semantic field alongside density and color, and NeSF~\citep{vora2021nesf} extends this idea to a generalizable 3D semantic segmentation network learned over the density field from only 2D semantic supervision. Both target fixed semantic categories under supervision rather than interactive, zero-shot selection of arbitrary objects. Other methods like NVOS~\citep{ren2022neural}, SA3D~\citep{cen2023segment}, and ISRF~\citep{goel2023interactive} proposed interactive segmentation but typically require per-scene fine-tuning or optimization. A parallel line of work focuses on feature alignment N3F~\citep{tschernezki2022neural}, DFF~\citep{kobayashi2022decomposing}, and on projecting 2D semantic features into the NeRF space, which can struggle with feature ambiguity and require retraining. LERF~\citep{kerr2023lerf} fuses CLIP~\citep{radford2021learning} features with the 3D scene to enable language-driven queries, but targets coarse, language-based concepts rather than high-precision interactive segmentation of arbitrary objects.

The most closely related NeRF-side baselines to DivAS are SA3D~\citep{cen2023segment} and SANeRF-HQ~\citep{liu2024sanerf}, which also combine SAM with 3D representations. SA3D lifts 2D SAM masks to NeRF and employs a self-prompting mechanism to propagate segmentation across views, but errors in the initial mask or occluded regions can easily propagate. SANeRF-HQ~\citep{liu2024sanerf} formulates the problem as a per-scene optimization, training a refinement network over SAM masks. This sacrifices SAM's zero-shot generalization and incurs significant computation time. In contrast, DivAS is fully optimization-free, preserves SAM's zero-shot capability through a depth-weighted refinement step guided by NeRF geometry, and replaces slow optimization with a real-time CUDA-accelerated voxel-fusion kernel that aggregates geometrically consistent masks.

\subsection{3D Segmentation in Gaussian Splatting}
\textbf{Per-primitive learned features.}
Gaussian Grouping~\citep{ye2024gaussian} attaches a learnable identity feature to every Gaussian primitive and trains it against video-object-segmentation pseudo-labels obtained from a 2D tracker. At query time, primitives are grouped by feature similarity. SAGA~\citep{cen2025segment} pushes this idea further, distilling SAM features into a per-Gaussian semantic vector and clustering similar primitives by feature affinity at query time. Both methods commit to a per-primitive learned feature volume that must remain resident with the Gaussian set during optimization and inference, and both run a per-scene optimization pass whose cost scales with the number of Gaussians and the number of training cameras. LangSplat~\citep{qin2024langsplat} attaches CLIP~\citep{radford2021learning} language features to 3D Gaussians for language-driven queries, sharing the same per-primitive feature-volume cost.

\textbf{Mask-score optimization.}
SA3D-GS~\citep{cen2024segment3dradiancefields} replays the SA3D inverse-rendering loop on top of Gaussian primitives by optimizing an unbounded per-Gaussian mask score against rendered SAM masks. Because the mask score is unconstrained, peripheral high-transparency Gaussians acquire inflated scores and, after thresholding, materialize as floater artifacts in the final 3D mask, degrading IoU even when the optimization itself converges.

\textbf{Common limitations.}
Every method above is optimization-based, commits to the 3DGS backbone, and offers no direct migration path to NeRF except SA3D. These methods must also keep learnable per-primitive features co-resident with the primitive cloud during the optimization pass. This coexistence constraint pushes the VRAM peak well beyond the envelope of consumer GPUs. SAGA explicitly works around this by operating at $\times 8$ image downsampling on consumer hardware, sacrificing linear resolution and thin-structure fidelity. DivAS sidesteps all three limitations on the GS representation. No per-primitive features are trained, no mask score is optimized, and it provides a representation-agnostic solution for both NeRF and GS.

 \subsection{Cross-Representation Interactive 3D Segmentation}
  Two lines of prior work span multiple 3D representations. SA3D~\citep{cen2023segment} has been ported from NeRF to 3DGS as SA3D GS~\citep{cen2024segment3dradiancefields}, but the port reuses the same unbounded mask-score optimization and yields visibly poorer segmentation quality on both benchmarks (Section~\ref{sec:exp-quant}). Closer in spirit is SAMa~\citep{fischer2026sama}, which targets material-aware selection rather than object segmentation. SAMa lifts the 2D predictions of a SAM-like model into a depth-projected 3D point cloud that acts as a representation-agnostic bridge, so a selection transfers to a NeRF, a 3DGS scene, or a mesh through nearest-neighbor lookups, optimization-free and at interactive speed. SAMa shows that a single depth-based selection pipeline can span multiple representations without per-asset optimization, and we regard it as the closest existing point of comparison to our representation-agnostic goal.
                                                                  
  DivAS shares this optimization-free, depth-based, cross-representation philosophy, but differs in the tasks it addresses and the geometric machinery it uses. First, DivAS performs object-level interactive segmentation rather than material-part decomposition, so the selection target and the evaluation protocol differ. Second, where SAMa uses depth only to build a point-cloud bridge and resolves selections by nearest-neighbor transfer, DivAS uses the rendered depth of the trained scene as a per-mask weighting and as a geometric-consistency gate during multi-view voxel voting, aggregating evidence into a probabilistic occupancy grid that is mapped back to NeRF density or to 2DGS surfels through representation-specific adapters. Third, DivAS bounds the user-effort budget to a fixed number of anchor and centroid views that is independent of the training-set size $N_\text{train}$, and it operates at the standard $\times 4$ downsampled image resolution within a 7–15 GB VRAM footprint, whereas the optimization-based GS baselines SAGA and Gaussian Grouping must keep per-Gaussian features resident and drop to $\times 8$ downsampling on consumer hardware, sacrificing thin-structure fidelity. We therefore position DivAS as an object-level interactive segmentation framework that combines depth-weighted voxel aggregation with a shared cross-representation interaction pipeline, enabling optimization-free segmentation on both NeRF and 2DGS scene representations.
% ================================================================
%  DivAS.tex  --  Unified, representation-agnostic Method section (TMLR)
%  Drop-in:  \input{DivAS}  inside main.tex.
%  One interactive 3D-segmentation framework, instantiated over both
%  NeRF (volumetric) and 2DGS (explicit-surfel) scene representations.
%  Organized component-wise: shared formulation first, then expanded
%  representation-specific realizations where they genuinely differ.
%  Requires: booktabs, multirow, graphicx, amsmath, amssymb, bbm,
%  algorithm, algpseudocode, enumitem, subcaption (loaded in preamble).
% ================================================================

\section{DivAS}
\label{sec:method}

% ----------------------------------------------------------------
\begin{figure*}[t]
\centering
\includegraphics[width=\linewidth]{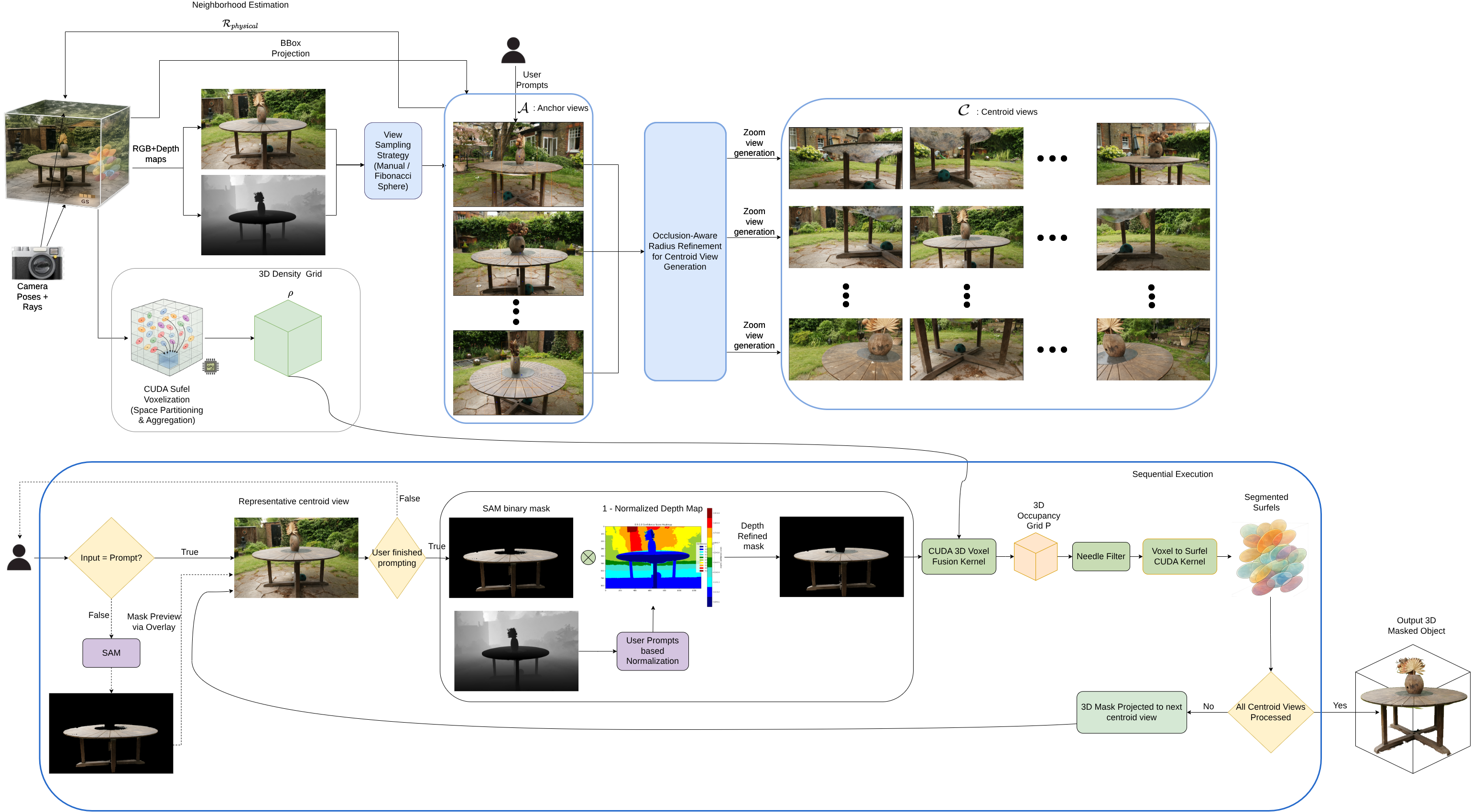}
\caption{\textbf{2DGS instantiation of DivAS.} An occupancy grid is first \emph{scattered} from anisotropic Gaussian surfels and rendered through surfel rasterization to obtain RGB and depth buffers. Unlike NeRFs, explicit 2DGS geometry is highly sensitive to sparse directional coverage and view-dependent occlusions; the Fibonacci anchor-view scheduler therefore adapts each sampling radius to remain within the valid training-camera shell and avoid unsupported viewpoints. An occlusion-aware centroid-view refinement stage further generates zoomed centroid views that suppress floaters and incomplete geometry. During interaction, SAM provides real-time mask previews as the user refines prompts in each centroid view, after which the resulting masks are fused synchronously using a CUDA-based voxel aggregation kernel. The fused occupancy grid is then projected back onto the Gaussian representation using a needle filter, which suppresses anisotropic sliver artifacts and preserves geometric continuity across surfels, followed by a voxel-to-surfel mapping kernel. The corresponding NeRF instantiation and its pipeline details are provided in Appendix~\ref{app:impl-nerf} (Figure~\ref{fig:nerf_arch}).}
\label{fig:gs_arch}
\end{figure*}
% ----------------------------------------------------------------
\subsection{Overview}
\label{sec:method-overview}
We present \textbf{DivAS}, a training-free, human-in-the-loop framework for interactive 3D segmentation built as a \emph{shared interaction-and-fusion framework with representation-specific adapters}. A single algorithmic skeleton operates over both volumetric NeRF using InstaNGP and explicit 2D Gaussian-Splatting (2DGS) representations as backbones, with only the rendering and occupancy adapters specialized per representation. DivAS couples the open-world generalization of a 2D foundation model (SAM~\citep{kirillov2023segment}) with the geometric consistency of a trained 3D scene, and requires no per-scene optimization, no learned per-primitive mask features, and no 2D mask tracker. As illustrated for the 2DGS instantiation in Figure~\ref{fig:gs_arch}, with the corresponding NeRF instantiation in Appendix~\ref{app:impl-nerf} (Figure~\ref{fig:nerf_arch}), a DivAS session proceeds through stages that are conceptually identical across representations. The user marks a few non-overlapping object regions with point prompts on global \emph{anchor views} (Section~\ref{sec:method-anchor}). Each prompt spawns a zoomed-in novel \emph{centroid view} centered on the selected point using the novel view rendering property of these 3D representations, which increases the object's apparent scale and is refined to be free of occluders (Section~\ref{sec:method-centroid}). On each centroid view, SAM masks are reweighted by the rendered depth so that pixels inconsistent with the prompt geometry are suppressed (Section~\ref{sec:method-sam}). The refined masks are aggregated by a single CUDA kernel that votes per voxel under geometric consistency checks, producing a probabilistic \emph{occupancy grid} (Section~\ref{sec:method-fusion}). For the explicit 2DGS backbone, this grid is then mapped back to per-surfel labels (Section~\ref{sec:method-vox2surfel}). Finally, the fused 3D mask is projected onto the next view as feedback, and the loop repeats until coverage is complete (Section~\ref{sec:method-propagation}).

\paragraph{Preliminaries.}
\label{para:prelim}
A NeRF scene representation models the scene as a continuous volumetric radiance field parameterized by a multi-resolution hash-grid encoding~\citep{muller2022instantngp}. Given a 3D position $\vx \in \sR^3$ and viewing direction $\vd$, a lightweight MLP predicts a volume density $\sigma(\vx)$ and view-dependent color $c(\vx,\vd)$. Rendering proceeds through differentiable volume rendering, where densities and colors sampled along each camera ray are composited using alpha accumulation to produce the final RGB image and depth map.

A 2D Gaussian Splatting model~\citep{huang20242d}  represents the scene as a set $\sG = \{g_i\}_{i=1}^{m}$ of oriented 2D surfels. Each surfel carries a center $\boldsymbol{\mu}_i \in \sR^3$, a unit quaternion $q_i$ encoding the surfel orientation, two tangent scales $(s_{i,1}, s_{i,2})$, an opacity
$\alpha_i$, and view-dependent color coefficients. Rendering proceeds by splatting each surfel onto the image plane and alpha-blending in front-to-back order along the ray.

Throughout, we denote the occupancy grid by a density field $\rho$ on a Morton-indexed voxel grid whose edge length we write $\Delta_x$ for the NeRF multi-resolution grid (cascade-dependent, $\Delta_x=2B_\ell/G$ at level $\ell$) and $h_v$ for the single-cascade 2DGS grid (a constant adaptive value per scene), the per-voxel segmentation probability by $P(v)\in[0,1]$, the raw and depth-refined SAM masks for view $s$ by $\mM_s$ and $\widetilde \mM_s$, and the rendered per-pixel depth buffer by the tuple $(\mD_\text{min}, \mD_\text{ref}, \mD_\text{max}, \mD_\text{mw}, \mN_\text{samp})$, where $\mD_\text{min}$ is the depth of the first significant weighted sample, $\mD_\text{max}$ is the depth of the sample when cumulative density reaches $\tau_\text{cw}$ beyond which background primitves could be sampled, $\mD_\text{mw}$ is the max-weighted sample depth, $\mD_\text{ref}$ is the backbone's primary surface estimate ($\mD_\text{exp}$, the expected depth, for NeRF; $\mD_\text{med}$, the median depth, for 2DGS) and $\mN_\text{samp}$ stores the number of primitives traversed up to the surface. For set of views the depth tuple is represented by $(\{\mD^{(s)}_\text{min}\}_{s=1}^S, \{\mD^{(s)}_\text{ref}\}_{s=1}^S, \{\mD^{(s)}_\text{max}\}_{s=1}^S, \{\mD^{(s)}_\text{mw}\}_{s=1}^S, \{\mN^{(s)}_\text{samp}\}_{s=1}^S)$ where symbols have usual meanings.

The proposed framework maintains a unified interaction, propagation, and geometric-consensus formulation across both NeRF and Gaussian-Splatting scene representations. While the high-level pipeline remains shared, several core components, including rendering, occupancy construction, and multi-view geometric fusion, require representation-specific realizations due to the fundamentally different geometric primitives underlying volumetric radiance fields and anisotropic Gaussian splats. Accordingly, the remainder of this section is organized component-wise: each subsection first introduces the shared formulation and then expands the representation-specific implementations where the underlying computational or geometric behavior differs substantially.

% ================================================================
\subsection{Anchor View Selection}
\label{sec:method-anchor}

\paragraph{Shared sampling and ranking.}
DivAS supports both bounded forward-facing scenes and unbounded $360^\circ$ scenes. For forward-facing captures, a small set of manually selected anchor views covers the object with minimal camera motion, since most of the geometry is visible from a narrow cone of viewpoints. For unbounded scenes, uniform angular coverage is required, and we generate a compact, low-discrepancy scene-specific candidate set with \emph{Fibonacci-sphere} sampling~\citep{gonzalez2010measurement}. This yields uniform spherical coverage in $O(N)$, even for small $N$ (we use $N{=}12$).

Candidate views are oriented toward the scene center and then ranked by a geometric \emph{informativeness} score balancing three factors: (i) \emph{diversity} $D_i$, the mean angular separation to the other views. (ii) \emph{cardinal coverage} $C_i$, proximity to the canonical axes $\{\pm\mathbf{x},\pm\mathbf{y},\pm\mathbf{z}\}$ and (iii) \emph{pitch extremity} which favors informative top-down and bottom-up perspectives that improve 3D understanding. The score for a view is a weighted sum:
$
I(i) = 0.4D_i + 0.3 C_i + 0.3 P_i
$
The top-ranked views are retained as anchor viewpoints for subsequent interaction. Since the ranking is purely geometric and independent of object semantics, the GUI provides a toggle to draw anchors from the \emph{full} Fibonacci set, allowing the user to prioritize views that better expose an occluded or off-center target in multi-object or object-centric scenes.

\paragraph{Representation-aware anchor placement.}
Both backbones place the $K$ retained anchors on a sphere centered at $\mathbf{c}_\text{scene}$ along the Fibonacci directions but differ in the \emph{radius} at which each anchor is placed, and this difference is dictated by how each representation behaves away
from the training cameras. For NeRF, a single global radius equal to the median training-camera distance suffices in every direction. The property of NeRF being bounded, normalized in a cube, its volumetric field renders a coherent image from any point on that shell and
degrades gracefully even where training views are sparse, with no hard occlusion penalty. The explicit 2DGS representation is substantially more sensitive to occlusions and incomplete directional coverage than volumetric NeRFs. Because anisotropic surfels are only reliably reconstructed within the shell spanned by the training cameras, using a fixed global sampling radius may place anchor views in unsupported regions, leading to floaters, holes, and severe view-dependent artifacts. We therefore make the anchor radius adaptive for each Fibonacci direction,
\begin{equation}
r_i = \min\bigl(r_i^{\mathrm{knn}}, r_i^{\mathrm{cap}}\bigr),
\label{eq:gs-radius}
\end{equation}
where $r_i^{\mathrm{knn}}$ estimates a locally representative radius from nearby training-camera density, and $ r_i^{\mathrm{cap}} $ constrains the anchor within the valid directional camera shell. This adaptive formulation keeps anchor views both geometrically representative and within reliably reconstructed regions, substantially improving robustness under sparse coverage and view-dependent occlusions. Since 2DGS camera poses are not canonically aligned during preprocessing, we additionally estimate a global scene up-direction before generating the Fibonacci sampling sphere. Further implementation details are deferred to the supplementary material.

\paragraph{Representation-specific rendering.}
The segmentation framework is agnostic to how a view is rendered. Only the rendering adapter differs. For \textbf{NeRF}, RGB and depth are produced by volumetric ray-marching of the trained density field, with the depth buffer read directly from the ray-integration loop. In NeRF, the expected depth $D_\text{exp}$ represents where the surface lies in the density cloud marched by the ray. For \textbf{2DGS}, the depth buffer is produced by surfel rasterization. In 2DGS, median depth $D_\text{med}$ is the primary surface proxy because it is robust to floaters near the camera.

% ================================================================
\subsection{Centroid-View Expansion and Zoom Refinement}
\label{sec:method-centroid}

\paragraph{Why centroid views are necessary.}
SAM's mask decoder produces a low-resolution $(256{\times}256)$ logit map, which is upsampled to the input resolution. When the target occupies a small fraction of a global anchor view or an image in the dataset, a common case for thin or distant structures, this coarse decoder produces ragged boundaries and systematic false negatives in fine regions. To counter this, every point prompt in anchor view spawns a \emph{centroid view}. A zoomed-in novel render in which the back-projected $3$D click point is set as the camera's look-at target and the camera is moved toward it, enlarging the object's apparent scale. Magnifying the object in the image plane directly mitigates the decoder's resolution bottleneck and improves recall on thin structures. The full centroid set across $A$ anchors and $K$ prompts per anchor is:
\begin{equation}
\mathcal{C} = \bigcup_{i=1}^{A}\bigcup_{j=1}^{K}\mathcal{A}_{ij}.
\label{eq:centroid-set}
\end{equation}

\paragraph{2DGS instantiation: occlusion-aware centroid refinement.}
The fixed-fraction zoom does not transfer to 2DGS, where scenes are unbounded and a constant push can drive the camera through the target or into foreground floaters. We instead size each centroid view physically from the local surfel neighborhood around the click, then refine the orbital radius with an occlusion-aware probe that pulls the camera in until the target is no longer dominated by occluders. The neighborhood radius estimate, the lateral-spread sizing, the dynamic occlusion test, and the per-click algorithm are detailed in Appendix~\ref{app:centroid-view}.

\subparagraph{Iterative centroid processing.}
Centroid views are processed one at a time, and after each fusion, the running 3D mask is projected back as an overlay (Section~\ref{sec:method-propagation}), so
the user re-prompts only regions not yet covered. This converts the set $\mathcal{C}$ of Equation~\ref{eq:centroid-set} into a sequence in which each new
view both adds a fresh viewpoint and corrects residual gaps from prior views, the mechanism by which sparse user input expands to complete object coverage.

% ================================================================
\subsection{Depth-Guided Interactive Segmentation}
\label{sec:method-sam}

\paragraph{Shared interaction loop.}
Centroid views improve SAM’s precision by enlarging object scale, yet SAM often produces false positives and boundary artifacts at this scale. Similar observations were made in SANeRF-HQ~\citep{liu2024sanerf}, which reported that segmentation errors in both 2D and 3D are more likely to occur near object boundaries. Motivated by this, we refine SAM masks using the rendered depth map from 3D representations as geometric priors. For each centroid, the user supplies positive (and optionally negative) point prompts. SAM produces a candidate mask $\mM_s$, mapped to a $[0,1]$ confidence by the sigmoid. We then reweight $\mM_s$ by the rendered depth, so that pixels whose depth is inconsistent with the prompted surface are suppressed, producing the depth-refined mask $\widetilde \mM_s$ used for fusion. The running fused 3D mask is overlaid on the current view so the user re-prompts only uncaptured regions in an interactive correction loop rather than a one-shot prediction. Two facets of this loop differ by backbone, and we present them as a single systems-design trade-off driven by rendering speed and scene boundedness, not as two separate pipelines.

\paragraph{Execution scheduling: asynchronous (NeRF) vs.\ synchronous (2DGS).}
The dominant interactive cost varies across backbones, so the optimal SAM scheduling differs accordingly. \textbf{NeRF} rendering is slow, so SAM inference is cheap by comparison and easy to hide: DivAS-NeRF runs SAM \emph{asynchronously} in a background worker, processing the preceding centroid views $\{c_{i-1},c_{i-2}\}$ while the user annotates $c_i$, and transitions from asynchronous to sequential execution only when (i) $c_i$ is the last view of its anchor group, or (ii) the cache holds at least three pending masks. At that point, pending inferences are synchronized and fused together as shown in Figure~\ref{fig:nerf_arch}. \textbf{2DGS} rendering is fast enough that each centroid view is generated in milliseconds compared to $1 \text{-} 2$s taken by NeRF, so a \emph{synchronous} call enables immediately incorporating the resulting mask into the interaction loop.

\paragraph{Mask-channel selection and view acceptance.}
SAM emits three masks per view, ordered by ambiguity. DivAS-GS always retains the single mid-granularity mask at channel index $1$, which was the most consistent across centroid views because each centroid view focuses on a part of an object and inherently gets masked by the mid-granularity mask of SAM. A view is \emph{rejected} (its mask discarded) if it carries no positive prompt, or if SAM's channel-$1$ score is below $0.5$, indicating SAM is not committed to a salient object and any forced mask would be noise. In this case, this centroid view is not considered in the fusion step.

\begin{figure}[t]
\centering
\begin{subfigure}[b]{0.48\linewidth}
\centering
\includegraphics[width=\linewidth]{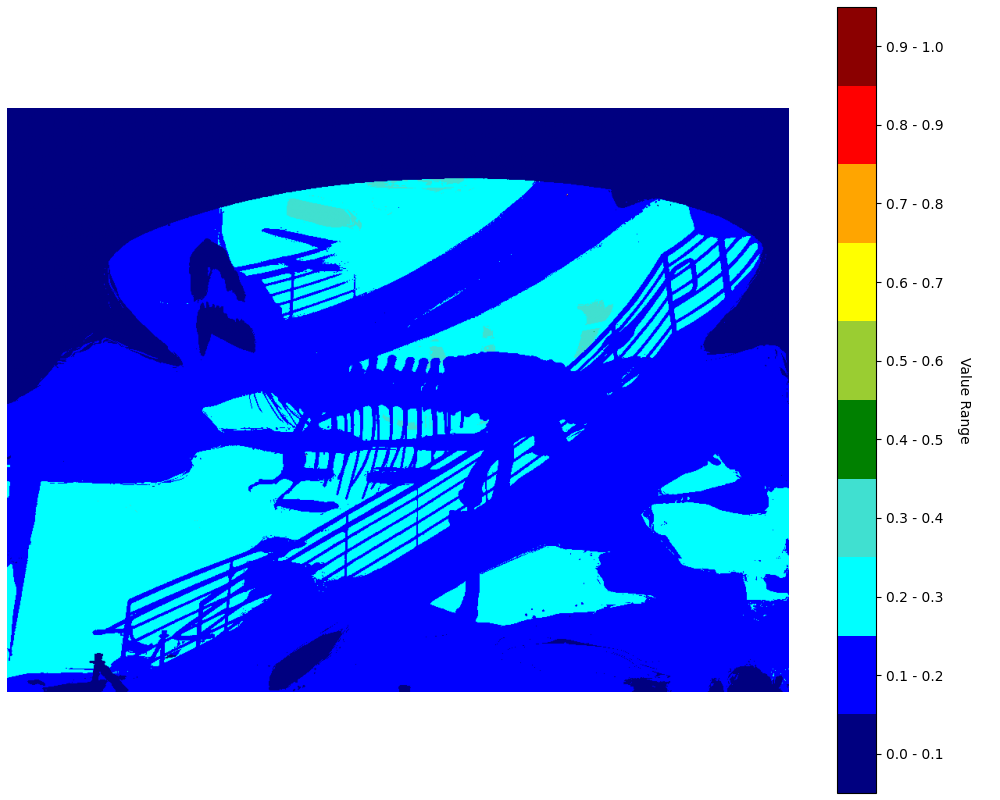}
\caption{NeRF-style min-max normalization on a 2DGS depth raster, far background compresses the range and the cue collapses.}
\label{fig:depthband-without}
\end{subfigure}\hfill
\begin{subfigure}[b]{0.48\linewidth}
\centering
\includegraphics[width=\linewidth]{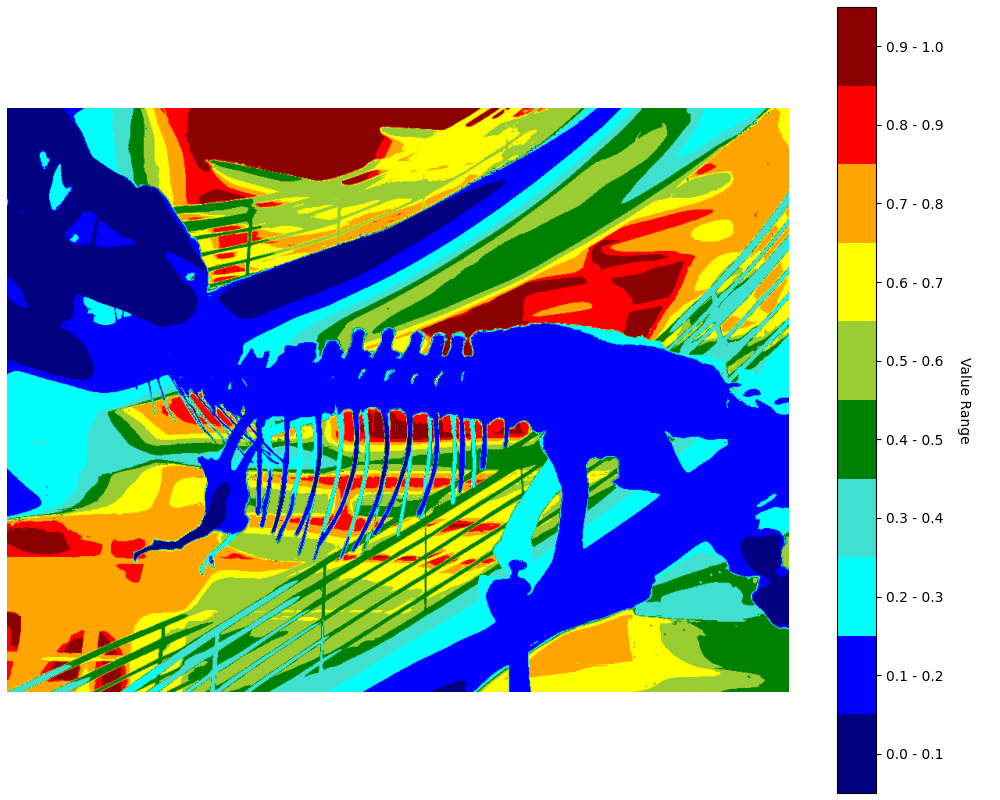}
\caption{Click-defined band $[d_{\min}^{(s)},d_{\max}^{(s)}]$ with $\mathcal{P}_{98}$ far estimate, the full object stays at unit weight.}
\label{fig:depthband-with}
\end{subfigure}
\caption{Depth-weighting for 2DGS (Eq.~\ref{eq:gs-bandpass}). The prompt-defined band replaces NeRF's min-max normalization and is robust to the unbounded depth
range of 2DGS scenes.}
\label{fig:depthband}
\end{figure}

\paragraph{Depth-weighting: bounded normalization (NeRF) vs.\ band-pass (2DGS).}
DivAS-NeRF weights the SAM mask by the inverse of the min-max-normalized expected depth,
\begin{equation}
\widetilde{\mM}_{s} = \mM_{s}\odot\bigl(1-\hat{\mZ}_{s}\bigr),
\qquad
\hat{\mZ}_{s} = \frac{\mD_\text{exp}-\min(\mD_\text{exp})}{\max(\mD_\text{exp})-\min(\mD_\text{exp})},
\label{eq:nerf-depthweight}
\end{equation}
which provides well-defined foreground-background separation precisely because the trained AABB clips each ray to a bounded, artifact-free depth range, suppressing distant pixels and avoiding large depth values, thereby preventing squashing of the normalized range. This rule \emph{cannot} transfer to 2DGS where surfels at sky distance or on far facades leave the median-depth raster uncapped, so a global min-max normalization (or any single-anchor depth falloff) collapses foreground and near-background to nearly identical weights and the depth separation vanishes (Figure~\ref{fig:depthband-without}). Because the object's depth extent is not known a priori, there is no AABB to bracket where the object lives. DivAS-GS turns the user's own prompts into a depth \emph{band}. Reading the median depth at every positive prompt and at the view center (a near-free anchor, since the centroid view looks directly at the target along the orbital axis), we form
\begin{equation}
\sD_s=\bigl\{\mD_\text{med}^{(s)}(p_{x,k})\bigr\}_{k=1}^{\sP_s}\cup\bigl\{\mD_\text{med}^{(s)}(\tfrac{w}{2},\tfrac{h}{2})\bigr\},
\quad
d_{\min}^{(s)}=\min\sD_s,\
d_{\max}^{(s)}=\max\sD_s.
\label{eq:depthband}
\end{equation}
A robust far estimate $d_\text{far}=\mathcal{P}_{98}(\mD_\text{med}^{(s)})$ discards the top $2\%$ tail (sky, far facades, reconstruction holes) so the falloff slope $1/(d_\text{far}-d_{\max}^{(s)})$ stays bounded and dimensionally consistent across scenes of very different scale. Pixels inside the band keep full weight. And, pixels outside fall off linearly toward $d_\text{far}$:
\begin{align}
\boldsymbol{\Delta}_-&=\max(d_{\min}^{(s)}-\mD_\text{med},0),\quad
\boldsymbol{\Delta}_+=\max(\mD_\text{med}-d_{\max}^{(s)},0),\nonumber\\
\mD_\text{band} &= \text{clip}\!\Bigl(\tfrac{\boldsymbol{\Delta}_- + \boldsymbol{\Delta}_+}{\max(d_\text{far}-d_{\max}^{(s)},\,10^{-6})},\,0,\,1\Bigr),
\qquad
\widetilde \mM_s = \mM_s\,(1-\mD_\text{band}).
\label{eq:gs-bandpass}
\end{align}
At most one of $\boldsymbol{\Delta}_-,\boldsymbol{\Delta}_+$ is nonzero per pixel. A pixel at $d_\text{far}$ receives weight $0$ and one at the band edge weight $1$, and the floored denominator prevents division blow-up when the object spans to the far edge (reducing the weight to a pure inside/outside indicator). The band keeps a spatially extended object, e.g., a Trex skeleton spanning meters between a rib click and a skull click at unit weight, while still rejecting background whose median depth lies outside the prompted band (Figure~\ref{fig:depthband-with}). The band-pass thus recovers normalized depth for the unbounded 2DGS case, the role that bounded normalization plays for NeRF.

% ================================================================
\subsection{Representation-Specific 3D Fusion}
\label{sec:method-fusion}

\paragraph{Shared voxel-voting concept.}
The fusion stage aggregates the depth-refined masks of all retained views into the occupancy grid. In both backbones, a single CUDA kernel assigns \emph{one
thread per voxel}. For active voxel $v$ (world center $\vx_v$, edge $h_v$) and view $s$, it projects $\vx_v$ into view $s$, reads the mask confidence
and depth channels at the projected pixel, and admits a vote only if the voxel is both \emph{spatially} and \emph{depth} consistent with the surface the renderer
reports there. Accepted votes are accumulated with a boundary-attenuating depth weight that is maximal when the voxel projects to the center of the surface
segment and decays toward its edges. Voxels missed by this center-aligned check are re-examined by a \emph{thin-structure} path that aggregates evidence over the
voxel's full $2$D footprint. After all $S$ views, the per-voxel probability is the depth-weighted, thresholded consensus over a thick-path view set $\sC_v$
and a thin-path set $\sT_v$:
\begin{equation}                                                                                                                 
  \begin{aligned}                                                                                                                  
  P(v) &=                                                                                                                          
  \begin{cases}                                                                                                                    
  \dfrac{\displaystyle\sum_{c\in\sC_v} m_v^{(c)}\,w_v^{(c)}                                                                        
  \;+\;\displaystyle\sum_{t\in\sT_v}\mathbbm{1}\!\bigl[t_v^{(t)}\ge\tau_\text{thin}\bigr]\,t_v^{(t)}}                              
  {\displaystyle\sum_{c\in\sC_v} w_v^{(c)}\;+\;\displaystyle\sum_{t\in\sT_v}\mathbbm{1}\!\bigl[t_v^{(t)}\ge\tau_\text{thin}\bigr]},
  & \text{denom}>\varepsilon,\\[6pt]                                                                                               
  0, & \text{otherwise,}                                                                                                           
  \end{cases}\\[8pt]                                                                                                               
  \sV_\text{seg} &= \bigl\{\, v : P(v) \ge \tau_\text{cons} \,\bigr\},                                                             
  \end{aligned}                                                                                                                    
  \label{eq:compact-fusion}                                                                                                        
  \end{equation}

where $m_v^{(c)}$ is the refined SAM confidence at the projected pixel, $w_v^{(c)}$ the depth weight, $t_v^{(t)}$ the thin-structure score, and $\tau_\text{thin}$ is the thin-structure acceptance threshold, and $\sV_\text{seg}$ the set of voxels reported as segmented at the consensus threshold $\tau_\text{cons}$. Equation~\ref{eq:compact-fusion} the \emph{compact fusion} rule is shared verbatim. What differs is how the occupancy grid $\rho$ is constructed and how the depth-consistency block is expressed in each backbone's native depth semantics. We detail the two realizations below.

% ----------------------------------------------------------------
\subsubsection{NeRF backbone 3D Fusion}
\label{sec:method-nerf-fusion}
For the NeRF backbone, occupancy comes from the ray-marched density field, and the depth-consistency block of Equation~\ref{eq:compact-fusion} is expressed in the renderer's per-pixel depth tuple $(\mD_\text{min}, \mD_\text{exp}, \mD_\text{max}, \mD_\text{mw}, \mN_\text{samp})$. A voxel votes through the thick path when its center projects within an adaptive spatial and depth tolerance of the rendered surface, weighted by a Gaussian falloff from the surface midpoint. Voxels whose center falls between the sparse mask-active pixels of a thin structure are recovered by a footprint-based thin path that aggregates mask and depth evidence over the voxel's full projected bounding box. The full derivation, the spatial and depth tolerances, the depth-weighted vote, the thin-structure coverage test, and the two kernel algorithms appear in Appendix~\ref{app:nerf-fusion}.

% ----------------------------------------------------------------
\subsubsection{Gaussian-Splatting Surfel Voxel Fusion}
\label{sec:method-gs-fusion}

Unlike the volumetric NeRF backbone, the 2DGS backbone operates on an explicit, anisotropic, and unbounded surfel representation, so the NeRF fusion kernel cannot be applied directly. The GS instantiation introduces representation-specific adaptations that address surfel anisotropy, thin-structure handling, and surfel-to-voxel propagation. These adaptations constitute the primary technical contribution of the 2DGS extension.

\paragraph{Occupancy by Morton-scattered opacity.}
\label{para:occupancy_gs}
The 2DGS occupancy grid is a scene-adaptive compression layer that turns millions of explicit surfels into a compact voxel field on which a single interactive segmentation kernel is feasible. Its design goal is to derive every quantity, the center, the extent, the voxel edge, and the occupancy threshold, from per-scene statistics so that it generalizes across scene scales without manual tuning.

\noindent\textbf{(i) Robust scene normalization.}
We first estimate a stable working frame directly from the surfel cloud using percentile-based estimators. The scene center $\vc_\text{scene}$ is the \emph{coordinate-wise median} of the Gaussian centers $\{\boldsymbol{\mu}_i\}_{i=1}^{m}$, and the scene extent is the \emph{$99$th percentile} of the radial distance to that center to avoid floaters,
\begin{equation}
\vc_\text{scene} = \text{median}_i(\boldsymbol{\mu}_i),
\qquad
r_\text{fg} = \mathcal{P}_{99}\!\bigl(\{\|\boldsymbol{\mu}_i - \vc_\text{scene}\|\}_{i=1}^{m}\bigr),
\qquad
b = \lceil 1.05\, r_\text{fg}\rceil,
\label{eq:scene-norm}
\end{equation}
The coordinate-wise median and the $99$th-percentile radius are insensitive to this outlier tail, so a few floaters shift neither the working origin $\vc_\text{scene}$ nor the symmetric cube half-extent $b$, which carries a five-percent margin to absorb borderline surfels. Gaussians outside the cube are removed by an AABB filter, and all downstream work proceeds in the re-centered frame.

The voxel edge is then set automatically from per-scene Gaussian statistics, combining the surfel-local resolution, the median Gaussian scale $\bar s$, with a scene-global lower bound:
\begin{equation}
h_v = \max\!\bigl(\alpha_v\,\bar s,\ \beta_v\,r_\text{scene}/g_0\bigr),
\qquad (\alpha_v,\,\beta_v,\,g_0) = (3.0,\, 1.0,\, 256),
\label{eq:voxel-size}
\end{equation}
The first term $\alpha_v\bar s$ follows the Gaussian $3\sigma$ rule so that a typical surfel lands inside one voxel rather than fragmenting across neighbors, while the second term $\beta_v r_\text{scene}/g_0$ caps grid growth on large scenes to keep GPU memory bounded. Their maximum balances geometric fidelity against memory with no scene-specific tuning, and the resolution $g$ is the smallest power of two with $g\ge 2b/h_v$, capped at $g_\text{max}$.

Surviving surfels are aggregated into the density grid $\boldsymbol{\rho}$ by a Morton-indexed opacity scatter-add. A voxel acts as an aggregator that stores only the summed opacity of its surfels, not their identities, and per-surfel labels are recovered later by re-projecting the segmented grid (Section~\ref{sec:method-vox2surfel}). Voxels above a transmittance floor $\tau_\text{TF}$ form a compact active-voxel list, in practice well under one percent of the dense $g^3$ grid, and every downstream kernel runs only on this list. This aggregation is what makes a single CUDA fusion kernel interactive. The working state collapses from millions of primitives to a sparse scalar field. This launch cost scales with foreground complexity rather than grid volume. Segmentation is decoupled from costly per-surfel visibility reasoning, and the construction transfers across scene scales because all of its quantities are per-scene statistics.

% \paragraph{Click-band depth check.}
% Each retained centroid view stores the user-defined depth band $[d_\text{min}^\text{clk},d_\text{max}^\text{clk}]$ together with the orbital camera-to-target distance $d_\text{cam}$ (Section~\ref{sec:method-sam}), passed in as a small per-view triple. The voxel is declared \emph{in band} when $d_\text{min}^\text{clk}\!\le\!x_d\!\le\!d_\text{max}^\text{clk}$ and used by the behind-surface gate below. The 2DGS backbone exposes this band-membership check inside the kernel itself because surfel depths are unbounded and the band is the only scene-anchored evidence available to bracket where the object lives along the ray; the NeRF backbone relies on the trained AABB for the same role.

\paragraph{Planar-to-ray depth correction.}
Both renderers report depth as camera-$Z$ (planar), whereas thick-path visibility is evaluated along the viewing ray. For an off-axis pixel the two differ by the obliquity factor $\cos\theta_i$, the cosine of the angle between the camera-forward axis $\hat{\ve}_\text{fwd}^i$ of view $i$ and the unit ray direction $\hat{\vd}_i$ through the projected pixel. We recover the ray-space depths by dividing each planar depth by this factor,
\begin{equation}
\cos\theta_i = \langle \hat{\ve}_\text{fwd}^i,\, \hat{\vd}_i \rangle,
\qquad
d_\bullet^\text{ray} = \frac{d_\bullet}{\cos\theta_i},
\qquad
\bullet \in \{\text{min},\, \text{med},\, \text{max}\}.
\label{eq:planar-to-ray}
\end{equation}
For NeRF this mismatch is largely absorbed by the thin-structure fallback, so its segmentation stays stable. For 2DGS, routing visible voxels through the thin path is undesirable, because it inflates fusion cost and weakens the discriminative power of the thick-path checks on anisotropic-surfel depths. We therefore apply Equation~\ref{eq:planar-to-ray} before the visibility test, so that the thick path correctly handles visible surfel-supported geometry, while the thin path is reserved for genuinely sub-voxel structures such as ribs and fern fronds. The resulting ray-space depths $d_\text{min}^\text{ray}, d_\text{med}^\text{ray}, d_\text{max}^\text{ray}$ feed the spatial gate of Equation~\ref{eq:gs-spatial}.

\paragraph{Footprint statistics via integral images.}
The per-pixel splat-traversal count $n_\text{samp}$ from the 2DGS rasterizer is the only scene-geometry signal the kernel sees about how the visible surface is built at a pixel, and two GS-specific effects make it informative only when summarized over the voxel's projected footprint rather than read at the center pixel. First, depth localization, NeRF fills object interiors, so its sample count grows with depth thickness and warrants \emph{relaxing} the depth tolerance, whereas 2DGS fits surfels only on the visible surface, so a high count signals stacked surfels and \emph{less} certain depth, which warrants the opposite reaction. Second, layered pixels: at narrow inter-part gaps (fern fronds, Trex, ribs, bonsai flowers) a ray crosses both the foreground splats bordering the gap and the background behind, giving an anomalously large $n_\text{samp}$ that flags a false-positive foreground pixel whose depth reference must switch from $d_\text{med}$ to the front-hit $d_\text{min}$. Both effects need the local mean $\bar n_\text{samp}$ and standard deviation $\sigma_n$ over each voxel's footprint, which we obtain in $O(1)$ per voxel from per-view integral images of $n_\text{samp}$ and $n_\text{samp}^2$.

\paragraph{Layered-pixel and see-through detection.}
Two pixel-level conditions signal that the median depth $d_\text{med}$ is \emph{not} the visible surface and must be replaced by the front-hit $d_\text{min}$:
\begin{equation}
\text{layered}\equiv n_\text{samp}>\bar n_\text{samp}+\sigma_n,
\qquad
\text{seethrough}\equiv d_\text{med}>\gamma_\text{ST}\,d_\text{cam},
\label{eq:seethrough}
\end{equation}
The layered predicate flags a splat-count anomaly relative to the local footprint, while the see-through predicate is a transmittance-gap test anchored on the orbital camera-target distance $d_\text{cam}$, which is scene-scale invariant and behaves consistently from LLFF to Mip-NeRF~$360^\circ$ with a single user-tunable $\gamma_\text{ST}$. Both arise from surfel anisotropy under oblique views, a ray can enter a slanted surfel near its front while most of its opacity lies deeper, so $d_\text{min}>d_\text{med}$, and rather than trust an ambiguous central vote, we defer such voxels to the behind-surface gate and the thin path.

\paragraph{Asymmetric depth-reference and depth gate.}
When the pixel is either layered or see-through (and $d_\text{min}$ is trusted), the reference depth switches from $d_\text{med}$ to the front surface $d_\text{min}$. The depth gate becomes one-sided:
\begin{equation}
\text{valid\_depth}=
\begin{cases}
(x_d-d_\text{min})\le t_\text{depth}, & \text{layered or see-through,}\\
\|x_d-d_\text{med}\|\le t_\text{depth}, & \text{otherwise.}
\end{cases}
\label{eq:asym-gate}
\end{equation}
The one-sided form $(x_d-d_\text{min})$, rather than $\|x_d-d_\text{min}\|$ is deliberate. Behind a thin transmissive front, a genuine object voxel still has $x_d>d_\text{min}$, and an absolute value would silently reject it. Empirically, this single change removes the distant false positives seen on the flower and leaf structures of \textit{Bonsai} and \textit{Garden}.

\paragraph{Per-sample depth-tolerance tightening.}
In 2DGS, a high per-pixel splat count signals stacked surfels and therefore \emph{less} certain depth, so the tolerance must \emph{tighten} where the count is anomalously high. We define the tolerance from the local \emph{excess} $E=\max(n_\text{samp}-\bar n_\text{samp},0)$:
\begin{equation}
\eta=\frac{1}{1+\beta E},
\qquad
t_\text{depth}=\gamma\,h_v\,\eta.
\label{eq:tightening}
\end{equation}
where $\beta,\gamma$ are fixed across datasets and $h_v$ is the voxel size. Tightening on the excess rather than the raw $n_\text{samp}$ prevents uniformly thick objects, where every footprint pixel has a similar count, from being penalized across the board, while still precisely shrinking the band where stacked surfels make the depth ambiguous.

\paragraph{Spatial gate.}
In the spatial check, the voxel center is projected onto the camera ray, $t^\star=\langle\vx_c-\vo_i,\hat{\vd}_i\rangle$, and clamped to the valid surface segment, $\hat t=\mathrm{clip}(t^\star,d_\text{min}^\text{ray},d_\text{ref}^\text{ray})$, with $d_\text{ref}^\text{ray}=d_\text{min}^\text{ray}$ when the pixel is layered or see-through and $d_\text{med}^\text{ray}$ otherwise. Here, $d_\text{min}^\text{ray}, d_\text{med}^\text{ray}$ are obtained from the planar depths $d_\text{min}, d_\text{med}$ through Equation~\ref{eq:planar-to-ray}. The clamped surface point $\vp^\star=\vo_i+\hat t\hat{\vd}_i$ must lie within the depth-gradient-modulated tolerance,
\begin{equation}
\|\vx_c-\vp^\star\|\le h_v\,\pi_\text{dg}(u_x^i,v_x^i),
\label{eq:gs-spatial}
\end{equation}
with the depth-gradient factor $\pi_\text{dg}$ that relaxes the tolerance on flat surfaces and tightens it at depth discontinuities.

\paragraph{Depth-weighted vote and view-coverage bitmask.}
A voxel that clears both gates votes with a boundary-attenuating Gaussian weight that is maximal at the surface-segment center and decays toward its edges,
\begin{equation}
w_\text{depth}^{(i)}=\exp(-\alpha_1\,\hat r^2),
\qquad
\hat r=\frac{\|\hat t - d_\text{ctr}^\text{ray}\|}{d_\text{hw}^\text{ray}},
\label{eq:gs-depthweight}
\end{equation}
with segment center $d_\text{ctr}^\text{ray}=\tfrac12(d_\text{min}^\text{ray}+d_\text{max}^\text{ray})$ and half-width $d_\text{hw}^\text{ray}=\tfrac12(d_\text{max}^\text{ray}-d_\text{min}^\text{ray})$ where $d_\text{max}^\text{ray}$ is ray-space equivalent of planar depth $d_\text{max}$. On every accepted thick vote, the kernel updates the running accumulators \emph{and} sets bit $i$ of the per-voxel view-coverage bitmask $\vv_x$ for the voxel $x$,
\begin{equation}
W_x \mathrel{+}= m_x^i\,w_\text{depth}^{(i)},\qquad
Z_x \mathrel{+}= w_\text{depth}^{(i)},\qquad
\vv_x \mathrel{|}= (1\!\ll\!i),
\label{eq:vmask}
\end{equation}
recording which views actually voted for $x$. This bitmask has no NeRF analogue and is the bridge from voxel space to surfel space (Section~\ref{sec:method-vox2surfel}).

\paragraph{Behind-surface gate before the thin path.}
Before the thin-path fallback, an asymmetric far-gate removes voxels lying strictly behind the visible surface:
\begin{equation}
d_\text{surf}=\begin{cases}d_\text{min}, & d_\text{min}\text{ trusted and see-through,}\\ d_\text{med}, & \text{otherwise,}\end{cases}
\qquad
\text{behind-fail}\equiv x_d>d_\text{surf}+2\,h_v.
\label{eq:behind-gate}
\end{equation}
Rejecting only voxels more than two voxel edges behind the surface, while leaving voxels in front of a far background pixel eligible, discards the bulk of the behind-surface false positives that the central pixel could not see, yet keeps genuine thin foreground that the central pixel happened to miss reachable by the thin path.

\paragraph{Thin-structure path.}
The thick path tests consistency at the voxel center only, so slender high-density structures (Trex ribs, bonsai flowers) project to too few center-aligned pixels and are silently rejected. The thin path instead aggregates depth- and mask-consistent evidence over the voxel's full projected footprint, gated by eligibility on the behind-surface test and the thin-structure density floor $\rho_\text{thin\_thresh}$. Over the footprint, the kernel keeps two counts, the depth-consistent pixels $n_\text{depth}$ and the SAM-and-depth-consistent pixels $n_{m\wedge d}$, and, to guard against 2DGS depths drifting along oblique surfaces, it unprojects each mask-positive pixel and tracks its squared distance to the voxel center together with the peak SAM confidence,

\begin{equation}
d_\text{min-to-surf}^2 \;\leftarrow\; \text{min}\!\bigl(d_\text{min-to-surf}^2,\,\|\vp_w^{(x_p,y_p)}-\vx_c\|^2\bigr),
\qquad
m_\text{max}\leftarrow\max(m_\text{max},\widetilde{\mM}_i(x_p,y_p))
\label{eq:thin-d2min}
\end{equation}
The unprojection has no NeRF analogue and is essential, since even depth-consistent pixels can unproject far from the voxel center on a tilted surface, and the distance term prevents certifying voxels whose footprint coverage is real but geometrically detached.

\textbf{Degeneracy floor, distance gate, final score.} Three conditions close the thin path. First, the depth-consistent pixels must contain sufficient support,
\begin{equation}
n_\text{depth}\ge n_\text{min},
\qquad
n_\text{min}=\text{max}\!\bigl(n_\text{abs},\,\lceil\rho_\text{occ}\,A_\sB\rceil\bigr),
\label{eq:thin-nmin}
\end{equation}
where $A_\sB=|\sB_x^i|$ is the pixel area of the projected footprint $\sB_x^i$. The absolute floor $n_\text{abs}{=}25$ suppresses sparsely sampled footprints, and the occupancy fraction $\rho_\text{occ}{=}0.1$ requires larger footprints to contribute proportionally more evidence, together rejecting isolated pixels and rasterization artifacts. Second, the coverage ratio is taken over the depth-consistent support,
\begin{equation}
p_\text{covered}=\frac{n_{m\wedge d}}{n_\text{depth}}.
\label{eq:gs-coverage}
\end{equation}
Dividing by $n_\text{depth}$ rather than by the footprint area $A_\sB$ prevents large geometrically irrelevant regions (background, occluders) from deflating the ratio, so it answers ``among the footprint parts that could plausibly belong to this voxel, how many does SAM call foreground?''. Third, the closest mask-valid surface intersection must be geometrically near the voxel center. When all three hold, the thin score is the peak footprint confidence, otherwise zero:
\begin{equation}
t_x^{\,i}=
\begin{cases}
m_\text{max}, & n_\text{depth}\ge n_\text{min}\ \wedge\ p_\text{covered}\ge\rho_\text{cover}\ \wedge\ d_\text{min-to-surf}\le (1.5\,h_v)^2,\\
0, & \text{otherwise.}
\end{cases}
\label{eq:thin-tthin-gs}
\end{equation}
The thin path enters the compact-fusion accumulator with the value $t_x^i$. The complete implementation, the preprocessing pipeline, and the CUDA execution scheme are provided in Appendix~\ref{app:gs-impl-details}, and algorithms in Algorithm~\ref{alg:gs-fusion}, Algorithm~\ref{alg:gs-thin}, and Appendix~\ref{app:gs-algorithms}.

% ================================================================
\subsection{Voxel-to-Surfel Mapping and Aggregation}
\label{sec:method-vox2surfel}

\paragraph{Why per-surfel labels are needed.}
For the volumetric NeRF backbone the segmented voxel grid \emph{is} the output. For the explicit 2DGS backbone, however, the deliverable is a per-Gaussian foreground indicator $\vf\in\{0,1\}^{M}$, used both to recolor surfels for the live overlay and to produce the final masked render, and a naive ``label every Gaussian that intersects a segmented voxel'' is geometrically inadequate because voxelization quantizes each anisotropic surfel to a single home voxel, so a disc that legitimately overlaps a segmented voxel may have its center in an adjacent background voxel, and elongated surfels fitted inside one voxel extend far beyond it and project as visible spikes when their home voxel is background. We therefore build $\vf$ as a monotone chain of three filter passes, $\vf^{(1)}\succeq\vf^{(2)}\succeq\vf$, each of which only turns surfels off: Pass $1$ admits a permissive starting set via a Morton-indexed bitfield lookup, Pass $2$ culls per-scene needle floaters, and Pass $3$ culls residual boundary bleeders by footprint-coverage analysis, yielding the final indicator $\vf\equiv\vf^{(3)}$(we use $\vf$ and $\vf^{(3)}$ interchangeably in what follows).

\paragraph{Pass 1: permissive Morton-indexed admission.}
To guarantee the mapping is geometrically identical to grid construction, so that a Gaussian is never assigned to a voxel it was not voxelized into, we re-execute the same NaN/AABB filter and Morton encoding on the live Gaussian set and read the segmentation bitfield $\tB^\text{seg}$ packed from $\tP$:
\begin{equation}
\boldsymbol{\mu}_i^\text{shft}=\boldsymbol{\mu}_i-\vc_\text{scene},
\quad
m_i=\texttt{morton3D}\!\bigl(\lfloor(\boldsymbol{\mu}_i^\text{shft}+b)/h_v\rfloor\bigr),
\quad
\vf_i^{(1)}=\tB^\text{seg}_{m_i}.
\label{eq:pass1}
\end{equation}
The pass is deliberately permissive: it admits every Gaussian whose center falls in a segmented voxel, including boundary surfels whose center sits just inside a segmented voxel while much of their splatted footprint projects to the background. Leaving these in would bleed into the masked render, so their removal is deferred to later passes. Pass $1$ mainly shrinks the candidate set to surfels on or near the object's surface.

\paragraph{Pass 2: scene-adaptive needle-floater culling.}
2DGS training minimizes photometric loss without object awareness, so it leaves needle-like surfels whose aspect ratio
\begin{equation}
a_i=\max(s_{i,1},s_{i,2})/(\min(s_{i,1},s_{i,2})+\epsilon)
\label{eq:aspect}
\end{equation}
exceeds roughly $1000$. These violate the one-surfel per-voxel assumption used to size $h_v$, spike density in unrelated voxels through their long axis, and project as visible spikes even when their opacity contribution to the segmented voxel is negligibl e. A fixed cull threshold either over-prunes or under-prunes across scenes, so we use a self-calibrating $99$th-percentile threshold
\begin{equation}
\tau_a=\mathcal{P}_{99}(\{a_i\}_{i=1}^{M}),
\label{eq:needle}
\end{equation}
and apply it as
\begin{equation}
\mathbb{F}_i^{(2)}=\mathbb{F}_i^{(1)}\wedge\bigl(a_i\le\tau_a\bigr).
\label{eq:pass-needle}
\end{equation}
Anchoring on the per-scene aspect distribution always discards exactly its top $1\%$ tail, and the cull is applied only to the segmentation-time surfel pool, not the photorealistic render. We place it at the mapping stage rather than at grid construction because a needle's harm to segmentation quality appears only when the masked viewpoint is rendered, and removing needles first also lowers the dominant Pass $3$ gather cost.

\paragraph{Pass 3: footprint-coverage culling of boundary bleeders.}
The remaining failure mode is boundary bleeders and surfels surviving Pass $2$, whose splatted disc mostly covers pixels that SAM never labeled as foreground. Pass $3$ measures this coverage directly with a tile-sorted gather that reuses the 2DGS rasterizer's tile decomposition but is semantically inverted: it counts coverage rather than blending color, so it carries no depth ordering, and its tile-sort keys hold only the tile id,
\begin{equation}
\text{key}_{(t,i)}=\text{tile\_id}(t)\ll 32,
\label{eq:keys-no-depth}
\end{equation}
Two design choices make the coverage geometrically faithful and thin-feature preserving. First, each candidate surfel is gated by the per-voxel view-coverage bitmask $\vv$ (Equation~\ref{eq:vmask}), so a surfel is scored in a view only when its mapped voxel was actually voted positive there, restricting evidence to genuinely contributing views. Second, the gather enumerates pixels within a tight $1\sigma$ projected box rather than the rasterizer's $3\sigma$ box: thin and distal structures (chair legs, fern fronds, Trex ribs) have a $1\sigma$ core that lands cleanly on the SAM mask but a $1\sigma\!\to\!3\sigma$ tail that bleeds into background, so restricting the count to the core both preserves these features and decisively culls near-boundary background surfels whose core lies off the mask, while the per-pixel acceptance inside the box stays the rasterizer's exact $3\sigma$ Cram\'er ellipse test. On each accepted pixel, the kernel accumulates two per-surfel counts, the visited footprint pixels and those SAM labels foreground,
\begin{equation}
\text{total\_fp}_i \mathrel{+}= 1, \qquad
\text{if}\ \widetilde{\mM}_s(x_p,y_p)\ge\tau_M \text{then } \text{mask\_fp}_i \mathrel{+}=1
\label{eq:cov-atomics}
\end{equation}
with no opacity, transmittance, or alpha-blending. The per-surfel coverage is the foreground fraction over visited pixels, folded across views by a running maximum,
\begin{equation}
\kappa_s^{(i)}=\frac{\text{mask\_fp}_i^{(s)}}{\max(\text{total\_fp}_i^{(s)},1)},
\qquad
\kappa_i \;\leftarrow\; \max\!\bigl(\kappa_i,\,\kappa_s^{(i)}\bigr).
\label{eq:kappa}
\end{equation}
where the maximum, rather than the mean, is taken because a single clean view is sufficient evidence of membership and averaging would penalize surfels occluded in other views. The final indicator then rejects bleeders below a coverage threshold $\tau_\kappa$,
\begin{equation}
\mathbb{F}_i=\mathbb{F}_i^{(2)}\wedge\bigl(\kappa_i\ge\tau_\kappa\bigr)
\label{eq:final-F}
\end{equation}
Because $\kappa_i$ measures each surfel's own foreground footprint coverage, $\tau_\kappa$ thresholds every surfel on its individual area rather than applying a single strict global score threshold across all primitives, as the optimization-based GS baselines do. Foreground membership is instead settled upstream by the multi-view voxel-occupancy consensus, so $\tau_\kappa$ acts purely as a per-surfel coverage cull and not as the background-rejection knob, a decoupling we analyze under render-resolution change in Appendix~\ref{app:gs-additional}.
After this, the masked render recolors the segmented Gaussians and composites the thresholded overlay on the RGB shown to the user after each centroid view. The complete implementation, the preprocessing, the tile-sorted gather pipeline, and the CUDA execution are provided in Appendix~\ref{app:vox2surfel-impl}, with the full pseudocode in Algorithm~\ref{alg:vox2surfel} (Appendix~\ref{app:gs-algorithms}). The chain preserves geometric continuity and topology while enforcing surfel-level mask consistency.

% ================================================================
\subsection{Iterative Mask Propagation}
\label{sec:method-propagation}

\paragraph{Shared propagation.}
After fusing the current view, DivAS projects the fused $3$D occupancy back onto the next centroid view and overlays it, so the user immediately sees what has been
captured and re-prompts only the missing regions. Views are processed sequentially, each fusion both incorporates a new viewpoint and corrects residual
gaps from earlier views, so sparse user input progressively expands to complete object coverage. The session terminates once all centroid views are processed, and
the segmented voxels (NeRF) or surfels (2DGS) are returned as the final $3$D mask.

\paragraph{Representation-specific execution.}
The backbones differ only in how the overlay is produced and timed, mirroring the SAM-scheduling trade-off of Section~\ref{sec:method-sam}. DivAS-NeRF relies on the \emph{asynchronous} SAM cache. Masks accumulate in the background and are fused in batches, so propagation feedback occurs only when a group synchronizes. DivAS-GS propagates in \emph{real time}, so after every centroid view, it runs the fusion kernel and the three-pass voxel-to-surfel mapping, recoloring segmented Gaussians for an immediate live overlay. In both cases, the per-update segmentation kernel stays below $70$\,ms across all benchmark scenes (Table~\ref{tab:appendix-divas-gs-efficiency}); this is the only per-update cost, and it grows with the number of committed centroid views rather than with $N_\text{train}$.

\medskip
\noindent In summary, DivAS is a single interactive-segmentation framework whose view scheduler, SAM interface, geometry-aware voxel fusion, and propagation loop
are shared, and whose only representation-specific components are the rendering and occupancy adapters. The NeRF and 2DGS instantiations differ in concrete,
well-motivated ways fixed-fraction vs.\ occlusion-aware centroid views, bounded normalization vs. band-pass depth weighting, asynchronous vs.
synchronous SAM, ray-marched density vs. scatter-and-map occupancy, and a 2DGS-only voxel-to-surfel stage. But these are realizations of one formulation, not two pipelines.

% ================================================================
%  Experiments.tex  --  DivAS-GS  (TMLR)
%  Drop-in section file:  \input{Experiments}  inside main.tex.
%  Requires:  booktabs, multirow, graphicx, amsmath (loaded by tmlr.sty
%  or by the main preamble).
% ================================================================

\section{Experiments}
\label{sec:experiments}

\subsection{Experimental Setup}
\label{sec:exp-setup}

\paragraph{Datasets.}
We evaluate our method on two types of scenes: bounded and unbounded. For bounded scenes, we use the LLFF~\citep{mildenhall2019local} dataset, which contains multiple forward-facing real-world scenes.  For unbounded scenes, we use the Mip-NeRF $360^\circ$~\citep{barron2022mip} dataset, which includes complex, large-scale environments.  To enable quantitative evaluation, we use the NVOS~\citep{ren2022neural} dataset. NVOS provides a hand-curated ground-truth mask for a single test view per scene in LLFF. For Mip-NeRF $360^\circ$, no true mask exists, so following SANeRF-HQ \cite{liu2024sanerf}, we evaluate on the publicly released
Mip-NeRF 360 benchmark masks distributed by the authors. These masks are part of the benchmark package and were generated using the annotation pipeline described in SANeRF-HQ rather than being the inference outputs of SANeRF-HQ itself. All methods are evaluated against the same released masks. We considered $6$ representative scenes from each dataset.

\paragraph{Baselines.}
We instantiate the DivAS framework on two scene representations and refer to the two instantiations as \textbf{DivAS} (InstaNGP~\citep{muller2022instantngp} backbone) and \textbf{DivAS-GS} (2DGS~\citep{huang20242d} backbone). We compare them against five representative optimization-based 3D-segmentation baselines that span both representations. On the \textbf{NeRF} side:
(a) \textbf{SA3D}~\citep{cen2023segment}, which lifts SAM masks into a TensoRF~\citep{chen2022tensorf} representation through per-scene mask-inverse-rendering optimization, (b) \textbf{SANeRF-HQ}~\citep{liu2024sanerf}, which aggregates SAM masks in a high-quality InstaNGP~\citep{muller2022instantngp} feature field. On the \textbf{Gaussian-Splatting} side: (c) \textbf{Gaussian Grouping}~\citep{ye2024gaussian}, which trains a per-Gaussian identity embedding supervised by a 2D mask tracker (DEVA), (d) \textbf{SAGA}~\citep{cen2025segment}, which learns a contrastive affinity feature field over the Gaussian scene and (e) \textbf{SA3D-GS}~\citep{cen2024segment3dradiancefields}, the 3DGS~\citep{kerbl20233d} backbone variant of SA3D, which optimizes a per-Gaussian mask score by inverse rendering. We evaluate the official \texttt{sa3d-gs} implementation released by the original SA3D authors through the public SegmentAnythingin3D repository. Since no separate peer-reviewed publication exists for this extension, we follow the released implementation and default settings provided by the authors. DivAS and DivAS-GS are both training-free instantiations of the single framework we introduce in this paper, and we report them together to demonstrate that the same pipeline transfers across scene representations rather than to serve as external points of comparison. All five baselines are run with their official open-source implementations and default hyperparameters. Gaussian Grouping, in particular, incurs a workflow cost that the IoU numbers alone do not capture. After the per-scene identity field is trained, the user has no GUI access to select a desired 3D object and must perform a linear scan of the rendered identity map across the scene's dataset images to identify which identity IDs correspond to the target. A single object is frequently split across multiple IDs (in the \textit{Garden} scene, we had to collect over a dozen disjoint IDs). These identity IDs are an artifact of Gaussian Grouping's own training and are specific to each trained model. Our reported Gaussian Grouping numbers apply a ``Union Merging'' step at evaluation time, merging all these foreground IDs into the final mask so that the results represent the \emph{best-case} performance of the method. A naive user might miss some identities, leading to holes in the final 3D mask. A timing asymmetry exists for methods that do not provide an integrated interactive interface. In SA3D-GS, the user supplies the initial object pixel before propagation begins, and the reported runtime therefore excludes the time required to identify the target location in the rendered scene. Similarly, Gaussian Grouping assumes that foreground Gaussian identities have already been identified from the rendered identity maps prior to execution, and this manual selection step is not included in the reported runtime. By contrast, SAGA and DivAS integrate target selection into the interactive workflow, so user interaction is naturally included in the measured end-to-end time. We follow the original evaluation protocols of all methods when reporting runtimes.

\paragraph{Evaluation protocol.}
\label{par:eval-protocol}
We evaluate each method at its recommended operating point and explicitly report the corresponding configuration. For methods whose performance depends on render resolution or threshold selection, we additionally provide a sensitivity analysis. For DivAS, SA3D, SANeRF-HQ, DivAS-GS, SA3D-GS, and Gaussian Grouping, the predicted mask is rendered \emph{directly at the native ground-truth (full) resolution} from the scene representation, against the NVOS ground-truth masks, no post-hoc upsampling of the predicted mask is performed at any stage. SAGA, by contrast, is strongest at its training resolution rather than at full resolution. We therefore report SAGA at its train-resolution operating point, and to keep the comparison free of information-free interpolation, we downsample the ground-truth mask to the render resolution rather than upsampling the prediction. We thus report a sensitivity analysis (Section~\ref{sec:exp-sensitivity}) that sweeps over resolution and threshold for all three density-thresholding methods. For DivAS-GS, the voxel-occupancy grid construction ($\sim 500$\, ms) is launched asynchronously and overlaps with the SAM model and UI loading at session start. It therefore introduces no additional time for preprocessing and is not counted separately in any reported timing.

\paragraph{Metrics.}
For segmentation quality, we report test-view \textbf{IoU} and pixel \textbf{accuracy} against the ground-truth mask in the LLFF dataset. We report mIoU (mean IoU) and mAcc (mean Accuracy) averaged across all evaluation views per scene in the Mip-$360^\circ$ dataset. For efficiency, we report two metrics that indicate the method's memory and time cost throughout its pipeline. First, the End-to-End Time from scene construction to 3D mask generation. This is because methods like Gaussian Grouping learn identity encodings per-Gaussian during scene reconstruction, whereas SA3D-GS alters the densification rule defined by the vanilla 3DGS method for segmentation. Second, the \textbf{peak VRAM} usage. For every method, we report \textbf{peak VRAM} as the maximum GPU memory observed while running it on a given scene (for the staged baselines SAGA, SA3D-GS, and Gaussian Grouping, this is the maximum taken across all stages of their pipeline), so that hardware-feasibility claims reflect the true worst-case footprint. 

\paragraph{Bounded view processing}
Unlike propagation-based methods that traverse all training views, DivAS processes only a small set of anchor and centroid views selected during interaction across its both instantiations. Table~\ref{tab:view-budget} reports the range of training images available per dataset, together with the number of centroid views processed by DivAS during a segmentation session. Although Mip-NeRF~$360^\circ$ scenes contain up to $301$ training images, DivAS processes at most $20$ centroid views during segmentation. Similarly, LLFF scenes require only $5$-$10$ processed centroid views despite containing up to $61$ training images. This bounded view-processing budget is reflected in the runtime results reported in Section~\ref{app:complexity}.

\begin{table}[h]
\centering
\caption{Training images available versus centroid views actually processed by DivAS during a segmentation session. Despite a large variation in capture size, the processed-view budget stays small and bounded, independent of the training-set size.}
\label{tab:view-budget}
\begin{tabular}{lcc}
\toprule
Dataset & Training Views & Processed Centroid Views \\
\midrule
LLFF                 & $19$-$61$   & $5$-$10$  \\
Mip-NeRF~$360^\circ$ & $175$-$301$ & $8$-$20$  \\
\bottomrule
\end{tabular}
\end{table}

% ================================================================
\subsection{Quantitative Results}
\label{sec:exp-quant}

Table~\ref{tab:summary} gives the overall summary of the results. It reports each method at its best stable operating point, showing both benchmarks, peak memory, and end-to-end runtime in a single view. The per-scene breakdowns that follow support every entry. DivAS-GS and Gaussian Grouping at full (native) resolution, SAGA at its stronger train-resolution setting, and SA3D-GS at full resolution. Gaussian Grouping has the worst performance across the three axes, i.e., IoU/mIoU scores, Peak VRAM, and End-to-End Time. SAGA dominates the mIoU score in the Mip-NeRF $360^\circ$ dataset, second best in the LLFF dataset, but has the maximum Peak VRAM and End-to-End Time close to the Gaussian Grouping time. SA3D-GS performs worst in segmentation quality, with the lowest IoU/mIoU scores, but remains efficient, ranking second-best in Peak VRAM and least in End-to-End Time. Our proposed method, DivAS-GS, achieves optimal performance across the three axes: the best IoU on the LLFF dataset, the second-best mIoU, the least VRAM, and the second-best End-to-End Time, with segmentation quality close to SOTA on Mip-NeRF~$360^\circ$ and better on LLFF.

% ----------------------------------------------------------------
\begin{table}
\centering
\caption{Overall benchmark summary. Each method is reported at the operating point that gives its best \emph{stable} quantitative performance (Section~\ref{par:eval-protocol}). Eval. Res. is Evaluation Resolution. End-to-End Time is in minutes:seconds (mm:ss). Second best per column is \underline{underline}, Best per column is \textbf{bold}.}
\label{tab:summary}
\resizebox{\linewidth}{!}{%
\begin{tabular}{lccccc}
\toprule
Method & Eval. & Mip-NeRF~360$^\circ$ & LLFF & Peak VRAM & End-to-End Time \\
       & Res.  & mIoU $\uparrow$ & IoU $\uparrow$ & (GB) $\downarrow$ & (mm:ss, $360^\circ$ / LLFF) $\downarrow$ \\
\midrule
Gaussian Grouping~\citep{ye2024gaussian} & full  & 0.8764 & 0.8534 & 60.95 & 53:20 / 34:15 \\
SAGA~\citep{cen2025segment}                          & train & \textbf{0.9594} & \underline{0.9151} & 65.52 & 52:42 / 33:28 \\
SA3D-GS~\citep{cen2024segment3dradiancefields}                       & full  & 0.6512 & 0.7502 & \underline{16.96} & \textbf{22:41 / 17:08} \\
DivAS-GS (ours)                          & full  & \underline{0.9476} & \textbf{0.9230} & \textbf{14.40} & \underline{23:55 / 20:03} \\
\bottomrule
\end{tabular}}
\end{table}

We compare DivAS (NeRF backbone) and DivAS-GS (2DGS backbone) against the baselines on both benchmarks. Tables~\ref{tab:cmp-quality-mip360} and ~\ref{tab:cmp-quality-llff} report segmentation quality (IoU and accuracy) per scene for all seven methods, grouped by representation backbone. Tables~\ref{tab:cmp-efficiency-mip360} and~\ref{tab:cmp-efficiency-llff} then report, \emph{per scene}, the efficiency by end-to-end time and peak VRAM for the Gaussian-Splatting methods. We report this per-scene to show the efficiency gain from DivAS-GS. Peak VRAM, in particular, shows that a scene that demands $60 \, \mathrm{GB}$ dictates the required GPU class, despite the low mean.

\paragraph{Backbone of the Gaussian-Splatting comparison.}
Throughout the Gaussian-Splatting comparison, DivAS-GS is evaluated on a 2DGS~\citep{huang20242d} reconstruction, while SAGA, Gaussian Grouping, and SA3D-GS are evaluated using their official 3DGS~\citep{kerbl20233d} implementations. We therefore interpret the reported results as properties of complete segmentation systems rather than isolated segmentation modules. Importantly, the same interaction-and-fusion skeleton also achieves competitive performance against SA3D and SANeRF-HQ in the NeRF setting, suggesting that the observed gains are not solely attributable to the 2DGS backbone.

% ----------------------------------------------------------------
\begin{table}
\centering
\caption{Quantitative comparison on the Mip-NeRF $360^\circ$ dataset. We report mIoU and mAcc scores for different methods. Best (\textbf{bold}) and second-best (\underline{underlined}) are marked \emph{within each representation block}, not across blocks.}
\label{tab:cmp-quality-mip360}
\resizebox{\linewidth}{!}{%
\begin{tabular}{l|cc|cc|cc||cc|cc|cc|cc}
\toprule
 & \multicolumn{6}{c||}{\textbf{NeRF backbone}} & \multicolumn{8}{c}{\textbf{Gaussian-Splatting backbone}} \\
\cmidrule(lr){2-7}\cmidrule(lr){8-15}
 & \multicolumn{2}{c|}{SA3D} & \multicolumn{2}{c|}{SANeRF-HQ} & \multicolumn{2}{c||}{DivAS (ours)} & \multicolumn{2}{c|}{Gaussian Grouping} & \multicolumn{2}{c|}{SAGA} & \multicolumn{2}{c|}{SA3D-GS} & \multicolumn{2}{c}{DivAS-GS (ours)} \\
Scene & mIoU $\uparrow$ & mAcc $\uparrow$ & mIoU $\uparrow$ & mAcc $\uparrow$ & mIoU $\uparrow$ & mAcc $\uparrow$ & mIoU $\uparrow$ & mAcc $\uparrow$ & mIoU $\uparrow$ & mAcc $\uparrow$ & mIoU $\uparrow$ & mAcc $\uparrow$ & mIoU $\uparrow$ & mAcc $\uparrow$ \\
\midrule
Bonsai              & \underline{0.9120} & \textbf{0.9910} & 0.9060 & \underline{0.9890} & \textbf{0.9200} & \textbf{0.9910} & 0.6012 & 0.9555 & \underline{0.9351} & \underline{0.9929} & 0.5608 & 0.9492 & \textbf{0.9396} & \textbf{0.9932} \\
Garden (no vase)    & 0.9030 & 0.9820 & \textbf{0.9520} & \textbf{0.9910} & \underline{0.9430} & \underline{0.9900} & \textbf{0.9745} & \textbf{0.9956} & \underline{0.9694} & \underline{0.9948} & 0.7769 & 0.9542 & 0.9620 & 0.9933 \\
Garden (with vase)  & 0.7580 & \underline{0.9540} & \underline{0.9280} & \textbf{0.9880} & \textbf{0.9360} & \textbf{0.9880} & 0.9567 & 0.9925 & \textbf{0.9726} & \textbf{0.9952} & 0.6672 & 0.9229 & \underline{0.9584} & \underline{0.9926} \\
Kitchen (lego)      & \textbf{0.9700} & \textbf{0.9960} & 0.8360 & 0.9780 & \underline{0.9170} & \underline{0.9880} & \underline{0.9570} & \underline{0.9942} & \textbf{0.9703} & \textbf{0.9959} & 0.8306 & 0.9719 & 0.9569 & 0.9940 \\
Counter (flowerpot) & 0.3640 & 0.9520 & \textbf{0.9020} & \textbf{0.9940} & \underline{0.8900} & \underline{0.9930} & \underline{0.9438} & 0.9958 & \textbf{0.9515} & \textbf{0.9968} & 0.4589 & 0.9522 & 0.9423 & \underline{0.9966} \\
Room (chair)        & 0.8310 & 0.9770 & \textbf{0.9520} & \textbf{0.9990} & \underline{0.8700} & \underline{0.9960} & 0.8254 & 0.9943 & \textbf{0.9576} & \textbf{0.9987} & 0.6126 & 0.9761 & \underline{0.9262} & \underline{0.9976} \\
\midrule
\textbf{Mean}       & \underline{0.7900} & 0.9750 & \textbf{0.9130} & \underline{0.9900} & \textbf{0.9130} & \textbf{0.9910} & 0.8764 & 0.9880 & \textbf{0.9594} & \textbf{0.9957} & 0.6512 & 0.9544 & \underline{0.9476} & \underline{0.9946} \\
\bottomrule
\end{tabular}}
\end{table}

Table~\ref{tab:cmp-quality-mip360} represents the segmentation quality on the Mip-NeRF $360^\circ$ dataset, across both scene representations. All methods are evaluated against the same SANeRF-HQ pseudo-ground-truth masks. The Gaussian-Splatting methods follow the best-stable protocol of Section~\ref{par:eval-protocol}, where SAGA is evaluated at training resolution. Under the NeRF backbone, SA3D performs well in simple object-centric settings (Kitchen), but struggles in complex, multi-material scenes (Garden with a vase) or in occluded scenes (Counter with a flowerpot), where single-view prompts fail to capture the target region across multiple viewpoints. Our approach maintains geometric consistency under such multi-material, occluded scenes by fusing multi-view cues via a 3D voxel kernel, resulting in smoother, more coherent 3D segmentation. Because the Mip-NeRF~$360^\circ$ references are SANeRF-HQ pseudo-labels rather than manually annotated ground truth, they occasionally penalize our IoU in scenes like Counter. The qualitative comparison (see Figure~\ref{fig:qualitative_mip}) shows that these discrepancies tend to coincide with imperfect pseudo-labels around thin or partially occluded structures, where our predictions remain geometrically consistent with the visible object boundaries. Overall, our method matches the quantitative accuracy of SANeRF-HQ~\citep{liu2024sanerf} while being multi-view consistent. This demonstrates the effectiveness of amortized (i.e., optimization-free) 3D fusion without per-scene optimization.

As shown in the GS backbone, SA3D-GS performs significantly worse than its NeRF counterpart, suggesting that the unbounded mask-score objective is more sensitive to floater artifacts and boundary leakage. Gaussian Grouping leads to inconsistent results. In the Garden scene, it achieves the maximum mIoU of $0.975$, whereas in the Bonsai scene it drops to $0.60$. This is because of the cumulative effect of SAM false negatives and the inability of 2D tracker (DEVA) to propagate masks across the views for thin or repeated structures like bonsai petals. On scenes where the target is a single connected blob of solid material (\textit{Garden-no-vase}, \textit{Kitchen-lego}), the tracker is reliable, and Gaussian Grouping is competitive, occasionally beating DivAS-GS on raw mIoU. SAGA is marginally ahead of DivAS-GS, winning four of six scenes with a mean mIoU $0.01$ higher. The largest gap is $0.03$ in the Room (chair) scene, where the floor and chair legs are not well-defined due to poorly learned geometry and reconstruction, leading to ambiguity about whether the query voxel belongs to the chair leg or the floor. 

% ----------------------------------------------------------------
\begin{table}
\centering
\caption{Quantitative comparison on the NVOS dataset. We report IoU and mAcc scores for different methods. Best(\textbf{bold}) and second-best (\underline{underlined}) are marked within each representation block.}
\label{tab:cmp-quality-llff}
\resizebox{\linewidth}{!}{%
\begin{tabular}{l|cc|cc|cc||cc|cc|cc|cc}
\toprule
 & \multicolumn{6}{c||}{\textbf{NeRF backbone}} & \multicolumn{8}{c}{\textbf{Gaussian-Splatting backbone}} \\
\cmidrule(lr){2-7}\cmidrule(lr){8-15}
 & \multicolumn{2}{c|}{SA3D} & \multicolumn{2}{c|}{SANeRF-HQ} & \multicolumn{2}{c||}{DivAS (ours)} & \multicolumn{2}{c|}{Gaussian Grouping} & \multicolumn{2}{c|}{SAGA} & \multicolumn{2}{c|}{SA3D-GS} & \multicolumn{2}{c}{DivAS-GS (ours)} \\
Scene & IoU $\uparrow$ & Acc $\uparrow$ & IoU $\uparrow$ & Acc $\uparrow$&  IoU $\uparrow$ & Acc $\uparrow$ & IoU $\uparrow$ & Acc $\uparrow$ & IoU $\uparrow$ & Acc $\uparrow$ & IoU $\uparrow$ & Acc $\uparrow$ & IoU $\uparrow$ & Acc $\uparrow$ \\
\midrule
Fern           & \underline{0.8220} & \underline{0.9430} & \textbf{0.8250} & \textbf{0.9440} & 0.8030 & 0.9330 & \underline{0.8230} & \underline{0.9420} & \textbf{0.8403} & \textbf{0.9479} & 0.7121 & 0.8842 & 0.8115 & 0.9374 \\
Fortress       & 0.9690 & \underline{0.9940} & \underline{0.9720} & \textbf{0.9950} & \textbf{0.9750} & \textbf{0.9950} & 0.9797 & 0.9962 & \underline{0.9808} & \underline{0.9964} & 0.9442 & 0.9891 & \textbf{0.9883} & \textbf{0.9978} \\
Horns (center) & \underline{0.9730} & \underline{0.9950} & 0.9690 & \underline{0.9950} & \textbf{0.9780} & \textbf{0.9960} & \underline{0.9769} & \underline{0.9960} & \textbf{0.9805} & \textbf{0.9966} & 0.8710 & 0.9747 & 0.9729 & 0.9953 \\
Horns (left)   & \underline{0.9360} & \textbf{0.9960} & \textbf{0.9390} & \textbf{0.9960} & 0.9350 & \textbf{0.9960} & 0.9262 & 0.9954 & \underline{0.9285} & \underline{0.9955} & 0.6428 & 0.9670 & \textbf{0.9351} & \textbf{0.9960} \\
Orchids        & \underline{0.8800} & \underline{0.9790} & 0.7780 & 0.9560 & \textbf{0.8950} & \textbf{0.9820} & 0.8578 & 0.9749 & \underline{0.9365} & \underline{0.9894} & 0.6458 & 0.9388 & \textbf{0.9408} & \textbf{0.9901} \\
Trex           & 0.8310 & 0.9770 & \underline{0.8410} & \underline{0.9790} & \textbf{0.8690} & \textbf{0.9820} & 0.5569 & 0.9430 & \underline{0.8241} & \underline{0.9755} & 0.6851 & 0.9478 & \textbf{0.8891} & \textbf{0.9853} \\
\midrule
\textbf{Mean}  & \underline{0.9020} & \textbf{0.9810} & 0.8870 & \underline{0.9780} & \textbf{0.9090} & \textbf{0.9810} & 0.8534 & 0.9746 & \underline{0.9151} & \textbf{0.9836} & 0.7502 & 0.9503 & \textbf{0.9230} & \textbf{0.9836} \\
\bottomrule
\end{tabular}}
\end{table}

Table~\ref{tab:cmp-quality-llff} evaluates segmentation quality on the NVOS benchmark across both NeRF and Gaussian-Splatting-based scene representations. Following the official training protocol of each method, we use the original inference pipelines rather than the user interaction strokes provided by NVOS. Under the NeRF backbone block, SA3D achieves competitive quantitative performance, indicating that optimization over volumetric voxel geometry remains considerably more stable than its Gaussian-Splatting counterpart. SANeRF-HQ~\citep{liu2024sanerf} shows a sharp performance drop in the Orchids scene. It incorrectly segments unprompted flower buds that visually resemble the user-selected flowers. This over-segmentation arises from its global optimization process, which lacks fine spatial prompt control. In contrast, our centroid-view design restricts segmentation to the prompted regions, enabling localized refinement. As a result, our method avoids false activations on nearby, visually similar structures. Our method, DivAS, achieves the highest mean IoU of $0.909$ and comparable pixel accuracy of $0.981$ among all baselines. Our method dominates in four out of six scenes, highlighting the advantage of interactive centroid-view refinement in reducing false negatives on fine object details.

Within the Gaussian-Splatting backbone, the performance gap between geometry-aware and feature-driven approaches becomes more pronounced on scenes with complex depth interactions. Similar to the observations on Mip-NeRF $360^\circ$, SA3D-GS performs substantially worse than its NeRF counterpart, suggesting that optimization over anisotropic Gaussian primitives is more sensitive to floater artifacts and boundary leakage. Gaussian Grouping shows degraded performance in the Trex scene. It is unable to segment Trex ribs and other parts. It confuses them with the background because of its proximity to the object parts in the image space. SAGA wins in two out of six scenes where there is a $0.02$-point IoU score difference in the fern scene with DivAS-GS at $0.811$ because of poor geometric separation between the background and the semi-transparent fern leaves, leading to background bleeding. SAGA optimizes this in image space at the cost of culling object parts, as shown in Figure~\ref{fig:qualitative_llff_gs}. Our approach maintains geometric consistency and an induced depth prior in the SAM mask, achieving a winning score in 4 of 6 scenes, with a mean IoU that is $0.007$ higher than the second-best SAGA method. The $0.065$ large margin in the Trex scene of SAGA at $0.8254$ is because depth-gated voxel aggregation of DivAS-GS separates the interleaved rib bones and background layers that SAGA's appearance-only affinity features merge. These results confirm that depth-guided SAM masks and voxel-level fusion yield stable and accurate segmentation across forward-facing scenes.

% ----------------------------------------------------------------
\begin{table}
\centering
\caption{Per-scene efficiency comparison on the Mip-NeRF $360^\circ$ dataset. For each method, we report \emph{E2E Time} (end-to-end wall-clock time) and \emph{Peak VRAM}. Best is \textbf{bold}, second-best \underline{underlined}.}
\label{tab:cmp-efficiency-mip360}
\resizebox{\linewidth}{!}{%
\begin{tabular}{l|cc|cc|cc|cc}
\toprule
 & \multicolumn{2}{c|}{Gaussian Grouping} & \multicolumn{2}{c|}{SAGA} & \multicolumn{2}{c|}{SA3D-GS} & \multicolumn{2}{c}{DivAS-GS (ours)} \\
Scene & \shortstack{E2E Time\\(mm:ss)$\downarrow$} & \shortstack{Peak VRAM\\(GB)$\downarrow$} & \shortstack{E2E Time\\(mm:ss)$\downarrow$} & \shortstack{Peak VRAM\\(GB)$\downarrow$} & \shortstack{E2E Time\\(mm:ss)$\downarrow$} & \shortstack{Peak VRAM\\(GB)$\downarrow$} & \shortstack{E2E Time\\(mm:ss)$\downarrow$} & \shortstack{Peak VRAM\\(GB)$\downarrow$} \\
\midrule
Bonsai              & 38:00 & 31.58 & 35:40 & 21.76 & \textbf{15:07} & \underline{8.06} & \underline{17:35} & \textbf{7.98} \\
Garden (no vase)    & 80:00 & 60.95 & 81:27 & 65.52 & \textbf{34:40} & \underline{16.96} & \underline{36:55} & \textbf{10.62} \\
Garden (with vase)  & 80:00 & 60.93 & 81:30 & 65.52 & \textbf{34:41} & \underline{16.74} & \underline{35:02} & \textbf{14.40} \\
Kitchen (lego)      & 42:00 & 29.94 & 43:34 & 23.10 & \textbf{19:01} & \textbf{9.19} & \underline{19:35} & \underline{13.93} \\
Counter (flowerpot) & 35:00 & 22.39 & 36:47 & 24.17 & \textbf{16:22} & \textbf{7.83} & \underline{17:25} & \underline{7.93} \\
Room (chair)        & 45:00 & 31.48 & 37:15 & 24.32 & \textbf{16:14} & \underline{8.50} & \underline{16:58} & \textbf{7.00} \\
\midrule
\textbf{Mean / max-VRAM} & 53:20 & 60.95 & 52:42 & 65.52 & \textbf{22:41} & \underline{16.96} & \underline{23:55} & \textbf{14.40} \\
\bottomrule
\end{tabular}}
\end{table}

Tables~\ref{tab:cmp-efficiency-mip360} and~\ref{tab:cmp-efficiency-llff} compare the computational efficiency of Gaussian-Splatting-based segmentation methods in terms of end-to-end time and peak GPU VRAM usage. In contrast to optimization-heavy feature-field approaches whose computational cost scales with prolonged feature optimization and scene-level processing, DivAS-GS maintains an amortized training-free pipeline with substantially lower end-to-end time overhead.

On the Mip-NeRF $360^\circ$ benchmark, DivAS-GS achieves segmentation quality comparable to the strongest-performing methods in Table~\ref{tab:cmp-quality-mip360} while requiring less than $25\%$ of the peak VRAM consumed by SAGA and reducing end-to-end runtime by approximately $2\times$-$2.2\times$. Compared to Gaussian Grouping, DivAS-GS provides competitive segmentation quality at a lower overall end-to-end cost. Although SA3D-GS achieves the lowest runtime and among the smallest memory footprints, its segmentation quality remains substantially lower than both SAGA and DivAS-GS.

A similar trend is observed on the LLFF dataset, where DivAS-GS maintains the strongest balance between segmentation quality and computational efficiency. Relative to SAGA and Gaussian Grouping, DivAS-GS reduces peak VRAM consumption to below $40\%$ while achieving approximately $1.5\times$-$1.7\times$ faster end-to-end execution. While SA3D-GS remains computationally lightweight, its segmentation accuracy exhibits a substantial gap compared to the results reported in Table~\ref{tab:cmp-quality-llff}. Overall, these results demonstrate that DivAS-GS achieves competitive segmentation quality while remaining significantly more practical for deployment on consumer-grade GPUs than optimization-intensive feature-field baselines.

% ----------------------------------------------------------------
\begin{table}
\centering
\caption{Per-scene efficiency comparison on the LLFF dataset. For each method, we report \emph{E2E Time} (end-to-end wall-clock time) and \emph{Peak VRAM}. Best is \textbf{bold}, second-best \underline{underlined}.}
\label{tab:cmp-efficiency-llff}
\resizebox{\linewidth}{!}{%
\begin{tabular}{l|cc|cc|cc|cc}
\toprule
 & \multicolumn{2}{c|}{Gaussian Grouping} & \multicolumn{2}{c|}{SAGA} & \multicolumn{2}{c|}{SA3D-GS} & \multicolumn{2}{c}{DivAS-GS (ours)} \\
Scene & \shortstack{E2E Time\\(mm:ss)$\downarrow$} & \shortstack{Peak VRAM\\(GB)$\downarrow$} & \shortstack{E2E Time\\(mm:ss)$\downarrow$} & \shortstack{Peak VRAM\\(GB)$\downarrow$} & \shortstack{E2E Time\\(mm:ss)$\downarrow$} & \shortstack{Peak VRAM\\(GB)$\downarrow$} & \shortstack{E2E Time\\(mm:ss)$\downarrow$} & \shortstack{Peak VRAM\\(GB)$\downarrow$} \\
\midrule
Fern           & 33:30 & 16.48 & 32:24 & 14.30 & \textbf{17:50} & \textbf{6.64} & \underline{23:19} & \underline{8.52} \\
Fortress       & 33:30 & 14.57 & 28:34 & 11.52 & \textbf{15:26} & \textbf{6.64} & \underline{17:06} & \underline{7.93} \\
Horns (center) & 34:03 & 16.78 & 34:05 & 17.81 & \textbf{18:09} & \textbf{6.64} & \underline{19:46} & \underline{7.95} \\
Horns (left)   & 34:03 & 16.78 & 33:34 & 17.81 & \underline{18:05} & \textbf{6.64} & \textbf{17:25} & \underline{6.94} \\
Orchids        & 39:41 & 22.23 & 40:01 & 21.96 & \textbf{18:11} & \textbf{6.69} & \underline{23:05} & \underline{8.11} \\
Trex           & 30:45 & 13.76 & 32:07 & 15.17 & \textbf{15:07} & \textbf{6.64} & \underline{19:40} & \underline{7.89} \\
\midrule
\textbf{Mean / max-VRAM} & 34:15 & 22.23 & 33:28 & 21.96 & \textbf{17:08} & \textbf{6.69} & \underline{20:03} & \underline{8.52} \\
\bottomrule
\end{tabular}}
\end{table}

\paragraph{NeRF instantiation efficiency.}
We evaluate the end-to-end runtime of DivAS (NeRF instantiation) against the optimization-based baseline SANeRF-HQ~\citep{liu2024sanerf} on a single NVIDIA A100 GPU. As shown in Table~\ref{tab:simple_timing}, DivAS achieves a consistent $1.7$-$2.1\times$ speedup across diverse scenes. The gain stems from the optimization-free design. While SANeRF-HQ requires substantial computation to train object fields (about $3$ minutes of non-interactive wait time), the DivAS CUDA fusion kernel incurs negligible latency, below $65$\,ms per call across all evaluated scenes (finer-grained kernel timings in Appendix~\ref{app:complexity}). The DivAS runtime is therefore dominated almost entirely by user interaction, enabling a fluid, real-time workflow that scales independently of the dataset size.

\begin{table}
\centering
\caption{\textbf{Total runtime comparison} (NeRF instantiation). Time in seconds. DivAS consistently outperforms the optimization-based baseline by eliminating the training bottleneck.}
\begin{tabular}{l|c|c|c}
\toprule
\textbf{Scene} & \textbf{SANeRF-HQ} & \textbf{DivAS} & \textbf{Speedup} \\
\midrule
Fern (Bounded)      & 231.0 & \textbf{120.0} & 1.92$\times$ \\
Trex (Bounded)      & 327.5 & \textbf{190.0} & 1.72$\times$ \\
Bonsai ($360^\circ$) & 514.0 & \textbf{258.0} & 1.99$\times$ \\
Kitchen ($360^\circ$) & 516.0 & \textbf{250.0} & 2.06$\times$ \\
\bottomrule
\end{tabular}
\label{tab:simple_timing}
\end{table}

\paragraph{Render-resolution and threshold robustness.} DivAS-GS is resolution-stable at its single canonical threshold ($\tau{=}0.1$), with mean IoU changing by at most $0.30$ points between full and training resolution, because it decides foreground membership upstream by multi-view voxel-occupancy consensus and uses $\tau$ only as a binarizer, whereas threshold-based baselines such as SAGA must balance background rejection against thin-structure retention through a single $\tau$. The full sensitivity table and analysis are provided in Appendix~\ref{app:gs-additional}.

% ================================================================
\subsection{Qualitative Analysis}
\label{sec:exp-qualitative}

For the NeRF instantiation, DivAS preserves both global structure and thin parts such as Trex ribs and hands while avoiding the over-segmentation of SANeRF-HQ and the single-view misses of SA3D, on both LLFF and Mip-NeRF~$360^\circ$. The full qualitative comparisons (Figure~\ref{fig:qualitative_llff} and Figure~\ref{fig:qualitative_mip}) are provided in Appendix~\ref{app:nerf-additional}, consistent with our use of the NeRF backbone as evidence that the framework transfers across representations.

\begin{figure}[h]
    \centering
    \includegraphics[width=0.9\linewidth]{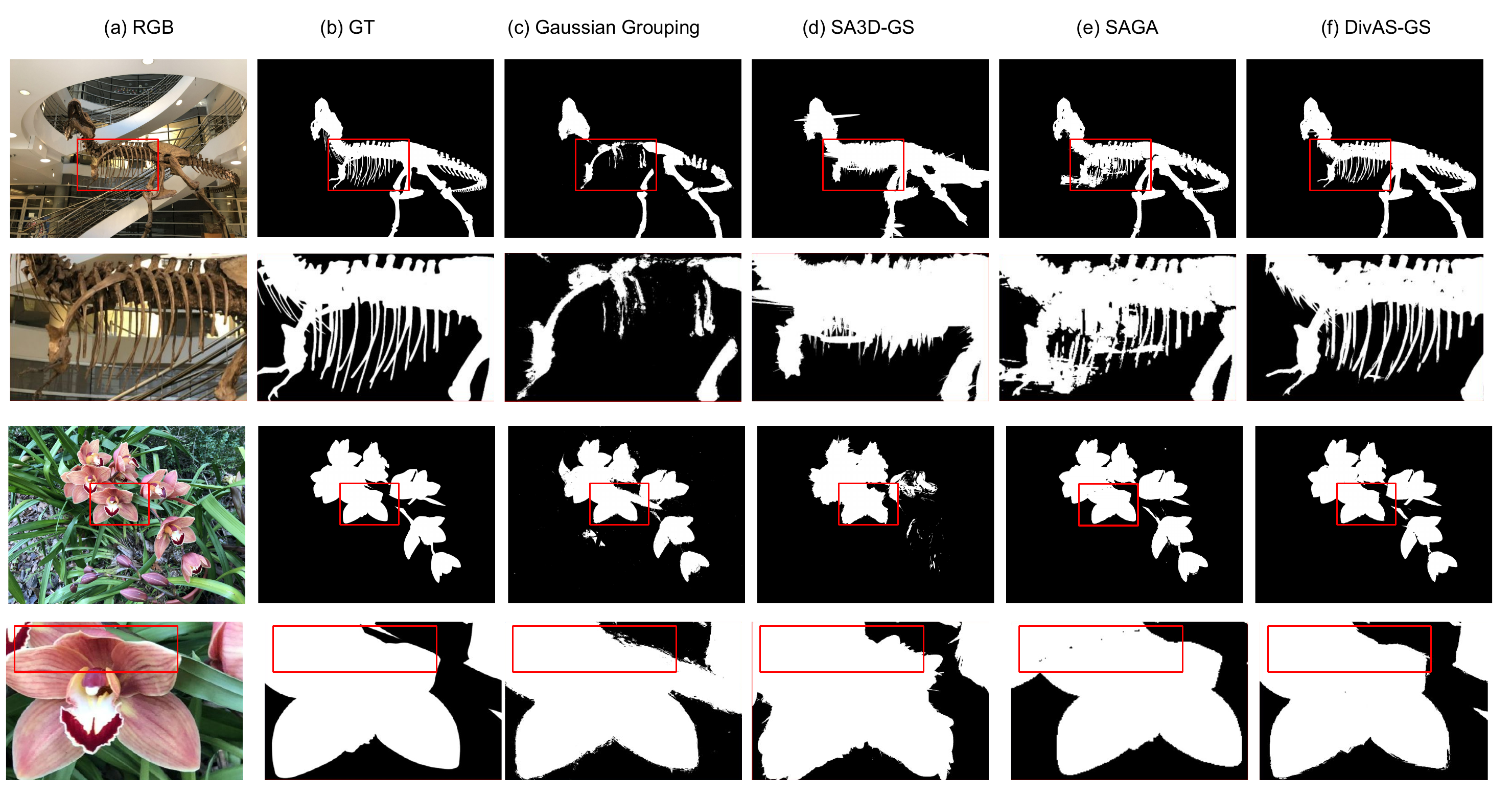}
    \caption{Results of different methods for GS backbone on LLFF.}
    \label{fig:qualitative_llff_gs}
\end{figure}

Figure~\ref{fig:qualitative_llff_gs} visualizes segmentation results on the LLFF dataset for the Gaussian-Splatting backbone. In rows $1$ and $2$ on the \textit{Trex} scene, Gaussian Grouping confuses the rib structures with the background, producing large false negatives in the mask. SA3D-GS misses global parts of the Trex such as the tail and simultaneously exhibits background bleeding into the rib region. We attribute this dual failure to the unbounded 
  mask-score objective, which does not suppress the persistent floaters in its 3DGS~\citep{kerbl20233d} representation. SAGA preserves the global structure (tail, ribs) but cannot separate the background from the ribs in the absence of geometric consistency, producing over-segmentation around ribs and hands. Driven by the depth-weighted mask refinement and the centroid-view strategy, DivAS-GS stays closest to the ground-truth mask, free of the artifacts visible in the other three methods. Row $3$ shows the \textit{Orchids} scene in which  Gaussian Grouping again suffers from background bleeding, while SA3D-GS fails to reconstruct all the orchids in the scene. In row $4$, the red box highlights random holes introduced by SAGA within the solid body of the object, the resolution-threshold pathology described in paragraph~\ref{para:saga_holes}, whereas DivAS-GS remains geometrically consistent, with no background bleeding or holes in the solid body.

\begin{figure}[h]
    \centering
    \includegraphics[width=0.9\linewidth]{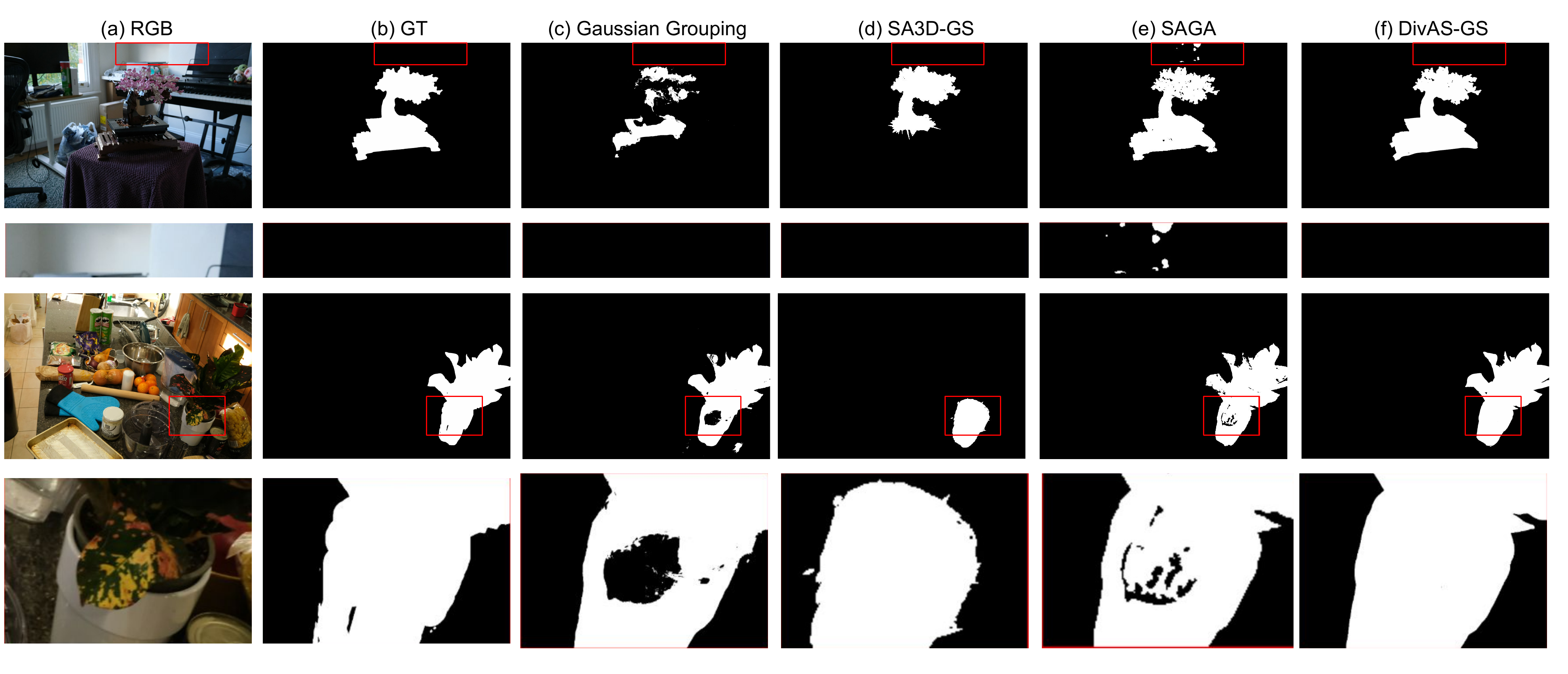}
    \caption{Results of different methods for GS backbone on Mip-NeRF~$360^\circ$.}
    \label{fig:qualitative_mip_gs}
\end{figure}

Figure~\ref{fig:qualitative_mip_gs} visualizes segmentation results on the Mip-NeRF~$360^\circ$ dataset for the Gaussian-Splatting backbone. In row $1$, Gaussian Grouping misses the stem of the plant, confusing it with the background, while SA3D-GS fails to propagate its single-view prompt across viewpoints, producing under-segmentation that drops parts of the object. Row $2$ illustrates the geometric inconsistency of SAGA on the \textit{Bonsai} plant, where background bleeders penetrate the foreground. DivAS-GS remains geometrically consistent with no background bleeders and produces a binary mask close to the ground truth. On the \textit{Counter (flowerpot)} scene in row $3$, SA3D-GS misses the plant leaves due to the occlusion limitations of single-view prompting. In row $4$, Gaussian Grouping treats a yellow leaf as background, producing a hole in the mask, while SAGA again exhibits random holes inside the solid body, degrading the resulting binary mask. DivAS-GS produces a geometrically consistent binary mask close to the references, which on this benchmark are SANeRF-HQ pseudo-labels rather than manual ground truth.

% ================================================================
\subsection{Ablation Studies}
\label{sec:exp-ablation}

We present ablations on two design decisions shared across the NeRF and GS instantiations.
Where the two instantiations produce qualitatively different behavior, we report and explain them separately.

% ---------------------------------------------------------------
\paragraph{Cumulative-weight threshold $\tau_\text{cw}$.}

\begin{figure}[t]
    \centering
    \includegraphics[width=0.95\linewidth]{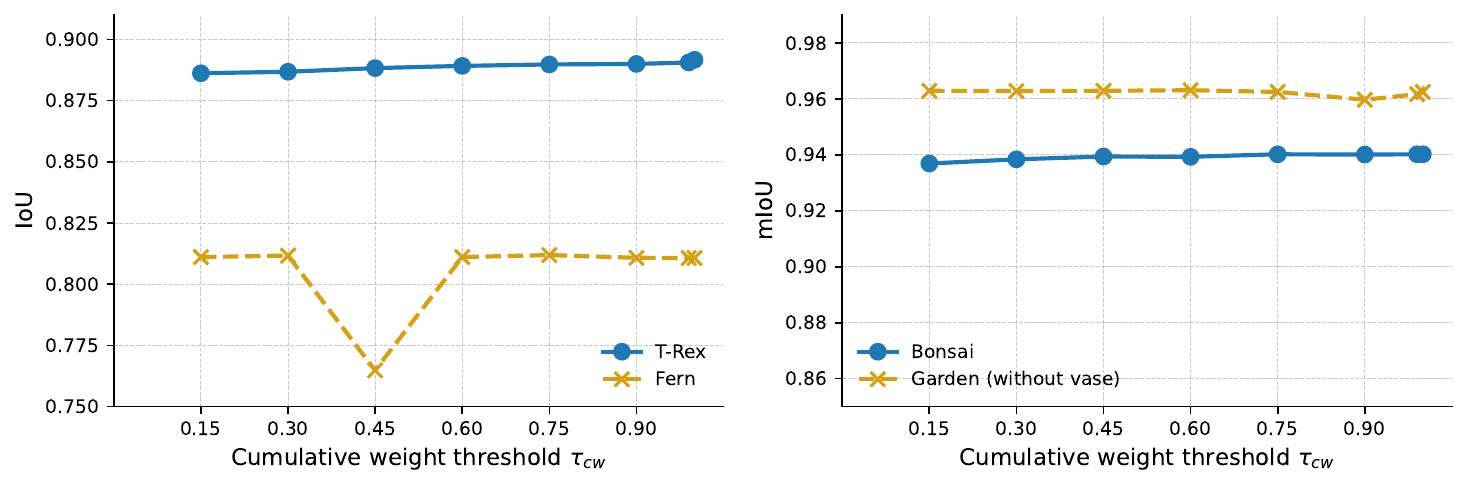}
    \caption{IoU vs.\ cumulative-weight target $\tau_\text{cw}$ for the 2DGS instantiation on representative LLFF and Mip-NeRF~$360^\circ$ scenes.}
    \label{fig:ablation_cumulative_weight_gs}
\end{figure}

The cumulative-weight cutoff $\tau_\text{cw}$ controls the point along each ray at which the renderer declares ``sufficient foreground density has been observed'', determining the upper bound of the depth range $\mD_\text{max}$ fed to the fusion kernel.
We sweep $\tau_\text{cw}$ over $\{0.15-0.9\}$ on four representative scenes \textit{Trex} and \textit{Fern} from LLFF, and \textit{Bonsai} and \textit{Garden (without vase)} from Mip-NeRF~$360^\circ$.

\textbf{NeRF instantiation} The NeRF backbone behaves differently, its IoU stays stable for $\tau_\text{cw}\!\in[0.3, 0.75]$ and degrades at higher cutoffs as the widening valid depth range admits background voxels. We report the full NeRF sweep, figure, and analysis in Appendix~\ref{app:cumulative-weight-nerf}.

\textbf{2DGS instantiation} As shown in Figure~\ref{fig:ablation_cumulative_weight_gs}, the 2DGS instantiation behaves \emph{differently} from NeRF at high $\tau_\text{cw}$: its IoU is essentially flat across the entire sweep, including the $\tau_\text{cw}\!\ge\!0.9$ regime where the NeRF IoU drops sharply. In the 2DGS rasterizer, $\tau_\text{cw}$ controls how many of the surfel layers stacked along the ray (the cumulative alpha accumulation) are folded into $\mD_\text{max}$. Because 2DGS surfels are packed tightly onto the visible surface with very little depth spread, raising $\tau_\text{cw}$ admits almost no additional depth, so $\mD_\text{max}$ and hence the segmentation remain nearly unchanged. This is the opposite of the NeRF behavior, where a high $\tau_\text{cw}$ widens the valid depth range enough to admit background voxels and degrade precision: the volumetric density field carries appreciable mass between the surface and the background, whereas the surfel field does not. Since the 2DGS score is essentially constant over the whole range, we adopt $\tau_\text{cw}{=}0.75$ to match the NeRF default for consistency across instantiations.

% ---------------------------------------------------------------
\paragraph{Thin-structure path and depth-weighted mask refinement.}

\begin{table}[!h]
\centering
\scriptsize
\begin{tabular}{lcccc}
\toprule
 & \multicolumn{2}{c}{\textbf{Thin-structure}} & \multicolumn{2}{c}{\textbf{Depth-weight}} \\
\cmidrule(lr){2-3}\cmidrule(lr){4-5}
\textbf{Scene} & \textbf{w (IoU)} & \textbf{w/o (IoU)} & \textbf{w (IoU)} & \textbf{w/o (IoU)} \\
\midrule
\multicolumn{5}{c}{\emph{NeRF instantiation}} \\
\midrule
\multicolumn{5}{l}{\textit{LLFF Dataset}} \\
Trex                  & \textbf{0.869} & 0.673          & \textbf{0.869} & 0.867 \\
Fern                  & 0.803          & \textbf{0.814} & \textbf{0.803} & 0.801 \\
Horns\_left           & \textbf{0.935} & 0.883          & \textbf{0.935} & 0.929 \\
\multicolumn{5}{l}{\textit{Mip-NeRF~$360^\circ$ Dataset}} \\
Bonsai                & \textbf{0.920} & 0.915          & \textbf{0.920} & 0.918 \\
Garden (without vase) & \textbf{0.943} & 0.899          & 0.943          & \textbf{0.946} \\
\midrule
\multicolumn{5}{c}{\emph{2DGS instantiation}} \\
\midrule
\multicolumn{5}{l}{\textit{LLFF Dataset}} \\
Trex                  & \textbf{0.889} & 0.597          & \textbf{0.889} & 0.885 \\
Fern                  & \textbf{0.812} & 0.544          & \textbf{0.812} & 0.809 \\
\multicolumn{5}{l}{\textit{Mip-NeRF~$360^\circ$ Dataset}} \\
Bonsai                & \textbf{0.940} & 0.822          & \textbf{0.940} & 0.938 \\
Garden (without vase) & \textbf{0.962} & 0.763          & \textbf{0.962} & 0.962 \\
\bottomrule
\end{tabular}
\caption{Ablations on the thin-structure handling path and depth-based SAM mask weighting, reported side-by-side. Bold marks the higher IoU within each ablation pair independently. Removing the thin-structure module reduces IoU, especially on scenes with fine details (e.g., \textit{Trex}, \textit{Garden}), and the drop is markedly larger for the 2DGS instantiation. Removing the depth-weight results in small but consistent IoU drops, with stronger qualitative effects (faint boundaries or background bleeding).}
\label{tab:ablation_thin_depth}
\end{table}

The left half of Table~\ref{tab:ablation_thin_depth} shows the impact of the thin-structure footprint path for both instantiations.
For the NeRF instantiation, removing the module causes a catastrophic drop on \textit{Trex} ($0.869 \to 0.673$) and a substantial drop on \textit{Horns\_left} and \textit{Garden}.
The \textit{Fern} scene shows a slight IoU degradation with the module ($0.803$ vs.\ $0.814$), which we attribute to high depth uncertainty in the pretrained NeRF for this scene: the relaxed 2D check operating on this ambiguous geometric prior can include a few background voxels.
This minor trade-off is outweighed by the recall gains on complex thin geometries.

For the 2DGS instantiation, the impact is considerably larger, IoU falls from $0.889$ to $0.597$ on \textit{Trex} and from $0.812$ to $0.544$ on \textit{Fern}, with substantial losses on \textit{Bonsai} ($-0.117$) and \textit{Garden} ($-0.199$). We attribute this difference to a structural property of the two representations.
In NeRF, the volumetric density field populates voxels both at the visible surface and inside the object volume, so the thick-structure path can recover a voxel that lies just behind the center-aligned ray by finding a neighboring density-filled voxel, and the accumulated opacity along any ray is spread across many adjacent samples, so a voxel missed by the strict center check is unlikely to be structurally isolated.
In the 2DGS representation, surfels are anisotropic, randomly oriented, and located \emph{only on the visible surface}, the object interior is hollow.
A voxel grid cell that the thick-structure path fails to vote for, which maps to a cluster of surface surfels with no surrounding filled neighbors to provide a fallback, and missing that voxel means losing the entire cluster of Gaussian primitives it controls.
The thin-structure footprint path, which aggregates evidence over the voxel footprint rather than just the voxel center, provides the only geometric path to recovering such isolated clusters, making it substantially more critical for the 2DGS instantiation.

\textbf{Depth-weighted mask refinement.}
The right half of Table~\ref{tab:ablation_thin_depth} shows that removing the depth-based mask weighting causes a negligible quantitative change but a clear qualitative degradation in both instantiations.
In the NeRF instantiation, zoomed regions exhibit faint ghost artifacts and background voxels are falsely activated as foreground, because denser sampling in zoomed views increases false SAM responses on nearby background pixels.
In the 2DGS instantiation, removing the band-pass depth weighting (Equation~\ref{eq:gs-bandpass}) causes consistent IoU drops of at most $0.004$ on all four scenes. The click-depth band already restricts most background activations through the behind-surface gate in the fusion kernel, and the SAM false positives are marked when the background-projection sits between two thin foreground structures, so depth weighting acts as a complementary soft suppression that cleans up these ambiguities.
Its primary observable effect is qualitative, without it, pixels at depths just outside the band that SAM ambiguously labels foreground survive into the fusion step. Depth weighting is therefore a cheap but qualitatively important refinement in both instantiations, and we keep it enabled in all experiments.

% ---------------------------------------------------------------
\paragraph{Qualitative effect of depth weighting.}

\begin{figure}[h]
    \centering
    \includegraphics[width=0.95\linewidth]{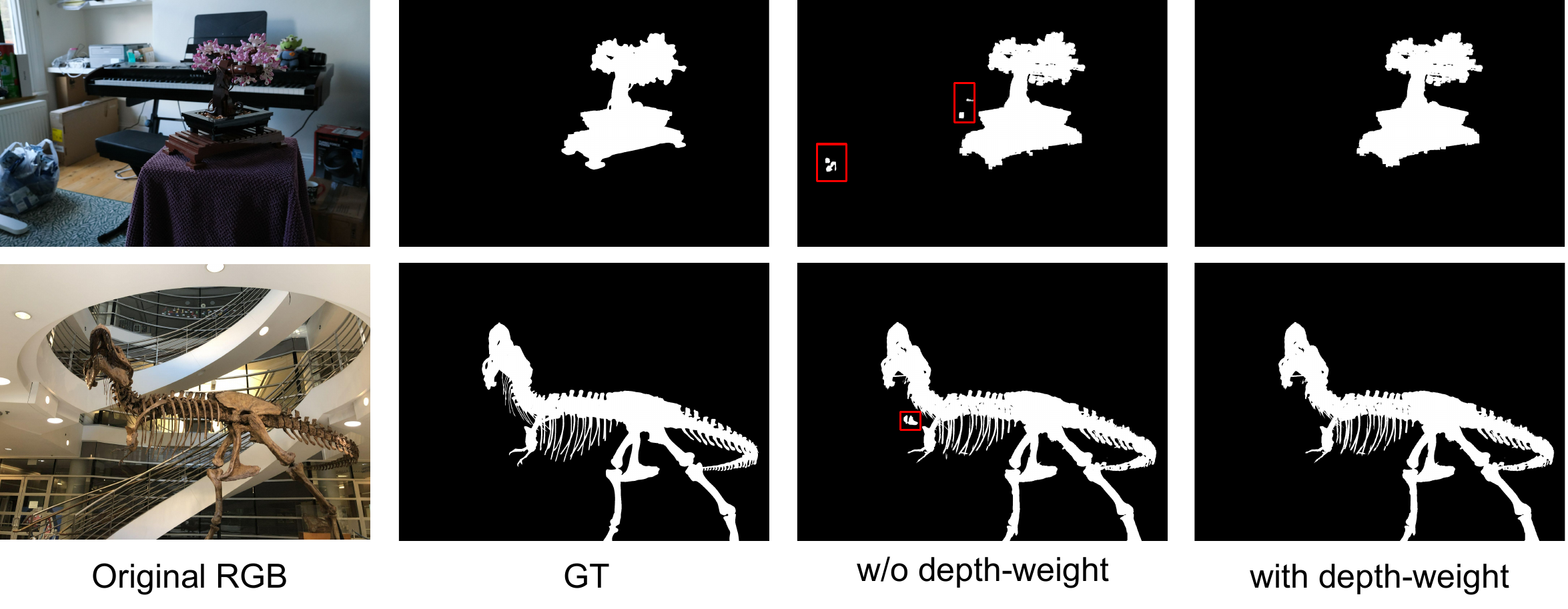}
    \caption{Qualitative ablation of depth weighting in the refined SAM mask. \textbf{Top row:} NeRF instantiation on the \textit{bonsai} scene from Mip-NeRF~$360^\circ$. \textbf{Bottom row:} 2DGS instantiation on the \textit{Trex} scene from LLFF. Red boxes highlight isolated false positives that the depth weight removes.}
    \label{fig:depth_weighting_ablation}
\end{figure}

\textbf{NeRF instantiation (top row).}
As shown in Figure~\ref{fig:depth_weighting_ablation}, without depth weighting, SAM responses on the \textit{bonsai} scene of Mip-NeRF~$360^\circ$ activate background voxels that do not belong to the object. The red boxes highlight isolated false positives. These errors arise because zoomed centroid views sample dense rays near the background, and thus background pixels are projected between the thin foreground structures, such as bonsai flower petals, etc., increasing the spurious high confidence of SAM on these pixels. The NeRF depth map suppresses these activations by penalizing pixels far away in the mask. As a result, the depth-weighted mask remains clean and boundary-aligned, while the non-weighted mask exhibits ghost artifacts at the leaf-background interface.

\noindent\textbf{2DGS instantiation (bottom row).}
The same failure mode appears on \textit{Trex} from LLFF with the 2DGS instantiation, but the underlying mechanism differs slightly. Without the band-pass depth weight of Equation~\ref{eq:gs-bandpass}, SAM ambiguities at the narrow gaps between adjacent rib bones leak onto the staircase that sits directly behind the skeleton. The zoomed centroid frames place the staircase only marginally further back than the bones along the camera ray, and SAM cannot disambiguate them from texture alone. The red boxes highlight these inter-rib bleeds in the non-weighted column. Once the user prompts are translated into a depth band $[d_{\text{min}}^{(s)},d_{\text{max}}^{(s)}]$ around the visible skeleton, the band-pass weight attenuates every pixel whose median depth falls outside the band, eliminating the inter-rib bleeds while leaving the on-bone activations untouched. The qualitative trend exactly mirrors the NeRF behavior in the top row with clean boundaries and no ghost artifacts, despite the underlying weighting being mechanism-specific (AABB normalization depth-weight factor vs. band-pass weight). This consistency across instantiations supports our claim that depth weighting is a representation-agnostic refinement.

% ---------------------------------------------------------------
\paragraph{Needle filter and surfel-footprint coverage (2DGS only).}

\begin{figure}[h]
    \centering
    \includegraphics[width=0.9\linewidth]{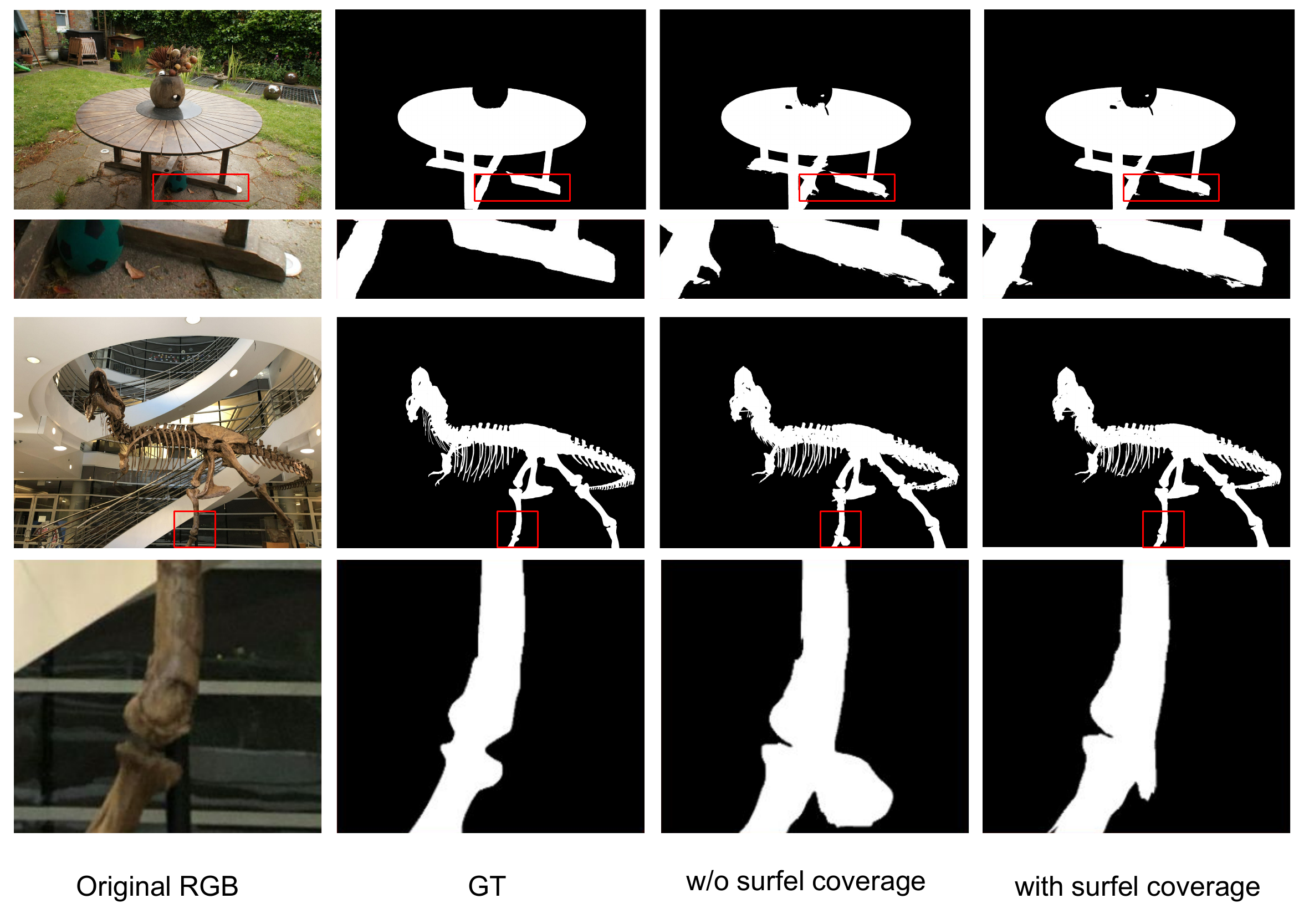}
    \caption{Qualitative effect of the surfel-footprint coverage filter (Pass $3$) in 2DGS instantiation. \textbf{Rows 1-2:} \textit{Garden} (without vase) from Mip-NeRF~$360^\circ$, with row $2$ showing zoomed-in crops of the highlighted regions. \textbf{Rows 3-4 :} \textit{Trex} from LLFF, with row $4$ showing the corresponding zoom-ins.}
    \label{fig:needle_surfel_ablation}
\end{figure}

\begin{table}[!h]
\centering
\small
\begin{tabular}{lcccc}
\toprule
\textbf{Variant} & \textbf{Needle Filter} & \textbf{Coverage Mapping} & \textbf{IoU (\textit{Trex})} & \textbf{mIoU (\textit{Garden})} \\
\midrule
Full (Pass 1+2+3) & \cmark & \cmark & \textbf{0.8891} & \textbf{0.9620} \\
Pass 1+2 only     & \cmark & \xmark & 0.8748          & 0.9482          \\
Pass 1 only       & \xmark & \xmark & 0.8748          & 0.9482          \\
\bottomrule
\end{tabular}
\caption{Ablation of the needle filter (Pass $2$) and the surfel-footprint coverage mapping (Pass $3$) of the 2DGS voxel-to-surfel mapping. Bold marks the best score per scene.}
\label{tab:ablation_needle_surfel}
\end{table}

\noindent\textbf{Needle filter (Pass~2).}
Needle filtering serves as a lightweight geometric prior that removes highly anisotropic surfels prior to coverage analysis. Since the subsequent coverage stage already rejects many boundary artifacts, the quantitative gains are modest, but the filter reduces the number of candidate surfels processed during refinement. Concretely, needle-shaped surfels with high aspect ratio defined in the Equation~\ref{eq:needle} typically have low individual opacity and are visually occluded by the denser, well-rounded surfels around them, so their removal does not change the rendered mask. They are still expensive to process in Pass $3$, because their elongated footprint produces a large screen-space bounding box. Pruning them up-front is therefore a compute optimization.

\noindent\textbf{Surfel-footprint coverage (Pass~3).}
In both scenes shown in Figure~\ref{fig:needle_surfel_ablation}, zoom-ins in rows $2$ and $4$, the without-coverage columns show characteristic background bleeding at object boundaries, disc surfels whose centers legitimately sit inside a segmented voxel but whose 2D footprint mostly covers non-foreground pixels (table leg in \textit{Garden}, bones joint in leg in \textit{Trex}). Pass $3$ measures, per surfel, the fraction of its in-ellipse pixels that the band-pass SAM mask labels foreground (Equation~\ref{eq:kappa}) and culls surfels whose best-view coverage falls below $\tau_\kappa$. The result is a sharper, more topologically accurate boundary on the photo-realistic render. The quantitative gains in Table~\ref{tab:ablation_needle_surfel} ($+0.014$ IoU on \textit{Trex} and $+0.014$ mIoU on \textit{Garden}) align with the qualitative sharpening.

% ================================================================
\section{Conclusion}
We presented DivAS, a training-free, zero-shot framework for interactive $3$D segmentation that operates over both pre-trained NeRF and 2DGS reconstructions through a single shared algorithmic skeleton with representation-specific adapters. By combining geometric priors from 3D representations with zero-shot $2$D SAM masks inside a shared depth-weighted CUDA voxel-fusion kernel, DivAS achieves accurate $3$D object extraction without per-scene optimization or per-primitive feature training on either representation. We discuss limitations and future work in Appendix~\ref{app:limitations}.

A central practical consequence of the optimization-free design is that, among the object-level GS-segmentation methods we evaluate, DivAS-GS is the only one that stays competitive with state-of-the-art segmentation quality while keeping its peak VRAM within the consumer-hardware envelope at the standard $\times 4$ downsampling, where SAGA and Gaussian Grouping do not. SA3D-GS is the only baseline that maintains a comparable footprint but at a substantially lower segmentation quality. DivAS-GS attains the best LLFF IoU and the second-best Mip-NeRF~$360^\circ$ mIoU among the Gaussian splatting baselines, while simultaneously holding the lowest peak VRAM and a near-best end-to-end runtime, a combination that no prior GS-segmentation method achieves under the same resource constraint. The same skeleton applied to NeRF matches or exceeds the optimization-based NeRF baselines at roughly $2\times$ lower end-to-end cost. Taken together, these results suggest that a single optimization-free lifting recipe, shared across backbones through representation-specific adapters, can deliver segmentation quality competitive with state-of-the-art optimization-based methods while operating well within a consumer-hardware memory budget across both NeRF and GS backbones, and we expect this direction to support future tools for interactive, optimization-free $3$D editing and scene understanding on the consumer hardware that its intended audience actually owns.

\bibliography{main}

\begin{thebibliography}{36}
\providecommand{\natexlab}[1]{#1}
\providecommand{\url}[1]{\texttt{#1}}
\expandafter\ifx\csname urlstyle\endcsname\relax
  \providecommand{\doi}[1]{doi: #1}\else
  \providecommand{\doi}{doi: \begingroup \urlstyle{rm}\Url}\fi

\bibitem[Badrinarayanan et~al.(2017)Badrinarayanan, Kendall, and
  Cipolla]{badrinarayanan2017segnet}
Vijay Badrinarayanan, Alex Kendall, and Roberto Cipolla.
\newblock Segnet: A deep convolutional encoder-decoder architecture for image
  segmentation.
\newblock \emph{IEEE transactions on pattern analysis and machine
  intelligence}, 39\penalty0 (12):\penalty0 2481--2495, 2017.

\bibitem[Barron et~al.(2021)Barron, Mildenhall, Tancik, Hedman, Martin-Brualla,
  and Srinivasan]{barron2021mip}
Jonathan~T Barron, Ben Mildenhall, Matthew Tancik, Peter Hedman, Ricardo
  Martin-Brualla, and Pratul~P Srinivasan.
\newblock Mip-nerf: A multiscale representation for anti-aliasing neural
  radiance fields.
\newblock In \emph{Proceedings of the IEEE/CVF international conference on
  computer vision}, pp.\  5855--5864, 2021.

\bibitem[Barron et~al.(2022)Barron, Mildenhall, Verbin, Srinivasan, and
  Hedman]{barron2022mip}
Jonathan~T Barron, Ben Mildenhall, Dor Verbin, Pratul~P Srinivasan, and Peter
  Hedman.
\newblock Mip-nerf 360: Unbounded anti-aliased neural radiance fields.
\newblock In \emph{Proceedings of the IEEE/CVF conference on computer vision
  and pattern recognition}, pp.\  5470--5479, 2022.

\bibitem[Caron et~al.(2021)Caron, Touvron, Misra, J{\'e}gou, Mairal,
  Bojanowski, and Joulin]{caron2021emerging}
Mathilde Caron, Hugo Touvron, Ishan Misra, Herv{\'e} J{\'e}gou, Julien Mairal,
  Piotr Bojanowski, and Armand Joulin.
\newblock Emerging properties in self-supervised vision transformers.
\newblock In \emph{Proceedings of the IEEE/CVF international conference on
  computer vision}, pp.\  9650--9660, 2021.

\bibitem[Cen et~al.(2023)Cen, Zhou, Fang, Shen, Xie, Jiang, Zhang, Tian,
  et~al.]{cen2023segment}
Jiazhong Cen, Zanwei Zhou, Jiemin Fang, Wei Shen, Lingxi Xie, Dongsheng Jiang,
  Xiaopeng Zhang, Qi~Tian, et~al.
\newblock Segment anything in 3d with nerfs.
\newblock \emph{Advances in Neural Information Processing Systems},
  36:\penalty0 25971--25990, 2023.

\bibitem[Cen et~al.(2024)Cen, Fang, Zhou, Yang, Xie, Zhang, Shen, and
  Tian]{cen2024segment3dradiancefields}
Jiazhong Cen, Jiemin Fang, Zanwei Zhou, Chen Yang, Lingxi Xie, Xiaopeng Zhang,
  Wei Shen, and Qi~Tian.
\newblock Segment anything in 3d with radiance fields, 2024.
\newblock URL \url{https://arxiv.org/abs/2304.12308}.

\bibitem[Cen et~al.(2025)Cen, Fang, Yang, Xie, Zhang, Shen, and
  Tian]{cen2025segment}
Jiazhong Cen, Jiemin Fang, Chen Yang, Lingxi Xie, Xiaopeng Zhang, Wei Shen, and
  Qi~Tian.
\newblock Segment any 3d gaussians.
\newblock In \emph{Proceedings of the AAAI conference on artificial
  intelligence}, pp.\  1971--1979, 2025.

\bibitem[Chen et~al.(2022)Chen, Xu, Geiger, Yu, and Su]{chen2022tensorf}
Anpei Chen, Zexiang Xu, Andreas Geiger, Jingyi Yu, and Hao Su.
\newblock Tensorf: Tensorial radiance fields.
\newblock In \emph{European conference on computer vision}, pp.\  333--350.
  Springer, 2022.

\bibitem[Chen et~al.(2017)Chen, Papandreou, Kokkinos, Murphy, and
  Yuille]{chen2017deeplab}
Liang-Chieh Chen, George Papandreou, Iasonas Kokkinos, Kevin Murphy, and Alan~L
  Yuille.
\newblock Deeplab: Semantic image segmentation with deep convolutional nets,
  atrous convolution, and fully connected crfs.
\newblock \emph{IEEE transactions on pattern analysis and machine
  intelligence}, 40\penalty0 (4):\penalty0 834--848, 2017.

\bibitem[Dosovitskiy(2020)]{dosovitskiy2020image}
Alexey Dosovitskiy.
\newblock An image is worth 16x16 words: Transformers for image recognition at
  scale.
\newblock \emph{arXiv preprint arXiv:2010.11929}, 2020.

\bibitem[Fischer et~al.(2026)Fischer, Georgiev, Groueix, Kim, Ritschel, and
  Deschaintre]{fischer2026sama}
Michael Fischer, Iliyan Georgiev, Thibault Groueix, Vladimir~G Kim, Tobias
  Ritschel, and Valentin Deschaintre.
\newblock Sama: Material-aware 3d selection and segmentation.
\newblock In \emph{2026 International Conference on 3D Vision (3DV)}, pp.\
  1812--1822. IEEE, 2026.

\bibitem[Fridovich-Keil et~al.(2022)Fridovich-Keil, Yu, Tancik, Chen, Recht,
  and Kanazawa]{fridovich2022plenoxels}
Sara Fridovich-Keil, Alex Yu, Matthew Tancik, Qinhong Chen, Benjamin Recht, and
  Angjoo Kanazawa.
\newblock Plenoxels: Radiance fields without neural networks.
\newblock In \emph{Proceedings of the IEEE/CVF conference on computer vision
  and pattern recognition}, pp.\  5501--5510, 2022.

\bibitem[Garbin et~al.(2021)Garbin, Kowalski, Johnson, Shotton, and
  Valentin]{garbin2021fastnerf}
Stephan~J Garbin, Marek Kowalski, Matthew Johnson, Jamie Shotton, and Julien
  Valentin.
\newblock Fastnerf: High-fidelity neural rendering at 200fps.
\newblock In \emph{Proceedings of the IEEE/CVF international conference on
  computer vision}, pp.\  14346--14355, 2021.

\bibitem[Goel et~al.(2023)Goel, Sirikonda, Saini, and
  Narayanan]{goel2023interactive}
Rahul Goel, Dhawal Sirikonda, Saurabh Saini, and PJ~Narayanan.
\newblock Interactive segmentation of radiance fields.
\newblock In \emph{Proceedings of the IEEE/CVF conference on computer vision
  and pattern recognition}, pp.\  4201--4211, 2023.

\bibitem[Gonz{\'a}lez(2010)]{gonzalez2010measurement}
{\'A}lvaro Gonz{\'a}lez.
\newblock Measurement of areas on a sphere using fibonacci and
  latitude--longitude lattices.
\newblock \emph{Mathematical geosciences}, 42\penalty0 (1):\penalty0 49--64,
  2010.

\bibitem[Huang et~al.(2024)Huang, Yu, Chen, Geiger, and Gao]{huang20242d}
Binbin Huang, Zehao Yu, Anpei Chen, Andreas Geiger, and Shenghua Gao.
\newblock 2d gaussian splatting for geometrically accurate radiance fields.
\newblock In \emph{ACM SIGGRAPH 2024 conference papers}, pp.\  1--11, 2024.

\bibitem[Kerbl et~al.(2023)Kerbl, Kopanas, Leimk{\"u}hler, Drettakis,
  et~al.]{kerbl20233d}
Bernhard Kerbl, Georgios Kopanas, Thomas Leimk{\"u}hler, George Drettakis,
  et~al.
\newblock 3d gaussian splatting for real-time radiance field rendering.
\newblock \emph{ACM Trans. Graph.}, 42\penalty0 (4):\penalty0 139--1, 2023.

\bibitem[Kerr et~al.(2023)Kerr, Kim, Goldberg, Kanazawa, and
  Tancik]{kerr2023lerf}
Justin Kerr, Chung~Min Kim, Ken Goldberg, Angjoo Kanazawa, and Matthew Tancik.
\newblock Lerf: Language embedded radiance fields.
\newblock In \emph{Proceedings of the IEEE/CVF international conference on
  computer vision}, pp.\  19729--19739, 2023.

\bibitem[Kirillov et~al.(2023)Kirillov, Mintun, Ravi, Mao, Rolland, Gustafson,
  Xiao, Whitehead, Berg, Lo, et~al.]{kirillov2023segment}
Alexander Kirillov, Eric Mintun, Nikhila Ravi, Hanzi Mao, Chloe Rolland, Laura
  Gustafson, Tete Xiao, Spencer Whitehead, Alexander~C Berg, Wan-Yen Lo, et~al.
\newblock Segment anything.
\newblock In \emph{Proceedings of the IEEE/CVF international conference on
  computer vision}, pp.\  4015--4026, 2023.

\bibitem[Kobayashi et~al.(2022)Kobayashi, Matsumoto, and
  Sitzmann]{kobayashi2022decomposing}
Sosuke Kobayashi, Eiichi Matsumoto, and Vincent Sitzmann.
\newblock Decomposing nerf for editing via feature field distillation.
\newblock \emph{Advances in neural information processing systems},
  35:\penalty0 23311--23330, 2022.

\bibitem[Liu et~al.(2024)Liu, Hu, Tang, and Tai]{liu2024sanerf}
Yichen Liu, Benran Hu, Chi-Keung Tang, and Yu-Wing Tai.
\newblock Sanerf-hq: Segment anything for nerf in high quality.
\newblock In \emph{Proceedings of the IEEE/CVF Conference on Computer Vision
  and Pattern Recognition}, pp.\  3216--3226, 2024.

\bibitem[Liu et~al.(2021)Liu, Lin, Cao, Hu, Wei, Zhang, Lin, and
  Guo]{liu2021swin}
Ze~Liu, Yutong Lin, Yue Cao, Han Hu, Yixuan Wei, Zheng Zhang, Stephen Lin, and
  Baining Guo.
\newblock Swin transformer: Hierarchical vision transformer using shifted
  windows.
\newblock In \emph{Proceedings of the IEEE/CVF international conference on
  computer vision}, pp.\  10012--10022, 2021.

\bibitem[Mildenhall et~al.(2019)Mildenhall, Srinivasan, Ortiz-Cayon, Kalantari,
  Ramamoorthi, Ng, and Kar]{mildenhall2019local}
Ben Mildenhall, Pratul~P Srinivasan, Rodrigo Ortiz-Cayon, Nima~Khademi
  Kalantari, Ravi Ramamoorthi, Ren Ng, and Abhishek Kar.
\newblock Local light field fusion: Practical view synthesis with prescriptive
  sampling guidelines.
\newblock \emph{ACM Transactions on Graphics (ToG)}, 38\penalty0 (4):\penalty0
  1--14, 2019.

\bibitem[Mildenhall et~al.(2020)Mildenhall, Srinivasan, Tancik, Barron,
  Ramamoorthi, and Ng]{mildenhall2020nerf}
Ben Mildenhall, Pratul~P Srinivasan, Matthew Tancik, Jonathan~T Barron, Ravi
  Ramamoorthi, and Ren Ng.
\newblock Nerf: Representing scenes as neural radiance fields for view
  synthesis.
\newblock In \emph{European Conference on Computer Vision}, pp.\  405--421.
  Springer, 2020.

\bibitem[M{\"u}ller et~al.(2022)M{\"u}ller, Evans, Schied, and
  Keller]{muller2022instantngp}
Thomas M{\"u}ller, Alex Evans, Christoph Schied, and Alexander Keller.
\newblock Instant neural graphics primitives with a multiresolution hash
  encoding.
\newblock In \emph{ACM SIGGRAPH}, 2022.

\bibitem[Qin et~al.(2024)Qin, Li, Zhou, Wang, and Pfister]{qin2024langsplat}
Minghan Qin, Wanhua Li, Jiawei Zhou, Haoqian Wang, and Hanspeter Pfister.
\newblock Langsplat: 3d language gaussian splatting.
\newblock In \emph{Proceedings of the IEEE/CVF Conference on Computer Vision
  and Pattern Recognition}, pp.\  20051--20060, 2024.

\bibitem[Radford et~al.(2021)Radford, Kim, Hallacy, Ramesh, Goh, Agarwal,
  Sastry, Askell, Mishkin, Clark, et~al.]{radford2021learning}
Alec Radford, Jong~Wook Kim, Chris Hallacy, Aditya Ramesh, Gabriel Goh,
  Sandhini Agarwal, Girish Sastry, Amanda Askell, Pamela Mishkin, Jack Clark,
  et~al.
\newblock Learning transferable visual models from natural language
  supervision.
\newblock In \emph{International conference on machine learning}, pp.\
  8748--8763. PmLR, 2021.

\bibitem[Ren et~al.(2022)Ren, Agarwala, Russell, Schwing, and
  Wang]{ren2022neural}
Zhongzheng Ren, Aseem Agarwala, Bryan Russell, Alexander~G Schwing, and Oliver
  Wang.
\newblock Neural volumetric object selection.
\newblock In \emph{Proceedings of the IEEE/CVF conference on computer vision
  and pattern recognition}, pp.\  6133--6142, 2022.

\bibitem[Strudel et~al.(2021)Strudel, Garcia, Laptev, and
  Schmid]{strudel2021segmenter}
Robin Strudel, Ricardo Garcia, Ivan Laptev, and Cordelia Schmid.
\newblock Segmenter: Transformer for semantic segmentation.
\newblock In \emph{Proceedings of the IEEE/CVF international conference on
  computer vision}, pp.\  7262--7272, 2021.

\bibitem[Tschernezki et~al.(2022)Tschernezki, Laina, Larlus, and
  Vedaldi]{tschernezki2022neural}
Vadim Tschernezki, Iro Laina, Diane Larlus, and Andrea Vedaldi.
\newblock Neural feature fusion fields: 3d distillation of self-supervised 2d
  image representations.
\newblock In \emph{2022 International Conference on 3D Vision (3DV)}, pp.\
  443--453. IEEE, 2022.

\bibitem[Vora et~al.(2021)Vora, Radwan, Greff, Meyer, Genova, Sajjadi, Pot,
  Tagliasacchi, and Duckworth]{vora2021nesf}
Suhani Vora, Noha Radwan, Klaus Greff, Henning Meyer, Kyle Genova, Mehdi~SM
  Sajjadi, Etienne Pot, Andrea Tagliasacchi, and Daniel Duckworth.
\newblock Nesf: Neural semantic fields for generalizable semantic segmentation
  of 3d scenes.
\newblock \emph{arXiv preprint arXiv:2111.13260}, 2021.

\bibitem[Xie et~al.(2021)Xie, Wang, Yu, Anandkumar, Alvarez, and
  Luo]{xie2021segformer}
Enze Xie, Wenhai Wang, Zhiding Yu, Anima Anandkumar, Jose~M Alvarez, and Ping
  Luo.
\newblock Segformer: Simple and efficient design for semantic segmentation with
  transformers.
\newblock \emph{Advances in neural information processing systems},
  34:\penalty0 12077--12090, 2021.

\bibitem[Xiong et~al.(2024)Xiong, Varadarajan, Wu, Xiang, Xiao, Zhu, Dai, Wang,
  Sun, Iandola, et~al.]{xiong2024efficientsam}
Yunyang Xiong, Bala Varadarajan, Lemeng Wu, Xiaoyu Xiang, Fanyi Xiao, Chenchen
  Zhu, Xiaoliang Dai, Dilin Wang, Fei Sun, Forrest Iandola, et~al.
\newblock Efficientsam: Leveraged masked image pretraining for efficient
  segment anything.
\newblock In \emph{Proceedings of the IEEE/CVF Conference on Computer Vision
  and Pattern Recognition}, pp.\  16111--16121, 2024.

\bibitem[Ye et~al.(2024)Ye, Danelljan, Yu, and Ke]{ye2024gaussian}
Mingqiao Ye, Martin Danelljan, Fisher Yu, and Lei Ke.
\newblock Gaussian grouping: Segment and edit anything in 3d scenes.
\newblock In \emph{European conference on computer vision}, pp.\  162--179.
  Springer, 2024.

\bibitem[Zhang et~al.(2023)Zhang, Han, Qiao, Kim, Bae, Lee, and
  Hong]{zhang2023faster}
Chaoning Zhang, Dongshen Han, Yu~Qiao, Jung~Uk Kim, Sung-Ho Bae, Seungkyu Lee,
  and Choong~Seon Hong.
\newblock Faster segment anything: Towards lightweight sam for mobile
  applications.
\newblock \emph{arXiv preprint arXiv:2306.14289}, 2023.

\bibitem[Zhi et~al.(2021)Zhi, Laidlow, Leutenegger, and Davison]{zhi2021place}
Shuaifeng Zhi, Tristan Laidlow, Stefan Leutenegger, and Andrew~J Davison.
\newblock In-place scene labelling and understanding with implicit scene
  representation.
\newblock In \emph{Proceedings of the IEEE/CVF International Conference on
  Computer Vision}, pp.\  15838--15847, 2021.

\end{thebibliography}
\bibliographystyle{tmlr}

\appendix
\appendix
\begin{center}
{\Large\textbf{Supplementary Material for:}}\\[0.4em]
{\Large\textbf{DivAS: Interactive 3D Segmentation by Depth-Weighted Voxel Aggregation}}
\end{center}
\vspace{1em}

% ================================================================
\section{Implementation Details}
\label{app:implementation}

\subsection{NeRF instantiation}
\label{app:impl-nerf}

\begin{figure}[h!]
\centering
\includegraphics[width=\linewidth]{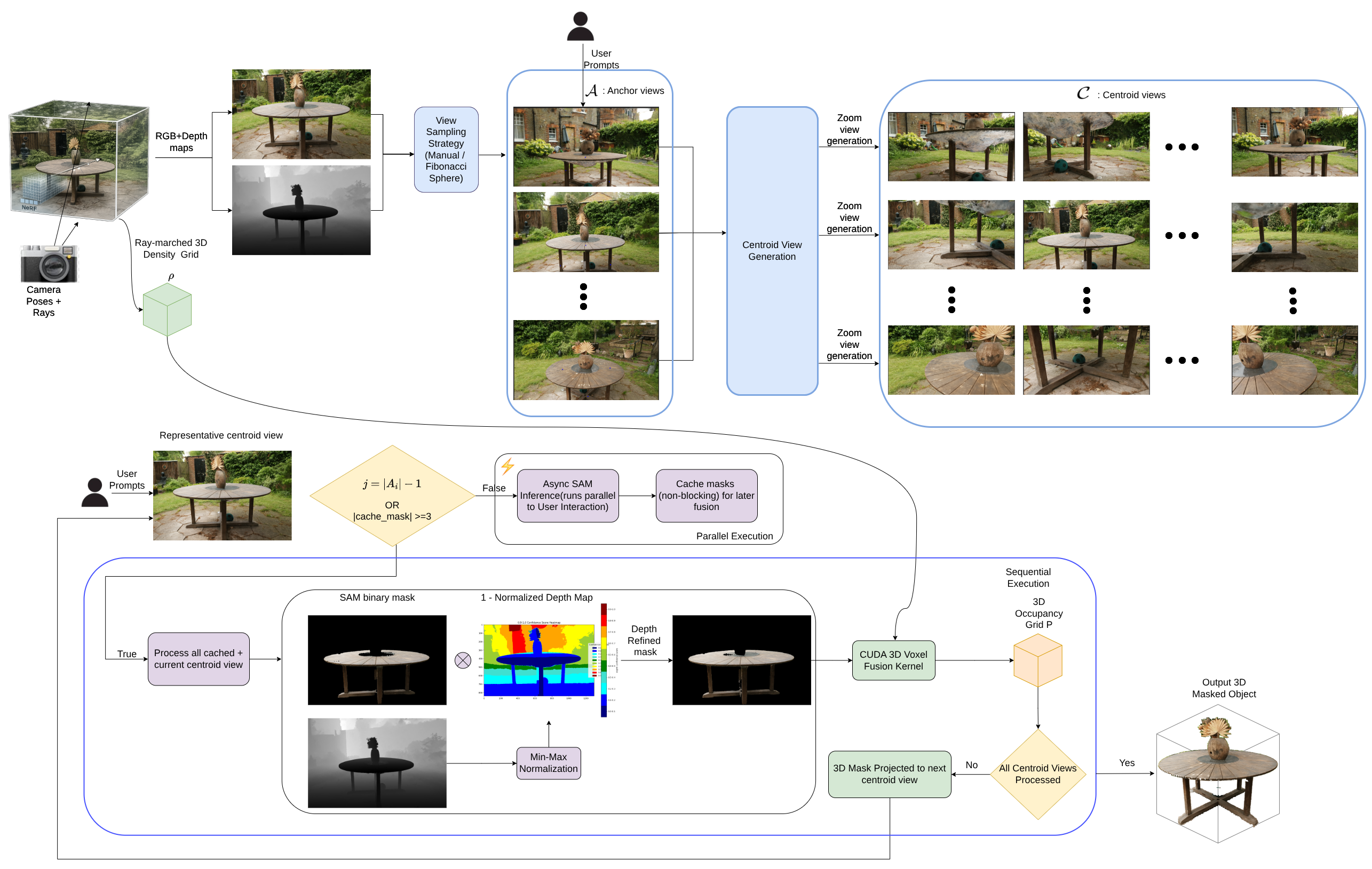}
\caption{\textbf{NeRF instantiation of DivAS.} RGB and depth are rendered through volumetric ray marching over the hash-grid NeRF representation. The occupancy grid is obtained by ray-marching the trained density field. For each centroid view $c_j$ in an anchor group $\mathcal{A}_i$, SAM inference runs asynchronously while the user continues interacting with subsequent views. When the cache holds at least three processed masks or the group ends, all cached and current masks are synchronized, refined using inverse-normalized NeRF depth maps, and fused by a CUDA-based $3$D voxel kernel into a probabilistic occupancy grid $\mathcal{P}$. The fused 3D mask is iteratively projected back onto subsequent centroid views to guide progressive scene-wide refinement. }
\label{fig:nerf_arch}
\end{figure}

\paragraph{NeRF training setup.}
\label{para:nerf-training-setup}
For LLFF, we use the original \texttt{torch-ngp} implementation with uniform sampling. For Mip-NeRF~$360^\circ$, we use the hierarchical proposal sampling from SANeRF-HQ~\citep{liu2024sanerf}. Following Instant-NGP~\citep{muller2022instantngp}, the scene-bound parameter is selected per scene as the smallest value that still covers the full scene while maintaining clear depth separation between foreground and background.

SANeRF-HQ~\citep{liu2024sanerf} uses a $16$-level hash grid with $2$-dimensional features for its NeRF field, and an additional $16$-level grid with feature dimension $8$ for its SAM feature field. Our LLFF configuration follows the Instant-NGP default, a faster $8$-level hash grid with $4$-dimensional features, which is well-suited to the forward-facing capture geometry of LLFF. For Mip-NeRF~$360^\circ$, we match SANeRF-HQ's $16$-level hash grid and $2$-dimensional feature configuration to handle the wider depth range of the unbounded scenes. These differences affect only the underlying NeRF reconstruction and do not change the DivAS segmentation pipeline.

The per-scene bound values used in our NeRF reconstructions are summarized in Table~\ref{tab:nerf_bound}.

\begin{table}[H]
\centering
\small
\begin{tabular}{l c}
\toprule
Scene & Bound (\texttt{aabb\_scale}) \\
\midrule
Fern     & $2$ \\
Fortress & $4$ \\
Orchids  & $4$ \\
Horns    & $8$ \\
Trex     & $32$ \\
\bottomrule
\end{tabular}
\caption{Bound parameters used for NeRF training. These depend only on the underlying NeRF reconstruction and are not part of the DivAS method itself.}
\label{tab:nerf_bound}
\end{table}

\paragraph{Voxel grids.}
We use a $128^3$ occupancy grid for LLFF and a $256^3$ grid for Mip-NeRF~$360^\circ$, balancing ray-marching speed with the resolution needed to avoid boundary leakage. Two $3$D grids are maintained for the NeRF instantiation, one for the NeRF density field used by the geometric consistency checks, and one for the voxel-wise segmentation probability produced by the fusion kernel. These grid resolutions were kept fixed across all NeRF experiments.

\paragraph{Training schedule.}
The underlying NeRF is trained for $25\text{K}$ iterations on LLFF and $20\text{K}$ iterations on Mip-NeRF~$360^\circ$, using images downsampled by $\times 4$ throughout. The same schedule is used consistently across all reported NeRF experiments.

\paragraph{CUDA fusion-kernel hyperparameters.}
The fusion kernel uses dataset-specific values for the tolerance and density hyperparameters in the upper block of Table~\ref{tab:kernel_hyperparams}, while the remaining entries are held fixed across both benchmarks. The tuned hyperparameters are specified per reconstruction pipeline rather than per scene. As mentioned at ~\ref{para:nerf-training-setup}, LLFF uses uniform ray sampling, whereas Mip-NeRF~$360$ employs the hierarchical proposal-sampling strategy. These reconstruction pipelines produce substantially different depth distributions, density accumulation profiles, and ray-sample statistics. Consequently, parameters such as $\gamma$ and $\lambda_{\text{range}}$ primarily serve as normalization factors that adapt the fusion kernel to the statistical scale of the underlying reconstruction pipeline, rather than altering the segmentation logic itself. These values were selected through small sweeps reported in Section~\ref{app:additional-ablations}. Once selected, a single parameter set is used for all scenes within a benchmark without further per-scene or per-object tuning. Importantly, these parameters are fixed before interactive segmentation begins and are not adjusted during user interaction. The user is therefore exposed to a single operating configuration for a given reconstruction pipeline.

\begin{table}
\centering
\scriptsize
\begin{tabular}{l c c p{0.4\linewidth}}
\toprule
\textbf{Parameter} & \textbf{LLFF} & \textbf{Mip-NeRF~$360^\circ$} & \textbf{Purpose} \\
\midrule
\texttt{base\_tolerance\_multiplier} $\gamma$ & $4.0$ & $0.5$ & Base scale for the depth tolerance used in geometric checks. \\
\texttt{per\_sample\_bonus} $\beta$ & $0.2$ & $0.2$ & Incremental relaxation added per supporting depth sample. \\
\texttt{max\_bonus} $b_{\max}$ & $10$ & $1$ & Upper bound on the accumulated tolerance bonus. \\
\texttt{depth\_range\_factor} $\lambda_\text{range}$ & $0.05$ & $5.0$ & Used to control the distance tolerance check. \\
\texttt{density\_thresh} $\rho_\text{thresh}$ & $10$ & $5$ & Density required for a voxel to be considered valid. \\
\texttt{thin\_density\_thresh} $\rho_\text{thin\_thresh}$ & $50$ & $30$ & Higher density threshold used to detect thin structures. \\
\texttt{thin\_percent\_cover} $\rho_\text{percent\_cover\_thresh}$ & $0.20$ & $0.30$ & Fractional coverage of voxel by the masked pixels.\\
\midrule
\texttt{grid\_resolution} & $128^3$ & $256^3$ & Occupancy and segmentation voxel-grid resolution. \\
\texttt{cumulative\_weight} $\tau_\text{cw}$ & $0.75$ & $0.75$ & Cumulative-weight cutoff fixing the upper depth bound $\mD_\text{max}$. \\
\texttt{thin\_accept} $\tau_\text{thin}$ & $0.6$ & $0.6$ & Thin-structure acceptance threshold in compact fusion. \\
\texttt{consensus\_threshold} $\tau_\text{cons}$ & $0.5$ & $0.5$ & Multi-view consensus threshold marking a voxel segmented. \\
\bottomrule
\end{tabular}
\caption{NeRF instantiation hyperparameters and the values used in all reported runs. The upper block holds the tolerance and density parameters that are tuned per reconstruction pipeline (LLFF versus Mip-NeRF~$360^\circ$), and the lower block holds the structural and threshold constants, which are held fixed across both benchmarks apart from the voxel-grid resolution. Sensitivity scans for the tunable parameters are provided in Section~\ref{app:additional-ablations}.}
\label{tab:kernel_hyperparams}
\end{table}

\subsection{NeRF-Backbone 3D Fusion Details}
\label{app:nerf-fusion}

\paragraph{Occupancy from ray-marched density.}
For the NeRF backbone, the trained scene is represented as a continuous volumetric density field. During rendering, each pixel is obtained by ray marching through the scene and accumulating density contributions along the camera ray. Besides the rendered RGB image, the renderer provides a depth tuple $(\mD_\text{min}, \mD_\text{ref}, \mD_\text{max}, \mD_\text{mw}, \mN_\text{samp})$ mentioned at Section.~\ref{para:prelim} for every pixel. These quantities capture the geometric structure observed along the ray and serve as the primary inputs to the 3D fusion kernel.

\paragraph{Voxel-to-view projection and geometric checks.}
Each voxel center $\vx_c$ projects to normalized image coordinates $(u_x^i,v_x^i)=\pi_i(\vx_c)$, out-of-frame voxels skip view $i$, otherwise the SAM probability $m_x^i$ is sampled at the pixel. A voxel passing the mask threshold ($m_x^i\ge 0.5$) and the density floor ($\boldsymbol{\rho}_x\ge\rho_\text{thresh}$) then undergoes two geometric checks. Let $d_\text{min}, d_\text{exp}, d_\text{max}, d_\text{mw}, n_\text{samp}$ be the values from the depth buffer tuple for the pixel $x_p, y_p$.
\textbf{Ray-voxel alignment} We enforce geometric consistency by measuring the distance between the voxel center and the image ray corresponding to the pixel $(u_x^i,v_x^i)$. The ray origin and direction $(\vo_i,\hat{\vd}_i)$ are derived through $\mK_i, \mT_i$, which are the intrinsic and extrinsic parameters of the camera related to the pixel. The voxel to ray depth is computed as the projection $t_\text{proj}=\langle\vx_c-\vo_i,\hat{\vd_i}\rangle$, clamps it to the NeRF surface band $t_c=\text{clamp}(t_\text{proj},d_\text{min},d_\text{max})$ which ensures that the voxel lies within the ray segment corresponding to the rendered surface. The closest point on the segment of the ray is then $\vp_c = \vo_i + \hat{\vd_i} t_c$ mapped through scene contraction for unbounded scenes. The geometric deviation used for consistency testing is: 
\begin{equation}
\label{eq:distance_cond}
    \delta_x^i = \|\mathbf{x}_c - \mathbf{p}_c||.
\end{equation}
This is the value on which the distance tolerance criterion is applied for geometric consistency.

\textbf{Depth-gradient factor} The spatial tolerance is modulated using a depth-gradient term $g_p^i=\pi_{\text{depth\_gradient}}(x_p,y_p,\mD_{\min},\mD_{\max})$, relaxing it on flat surfaces and tightening it at depth discontinuities.

\textbf{Distance tolerance}: We define a spatial proximity threshold between a voxel and its corresponding ray as:
\begin{equation}
\label{eq:dist_tol}
\tau_\text{spatial}=\Delta_x\,g_p^i + \lambda_\text{range}(d_\text{max}-d_\text{min}).
\end{equation}
where $\Delta_x$ is the cascade-level voxel edge defined earlier and the multi-resolution grid follows~\citep{muller2022instantngp}, and $\lambda_{\text{range}}$ is a scene-dependent hyperparameter. This adaptively increases the tolerance in smooth regions and decreases it at depth edges. The second term, $\lambda_\text{range}(d_\text{max}-d_\text{min})$, adds slack in proportion to the width of the rendered depth band $[d_\text{min},d_\text{max}]$, which is wide for NeRF because the volumetric density fills the solid interior of the object. A voxel lying anywhere between the first surface hit and the back of the object is therefore legitimately consistent with the rendered surface, and this term admits such interior voxels. The 2DGS instantiation omits it, since its surfels lie only on the visible surface and report a precise single-surface depth with no interior band to absorb.

\textbf{Depth tolerance}
To prevent background leakage, we impose a depth consistency check on each voxel. Let $x_d$ denote the voxel depth in this current camera view. The depth tolerance adapts to the NeRF sampling density $n_{\text{samp}}$.
\begin{equation}
\label{eq:depth_tol}
    \tau_{\text{depth}} = (\gamma + b)\Delta_x,\; \text{where}\; b = \min(\beta \cdot n_{\text{samp}},b_{\max}) 
\end{equation}

with $\beta, b_{\max}, \gamma$ are hyperparameters.
A voxel $x$ is considered geometrically consistent if
\begin{equation}
\label{eq:nerf-accept}
    \delta_x^i \leq \tau_{\text{spatial}} \; \land \; \|x_d - d_{\text{exp}}\| \leq \tau_{\text{depth}}
\end{equation}
Voxels satisfying both conditions follow the \textit{thick-structure} path for depth-weighted score aggregation, while others are handled by the thin-structure refinement step.

\paragraph{Depth-weighted vote aggregation.}
Following the volumetric-rendering property of NeRF, along each camera ray, most of the density concentrates near the surface intersection, while contributions from distant samples fade. Hence, voxels projected to the expected surface center should have a stronger influence than those at the depth boundaries. The accepted votes are weighted by a Gaussian falloff from the segment midpoint $\mu_d=\tfrac12(d_\text{min}+d_\text{max})$ with half-width $h_d=\max(\tfrac12(d_\text{max}-d_\text{min}),\epsilon)$ and normalized offset $r=|t_c-\mu_d|/h_d$:
\begin{equation}
w_\text{depth}=\exp(-\alpha_1 r^2),
\qquad
p_x=\frac{\sum_i m_x^i\,w_\text{depth}}{\sum_i w_\text{depth}}.
\label{eq:nerf-vote}
\end{equation}
where $\alpha_1$ is a fixed falloff constant controlling how quickly confidence decays with depth distance. Voxels near the range midpoint $(r \approx 0)$ thus receive higher weights, while those farther away are smoothly down-weighted, approximating a Gaussian attenuation around the true surface. Because the weights are normalized, $\alpha_1$ affects only the relative sharpness and not the scale of fusion.

\paragraph{Thin-structure path: motivation.}
The thick path of Equation~\ref{eq:nerf-accept} tests consistency at a single projected pixel at the voxel center. This center-ray assumption breaks down for thin, high-density structures (Trex ribs, wires, etc.) whose entire $3$D extent projects to a handful of image pixels. In these cases, the voxel center frequently lands \emph{between} the few mask-active pixels or at the background, so the SAM probability sampled at the center is below $0.5$, and the thick gate silently rejects an otherwise valid voxel. Recovering these structures requires \emph{footprint-based} reasoning, in which evidence is aggregated over the voxel's full $2$D projected extent rather than at a single pixel, and thus we need a thin-structure path.

\paragraph{Three-criterion eligibility.}
To avoid spuriously activating background voxels, the thin path is run only when all three conditions hold for the voxel $x$: (i) the view is sufficiently \emph{zoomed in} so the projected footprint is sufficiently large to be locally probed, already satisfied by the centroid view (ii) the SAM mask at the
projected pixel is non-trivial, $m_x^i > 0.1$, ruling out a clear background and (iii) the NeRF-predicted density at the voxel exceeds the higher thin-structure threshold, $\rho_x \ge \rho_\text{thin\_thresh}$, restricting evidence to geometrically plausible fine structures.

\paragraph{Voxel footprint construction.}
Given voxel center $\mathbf{x}_c$ and edge $\Delta_x$, we form the eight corners
$
\mathbf{x}_j^{i} = \mathbf{x}_c + \tfrac{\Delta_x}{2}\,\boldsymbol{\epsilon}_j,
\qquad \boldsymbol{\epsilon}_j \in \{-1,+1\}^3,\quad j = 1,\dots,8,
$
Project each corner of the voxel into view $i$ using its camera parameters and take the axis-aligned bounding box normalized coordinates converted to discrete pixel indices using the spatial resolution of the view given as $\sB_x^i = \{(x_p,y_p) | x_\text{start} \leq x_p \leq x_\text{end}, y_\text{start} \leq y_p \leq y_\text{end}\}$. This defines the footprint of total pixel count $N_\text{pixels} = (x_\text{end}-x_\text{start}+1) \times (y_\text{end}-y_\text{start}+1)$. This is the discrete $2$D region over which mask and depth evidence is aggregated for voxel $x$ in view $i$.

\paragraph{Per-pixel depth gate and supportive-pixel counting.}
Let $x_d$ be the voxel depth along the camera-forward axis in view $i$. For each footprint pixel $(x_p, y_p)\in\sB_x^i$, the kernel reads the mask $m = \mM_i(x_p, y_p)$. Let $d_\text{min}, d_\text{exp}, d_\text{max}, d_\text{mw}, n_\text{samp}$ be the values from the depth buffer tuple for the pixel $(x_p, y_p)$. It then evaluates a footprint-local depth tolerance that reuses the per-sample bonus of Equation~\ref{eq:depth_tol} but with a \emph{doubled} base multiplier to relax the depth tolerance because of the wide depth band learned by the NeRF for the thin structures, reflecting the larger effective scale of the bounding-box test compared with the single-pixel thick gate:
\begin{equation}
\tau_\text{depth}^\text{thin}(x_p,y_p) = (2\gamma + b)\,\Delta_x
\qquad \text{where, }
b = \min\bigl(\beta\,n_\text{samp},\,b_\text{max}\bigr)
\label{eq:nerf-thindepth}
\end{equation}
A pixel is declared \emph{supportive} when it is simultaneously claimed by SAM and depth-consistent with the voxel:
\begin{equation}
\text{support}(x_p, y_p) \;=\;
\mathbbm{1}\!\bigl[\,m > 0.5 \;\wedge\; \|\,x_d - d_\text{exp}\,\| \le \tau_\text{depth}^\text{thin}(x_p,y_p)\bigr].
\label{eq:nerf-support}
\end{equation}
Two scalars are accumulated as the kernel sweeps the bounding box. The support count and the maximum mask value seen anywhere in the footprint,
\begin{equation}
n_\text{valid} \;=\; \!\!\!\!\sum_{(x_p,y_p)\in\sB_x^i}\!\!\!\!\text{support}(x_p,y_p),
\qquad
m_\text{max} \;=\; \max_{(x_p,y_p)\in\sB_x^i}\, \mM_i(x_p, y_p).
\label{eq:nerf-counters}
\end{equation}

\paragraph{Coverage ratio and partial-credit confidence.}
The coverage ratio is the supportive fraction of the footprint,
\begin{equation}
p_\text{covered} \;=\; n_\text{valid}\,/\,N_\text{pixels}
\label{eq:nerf-coverage}
\end{equation}
and the thin-structure confidence for voxel $x$ in view $i$ is assigned by a hard coverage gate that returns the peak mask activation in the footprint only
when the supportive fraction meets the threshold:
\begin{equation}
t_x^{\,i} \;=\;
\begin{cases}
m_\text{max}, & p_\text{covered} \ge \rho_\text{cover}, \\
0, & \text{otherwise.}
\end{cases}
\label{eq:nerf-thin}
\end{equation}
Returning $m_\text{max}$ rather than $p_\text{covered}$ when the gate fires gives the voxel partial credit at the strongest evidence the footprint contains, while a sub-threshold footprint contributes nothing, leaving the denominator of Equation~\ref{eq:compact-fusion} unchanged.

\paragraph{Integration with compact fusion.}
The thin path enters the compact fusion of Equation~\ref{eq:compact-fusion} through the thin-view set $\sT_v$. A view $i$ joins $\sT_v$ when $t_x^{\,i} \ge \tau_\text{thin}$, contributing $t_x^{\,i}$ to the numerator and $1$ to the denominator. Because the thick path is exclusive, the same $(x,i)$ pair does not contribute through both paths. Thin evidence \emph{augments} rather than dilutes the thick aggregation, recovering thin structures that the center-ray test would otherwise drop.
The complete pseudocode of the 3D voxel fusion kernel is defined in Algorithm~\ref{alg:nerf-fusion}, which carries out the thick-structure segmentation via center-ray depth-weighted aggregation, and Algorithm~\ref{alg:nerf-thin} carries out the thin-structure recovery via footprint-based coverage analysis. Their outputs are combined per voxel by the compact-fusion rule of Equation~\ref{eq:compact-fusion}.

\begin{algorithm}
\caption{Thick-Structure Segmentation via Depth-Weighted Voxel Aggregation}
\label{alg:nerf-fusion}
\begin{algorithmic}
\Require voxel grid $\sS$, refined masks $\{\widetilde{\mM}_i\}_{i=1}^{S}$ with associated depth maps $(\{\mD^{(s)}_\text{min}\}_{s=1}^S, \{\mD^{(s)}_\text{exp}\}_{s=1}^S, \{\mD^{(s)}_\text{max}\}_{s=1}^S, \{\mN^{(s)}_\text{samp}\}_{s=1}^S)$, density grid $\boldsymbol{\rho}$, 3D occupancy grid $\tP$, cameras $\{\mK_i,\mT_i\}$, hyperparameters $(\gamma,\beta,b_\text{max},\lambda_\text{range},\rho_\text{thresh},\rho_\text{thin\,thresh},\alpha_1,\tau_\text{thin})$

\For{each voxel $x \in \sS$ \textbf{in parallel}}
   \State retrieve voxel center $\vx_c$ and edge $\Delta_x$
   \If{$\boldsymbol{\rho}_x < \rho_\text{thresh}$}
       \State $P^{\text{thick}}(x) \gets 0$;\quad \textbf{continue} \Comment{density-based empty-voxel skip}
   \EndIf
   \State $W_x \gets 0,\ Z_x \gets 0$
   \For{each view $i = 1,\dots,S$}
      \State project voxel center: $(u_x^i, v_x^i) \gets \pi_i(\vx_c)$;\quad \textbf{if} out-of-frame \textbf{continue}
      \State sample $m_x^i \gets \widetilde{\mM}_i(u_x^i, v_x^i)$ and obtain voxel depth $x_d$ in view $i$
      \State $\text{check\_thin} \gets \textbf{True}$ \Comment{flag for thin-path fallback eligibility}
      \If{$m_x^i \ge 0.5$} \Comment{thick path entry: center-pixel mask gate}
         \State $d_\text{min}, d_\text{exp}, d_\text{max}, n_\text{samp} \gets (\mD^{(i)}_\text{min}, \mD^{(i)}_\text{exp}, \mD^{(i)}_\text{max}, \mN^{(i)}_\text{samp})_{u^i_x,v^i_x}$
         \State compute closest point on the camera ray and the ray-voxel distance $\delta_x^i$ \Comment{Equation~\ref{eq:distance_cond}}
         \State $g_p^i \gets \pi_\text{depth\_gradient}(u_x^i, v_x^i, d_\text{min}, d_\text{max})$
         \State $\tau_\text{spatial} \gets \Delta_x\,g_p^i + \lambda_\text{range}(d_\text{max} - d_\text{min})$ \Comment{Equation~\ref{eq:dist_tol}}
         \State $b \gets \min(\beta\,n_\text{samp},\,b_\text{max})$, \quad $\tau_\text{depth} \gets (\gamma + b)\,\Delta_x$ \Comment{Equation~\ref{eq:depth_tol}}
         \If{$\delta_x^i \le \tau_\text{spatial}$ \textbf{and} $\|x_d - d_\text{exp}\| \le \tau_\text{depth}$} \Comment{Equation~\ref{eq:nerf-accept}}
            \State $\mu_d \gets \tfrac12(d_\text{min}+d_\text{max})$, $h_d \gets \max(\tfrac12(d_\text{max}-d_\text{min}),\epsilon)$
            \State $r \gets |t_c-\mu_d|/h_d$
            \State $w_\text{depth}=\exp(-\alpha_1 r^2)$          \Comment{Equation~\ref{eq:nerf-vote}}
            \State $W_x \mathrel{+}= m_x^i\,w_\text{depth}$;\quad $Z_x \mathrel{+}= w_\text{depth}$ \Comment{depth-weighted center-ray vote}
            \State $\text{check\_thin} \gets \textbf{False}$ \Comment{thick vote accepted, skip thin fallback}
         \EndIf
      \EndIf
      \State \emph{// Thin-structure fallback: invoke Algorithm~\ref{alg:nerf-thin} when the thick path is skipped or rejected}
      \If{$\text{check\_thin}$ \textbf{and} $m_x^i > 0.1$ \textbf{and} $\boldsymbol{\rho}_x \ge \rho_\text{thin\,thresh}$}
         \State $t_x^{\,i} \gets \textsc{ThinStructure}(x, i)$ \Comment{call Algorithm~\ref{alg:nerf-thin}}
         \If{$t_x^{\,i} \ge \tau_\text{thin}$}
            \State $W_x \mathrel{+}= t_x^{\,i}$;\quad $Z_x \mathrel{+}= 1$ \Comment{thin-pathway contribution}
         \EndIf
      \EndIf
   \EndFor
   \State \emph{// Multi-view consensus}
   \If{$Z_x > \varepsilon$}
      \State $\tP_x \gets W_x / Z_x$
   \Else
      \State $\tP_x \gets 0$
   \EndIf
\EndFor
\State \Return $\tP$
\end{algorithmic}
\end{algorithm}

\begin{algorithm}
\caption{Thin-Structure Segmentation via Footprint-Based Coverage Analysis}
\label{alg:nerf-thin}
\begin{algorithmic}
\Require voxel center $\vx_c$, edge $\Delta_x$, voxel depth $x_d$, depth maps $(\{\mD^{(s)}_\text{min}\}_{s=1}^S, \{\mD^{(s)}_\text{exp}\}_{s=1}^S, \{\mD^{(s)}_\text{max}\}_{s=1}^S, \{\mN^{(s)}_\text{samp}\}_{s=1}^S)$, refined mask $\widetilde{\mM}_i$ of view $i$, camera $(\mK_i,\mT_i)$; hyperparameters $(\gamma,\beta,b_\text{max},\rho_\text{cover})$
\State Form the $8$ voxel corners $\{\vx_j^i = \vx_c + \tfrac{\Delta_x}{2}\boldsymbol{\epsilon}_j\}_{j=1}^{8}$, $\boldsymbol{\epsilon}_j \in \{-1,+1\}^3$ 
\State $(u_j^x, v_j^x) \gets \pi_i(\vx_j^i)$ Project corners.  
\State Construct the axis-aligned bounding box $\sB_x^i = [u_\text{min},u_\text{max}] \times [v_\text{min},v_\text{max}]$ in pixel coordinates, clipped to the image
\State $N_\text{pixels} \gets \|\sB_x^i\|$;\quad $n_\text{valid} \gets 0$;\quad $m_\text{max} \gets 0$
\For{each pixel $(x_p, y_p) \in \sB_x^i$}
   \State $m \gets \widetilde{\mM}_i(x_p, y_p)$
   \State $d_\text{min}, d_\text{exp}, d_\text{max}, n_\text{samp} \gets (\mD^{(i)}_\text{min}, \mD^{(i)}_\text{exp}, \mD^{(i)}_\text{max}, \mN^{(i)}_\text{samp})_{x_p,y_p}$
   \State $b \gets \min(\beta\,n_\text{samp},\,b_\text{max})$, \quad $\tau_\text{depth}^{\text{thin}} \gets (2\gamma + b)\,\Delta_x$ \Comment{Equation~\ref{eq:nerf-thindepth}}
   \If{$m > 0.5$ \textbf{and} $\|x_d - d_\text{exp}(x_p, y_p)\| \le \tau_\text{depth}^{\text{thin}}$} \Comment{Equation~\ref{eq:nerf-support}}
      \State $n_\text{valid} \mathrel{+}= 1$ 
   \EndIf
   \State $m_\text{max} \gets \max(m_\text{max},\, m)$ \Comment{Equation~\ref{eq:nerf-counters}}
\EndFor
\State $p_\text{covered} \gets n_\text{valid}\,/\,N_\text{pixels}$ \Comment{Equation~\ref{eq:nerf-coverage}}
\If{$p_\text{covered} \ge \rho_\text{cover}$}
   \State \Return $m_\text{max}$ \Comment{Equation~\ref{eq:nerf-thin}}
\Else
   \State \Return $0$
\EndIf
\end{algorithmic}
\end{algorithm}

\subsection{2DGS instantiation}
\label{app:impl-gs}
All timings and memory footprints are measured on a single NVIDIA A100 card. 2DGS models are trained at default hyperparameters~\citep{huang20242d}. DivAS-GS adaptively determines the occupancy grid structure from the underlying Gaussian statistics of the reconstructed scene, avoiding manual voxel-resolution tuning across datasets. We therefore do not hand-select a per-dataset grid size. SAM-ViT-H~\citep{kirillov2023segment} is used throughout. All scenes are trained by downsampling the images by $4$. As the Gaussian Grouping authors noted, the SAM mask quality affects 3D segmentation results. We generate the SAM masks on images downsampled by $2$ to meet the memory-footprint constraint, then downsample to match the training image resolution. For consistency across methods and to account for VRAM limitations, we train on $\times 4$ downsampled images of the scene. For other methods, the SAM mask is generated at the same resolution as the training images.

The DivAS-GS pipeline exposes a single set of hyperparameters that is shared across LLFF and Mip-NeRF~$360^\circ$ (unlike the NeRF instantiation, the 2DGS kernels are not retuned per dataset). Table~\ref{tab:gs-hyperparams} groups them by pipeline stage. Voxel-grid construction is fully data-driven from the per-scene Gaussian statistics and exposes only the four constants of the adaptive rule, the segmentation kernels expose the kernel-side constants used by the thick path, the thin-structure path, and the voxel-to-surfel mapping. All values listed are the defaults used to produce the numbers in Section~\ref{sec:exp-quant}.

\begin{table}
\centering
\small
\caption{DivAS-GS hyperparameters used in all reported runs. The constants below are dataset-independent unless noted.}
\label{tab:gs-hyperparams}
\begin{tabular}{lll}
\toprule
Stage / Symbol & Value & Role \\
\midrule
\multicolumn{3}{l}{\emph{Voxel-grid construction (Section~\ref{sec:method-gs-fusion})}}\\
$\alpha_v$ & $3.0$ & Voxel-size scale on the median Gaussian footprint ($3\sigma$ rule) \\
$\beta_v$  & $1.0$ & Voxel-size scale on the scene radius/reference resolution \\
$g_0$ & $256$ & Reference grid resolution in the scene-scale term \\
$g_\text{max}$ & $512\,/\,1024$ & Grid-resolution cap, LLFF\,/\,Mip-NeRF~$360^\circ$ \\
$\tau_\text{TF}$ & $0.5$ & Transmittance floor defining the active-voxel list \\
AABB safety pad & $1.05$ & $b=\lceil 1.05\,R_\text{fg}\rceil$ \\
\midrule
\multicolumn{3}{l}{\emph{Fusion kernel thick path (Algorithm~\ref{alg:gs-fusion})}}\\
$\gamma$ & $1.0$ & Base depth-tolerance multiplier (in units of $h_v$) \\
$\beta$ & $0.05$ & Inverted per-sample tightening strength \\
$b_\text{max}$ & $0.5$ & Cap on the legacy bonus term \\
$\gamma_\text{ST}$ & $1.5$ & See-through gap as a multiple of $d_\text{cam}$ \\
$\tau_\text{thin}$ & $0.6$ & Thin-structure acceptance threshold \\
$\tau_\text{cons}$ & $0.5$ & Multi-view consensus threshold \\
Behind-margin & $2\,h_v$ & Asymmetric far-gate offset on the surface depth \\
\midrule
\multicolumn{3}{l}{\emph{Fusion kernel thin path (Algorithm~\ref{alg:gs-thin})}}\\
$\rho_\text{cover}$ & $0.5$  & Coverage ratio threshold on $n_{m\wedge d}/n_\text{depth}$ \\
$\rho_\text{occ}$ & $0.1$ & Fractional pixel floor of voxel footprint for the same pool \\
$n_\text{abs}$ & $25$ & Absolute pixel-count floor on the projected footprint \\
$3$D distance gate & $1.5\,h_v$ & Max unprojected-pixel distance to the voxel center \\
\midrule
\multicolumn{3}{l}{\emph{Voxel-to-surfel mapping (Algorithm~\ref{alg:vox2surfel})}}\\
$\tau_a$ & $\mathcal{P}_{99}(\{a_i\}_{i=1}^{M})$ & Per-scene needle aspect threshold \\
Tangent containment & $1\sigma$ & Pass 3 disc-overlap criterion \\
$\tau_M$ & $0.5$ & Pixel-shader mask threshold in Pass 3 \\
$\tau_\kappa$ & $0.5$ & Final boundary-bleeder coverage threshold \\
\bottomrule
\end{tabular}
\end{table}

\section{2DGS Fusion: Implementation Details}
\label{app:gs-impl-details}

These notes describe the execution mechanics that the main text (Section~\ref{sec:method-gs-fusion}) summarizes, and they accompany the 2DGS pseudocode in Algorithm~\ref{alg:gs-fusion} and Algorithm~\ref{alg:gs-thin}. The symbols follow the definitions in Section~\ref{sec:method-gs-fusion}.

\subsection{Occupancy grid construction}
The occupancy grid is built in four passes over the trained surfels. Surfels are first filtered against the symmetric cube by an axis-aligned bounding-box test in the re-centered frame. Each survivor is mapped to an integer voxel coordinate and to its one-dimensional Morton ($Z$-order) index, the standard bit-interleaving that places spatially adjacent voxels at adjacent linear addresses so that scatter and lookup remain cache-coherent. Because many surfels typically fall in a single voxel, the density grid is then populated by a Morton-indexed scatter-add of surfel opacities, with each voxel retaining only the aggregate, not the contributing surfel identities. Finally, voxels below the transmittance floor $\tau_\text{TF}$ are dropped, the thresholded grid is materialized in packed form for fast occupancy lookup, and a compact active-voxel list $\{m\,:\,\boldsymbol{\rho}[m] \ge \tau_\text{TF}\}$ of surviving indices is assembled alongside it. The active fraction is small in practice. On \textit{Fern} at $g{=}512$, for example, only about $0.07\%$ of voxels survive, with roughly nine surfels sitting on each active voxel on average, so all subsequent kernels iterate this list rather than the dense grid of $g^3$ entries.

\noindent\textbf{Computational advantages.}
The scene-adaptive compression of the explicit Gaussian scene, tailored for interactive 3D segmentation. Aggregating many surfels into one voxel is beneficial on several fronts that compound: it (a) reduces the working state from millions of primitives to a $g^3$ scalar field plus a sparse active list, cutting both kernel memory and warp-divergence costs, (b) lets the segmentation kernel iterate only over active voxels, so its launch cost scales with the foreground complexity of the scene rather than with the dense grid volume, (c) decouples downstream segmentation from per-surfel propagation, which would otherwise require expensive visibility and consensus reasoning on every primitive in every view, and (d) generalizes across scene scales without manual retuning, because every quantity used to build the grid center, extent, voxel edge, and occupancy threshold are derived from per-scene statistics. The resulting representation makes a single CUDA fusion kernel feasible at interactive rates for scenes containing millions of surfels.

\paragraph{Sparse active-voxel iteration.}
The fusion kernel assigns one CUDA thread to one entry of the active-voxel list. The thread decodes the Morton index $m$ back to integer grid coordinates, reconstructs the world-space voxel center $\vx_c$ in the scene-centered frame at edge length $h_v$, and skips the dense empty-voxel density test $\boldsymbol{\rho}_x\!<\!\rho_\text{thresh}$ because membership in the active list already implies occupancy. For each foreground view, the thread projects the center to normalized image coordinates $(u_x^i,v_x^i)=\pi_i(\vx_c)$, skips out-of-frame projections, samples the refined SAM confidence $m_x^i\!=\!\widetilde{\mM}_i(u_x^i,v_x^i)$ at the projected pixel, reads the per-pixel depth tuple $(\mD_\text{min}, \mD_\text{med}, \mD_\text{max}, \mD_\text{exp}, \mN_\text{samp})_i$ at the same location, and computes the voxel planar depth $x_d$ in that view.

\subsection{Footprint statistics via integral images}
Recovering the local mean and standard deviation of the splat-traversal count over a voxel's projected footprint by direct per-voxel summation would be prohibitive, since the grid can approach a billion voxels and a single footprint can span hundreds of pixels. Instead, once per view the kernel precomputes summed-area tables of the per-pixel splat count and of its square. The footprint mean and variance then follow from a four-corner inclusion-exclusion query in constant time per voxel. The same two tables serve both the thick-path tolerance tightening and the thin-path coverage gate, so the per-view preprocessing cost is amortized across all voxels that project into the view.

\subsection{Thin-structure path execution}
The thin path is entered only when the thick path did not vote for the current view and the voxel passes both the behind-surface gate and the thin-structure density floor. The kernel forms the eight voxel corners $\{\vx_c\pm\tfrac{h_v}{2}\boldsymbol{\epsilon}_j\}_{j=1}^{8}$, $\boldsymbol{\epsilon}_j\in\{-1,+1\}^3$, projects them into the view, and takes their axis-aligned pixel bounding box as the footprint, reusing the same integral-image statistics $(\bar n_\text{samp},\sigma_n)$ as the thick path. It then sweeps the footprint pixels, applying the per-pixel tightened depth gate and the layered or see-through asymmetric test, and increments the depth-consistent and the mask-and-depth-consistent counters accordingly. For each mask-positive pixel, the kernel unprojects it with the layered-aware depth, using the front-hit depth when the pixel is layered and the median depth otherwise, and updates the running minimum squared distance to the voxel center and the running maximum SAM confidence. The accumulated counters and distances are what feed the three closing conditions stated in the main text.

\subsection{Voxel-to-Surfel Mapping: Implementation Details}
\label{app:vox2surfel-impl}

These notes describe the execution of the three-pass voxel-to-surfel mapping summarized in Section~\ref{sec:method-vox2surfel}, and they accompany the pseudocode in Algorithm~\ref{alg:vox2surfel}. No new equations are introduced here; the symbols follow the definitions in Section~\ref{sec:method-vox2surfel}.

\paragraph{Tile-sorted gather pipeline}
Pass $3$ reuses the structure of the 2DGS rasterizer of \citep{huang20242d}, namely, preprocess, prefix sum, tile-key duplication, radix sort, tile-range identification, and a tile-parallel pixel kernel, because that tile decomposition is what makes a per-surfel gather tractable across millions of surfels. Only the differences from the rasterizer are described below, whereas the shared scaffolding is borrowed without modification.

\paragraph{Preprocess and view gating}
In the preprocess stage, each surfel surviving the needle cull is first gated against the per-voxel view-coverage bitmask $\vv$ produced by the fusion kernel (Equation~\ref{eq:vmask}). Surfel $i$ is admitted into a view $s$ only when the bit for its mapped voxel $x$ is set, which has no rasterization analogue and bounds the effective per-view surfel count to those that genuinely contributed segmentation evidence. The remaining preprocess work is identical to the rasterizer's, recovering the world-space tangent semi-axes  $(\vt_i^{(0)},\vt_i^{(1)})$ from $(\vq,\vs)$ from the surfel rotation and scales, projecting the center and the four tangent tips  $\boldsymbol{\mu}_i\pm\vt_i^{(0,1)}$ to pixel coordinates, computing the projected-ellipse inverse determinant for the in-pixel test, and counting the tiles each surfel touches.

\paragraph{Depth-free tile sort}
Because coverage counting has no front-to-back ordering dependency, the duplicated tile keys carry only the tile id and leave the lower $32$ bits zero, so the CUB radix sort runs on the tile-id bits ($32 + \lceil\log_2 n_\text{tiles}\rceil$) alone, saving sort work and memory bandwidth relative to the rasterizer, which packs per-tile depth into the low bits for blending.

\paragraph{Coverage gather kernel}
The tile-parallel kernel loads each tile's surfels into shared memory in the same cooperative batches as the 2DGS~\citep{huang20242d} rasterizer and visits the tile's pixels. The only divergence is the bounding box, which determines which pixels are visited. DivAS-GS emits a tight one-sigma box whose half-extents are the projected lengths of the longer tangent semi-axis in each image axis, with no $3\times$ scaling, so the one-sigma to three-sigma annulus is never enumerated. On every visited pixel, the kernel applies the rasterizer's exact three-sigma ellipse cutoff in tangent-plane coordinates and, on a hit, issues two atomic increments to the per-surfel visited and mask-positive counters. After a view completes, a per-surfel reduction forms the view coverage and folds it into the running cross-view maximum, which is used by the final threshold $\tau_\kappa$.

\subsection{2DGS Kernel Algorithms}
\label{app:gs-algorithms}
This subsection lists the three CUDA kernels of the 2DGS instantiation referenced in the main text: the thick-path depth-weighted fusion kernel (Algorithm~\ref{alg:gs-fusion}), the thin-structure footprint-coverage fallback (Algorithm~\ref{alg:gs-thin}), and the voxel-to-surfel mapping (Algorithm~\ref{alg:vox2surfel}).

\begin{algorithm}
\caption{Thick-Structure Segmentation for 2DGS via Depth-Weighted Voxel Aggregation}
\label{alg:gs-fusion}
\begin{algorithmic}
\Require active-voxel list $\{m_a\}_{a=1}^{n_\text{active}}$; refined masks $\{\widetilde{\mM}_i\}_{i=1}^{S}$ with depth maps $(\{\mD^{(s)}_\text{min}\}_{s=1}^S, \{\mD^{(s)}_\text{med}\}_{s=1}^S, \{\mD^{(s)}_\text{max}\}_{s=1}^S, \{\mN^{(s)}_\text{samp}\}_{s=1}^S)$, per-view camera-target distances $\vd_\text{cam}$, integral images $\{\tI_n,\tI_{n^2}\}$, density grid $\boldsymbol{\rho}$, 3D occupancy grid $\tP$,, view camera matrices $\{\mC_i\}_{i=1}^{S}$ (each $\mC_i$ represent the c2w matrix), voxel view mask $\vv$, hyperparameters $(\gamma,\beta,\gamma_\text{ST},\rho_\text{thin\_thresh},\tau_\text{thin},\alpha_1,\tau_\text{cons},\varepsilon)$

\For{each $a=1,\dots,n_\text{active}$ \textbf{in parallel}}
   \State decode Morton index $m_a$ to grid coordinates and reconstruct world center $\vx_c$, edge $h_v$, density $\boldsymbol{\rho}_x$
   \State $W_x \gets 0,\ Z_x \gets 0,\ \text{vmask} \gets 0$
   \For{each view $i=1,\dots,S$}
      \State extract camera-forward axis $\hat{\ve}_\text{fwd}^i \gets \mC_i[2]$ and camera origin $\vo_i \gets \mC_i[3]$ \Comment{rows of c2w matrix}
      \State project: $(u_x^i,v_x^i) \gets \pi(\vx_c,\mC_i)$;\quad \textbf{if} out-of-frame \textbf{continue}
      \State sample $m_x^i \gets \widetilde{\mM}_i(u_x^i,v_x^i)$ and \quad obtain voxel depth $x_d$ in view $i$
      \State $d_\text{min},d_\text{med},d_\text{max},n_\text{samp} \gets (\mD^{(i)}_\text{min},\mD^{(i)}_\text{med},\mD^{(i)}_\text{max},\mN^{(i)}_\text{samp})_{u_x^i,v_x^i}$
      \State Derive pixel ray direction $\hat{\vd}_i$ through $(u_x^i,v_x^i)$ from $\mC_i$
      \State $\cos\theta \gets \langle\hat{\ve}_\text{fwd}^i,\hat{\vd}_i\rangle$, \quad
             $d_\bullet^\text{ray} \gets d_\bullet/\cos\theta$ \Comment{ray-space conversion, Equation~\ref{eq:planar-to-ray}; $\bullet\in\{\text{min},\text{med},\text{max}\}$}
      \State $\text{check\_thin} \gets \textbf{True}$
      \If{$m_x^i \ge 0.5$} \Comment{thick path entry}
         \State $(\bar n_\text{samp},\sigma_n) \gets$ four-corner query of $(\tI_n,\tI_{n^2})_i$ over the projected footprint
         \State $\text{layered} \gets (n_\text{samp}>\bar n_\text{samp}+\sigma_n)$,\quad
                $\text{seethru} \gets (d_\text{med}>\gamma_\text{ST}\,d_\text{cam}^{(i)})$ \Comment{Equation~\ref{eq:seethrough}}
         \State $\text{trusted} \gets (d_\text{min}\le d_\text{med})$,\quad
                $\text{front} \gets (\text{layered}\vee\text{seethru})\wedge\text{trusted}$
         \If{not $\text{trusted}$}
            \State skip thick path vote for this view \Comment{tilted-pixel anomaly; defer to thin path}
         \Else
            \State $d_\text{ref} \gets d_\text{min}$ if front else $d_\text{med}$,\quad
                   $\hat t \gets \text{clamp}\bigl(\langle\vx_c-\vo_i,\hat{\vd}_i\rangle,\,d_\text{min},\,d_\text{ref}\bigr)$
            \State $\delta_\text{dist} \gets \|\vx_c-(\vo_i+\hat t\,\hat{\vd}_i)\|$,\quad
                   $\tau_\text{spatial} \gets h_v\,\pi_\text{dg}(u_x^i,v_x^i)$ \Comment{Equation~\ref{eq:gs-spatial}}
            \State $E \gets \max(n_\text{samp}-\bar n_\text{samp},0)$,\quad
                   $t_\text{depth} \gets \gamma h_v/(1+\beta E)$ \Comment{Equation~\ref{eq:tightening}}
            \State $\text{valid\_d} \gets$ asymmetric gate using $x_d$, $d_\text{min}$, $d_\text{med}$, front \Comment{Equation~\ref{eq:asym-gate}}
            \If{$\delta_\text{dist}\le\tau_\text{spatial}$ \textbf{and} $\text{valid\_d}$}
               \State $d_\text{ctr} \gets \tfrac12(d_\text{min}^\text{ray}+d_\text{max}^\text{ray})$, \quad
                      $d_\text{hw} \gets \text{max}\bigl(\tfrac12(d_\text{max}^\text{ray}-d_\text{min}^\text{ray}),\varepsilon\bigr)$
               \State $w_\text{depth} \gets \exp\bigl(-\alpha_1\,((\hat t-d_\text{ctr})/d_\text{hw})^2\bigr)$ \Comment{Equation~\ref{eq:gs-depthweight}}
               \State $W_x \mathrel{+}= m_x^i\,w_\text{depth}$, \quad $Z_x \mathrel{+}= w_\text{depth}$, \quad
                      $\text{vmask} \mathrel{|}= (1\!\ll\!i)$, \quad $\text{check\_thin} \gets \textbf{False}$
            \EndIf
         \EndIf
      \EndIf
      \Statex \emph{// Behind-surface gate before thin fallback}
      \State $d_\text{surf} \gets d_\text{min}$ if (trusted $\wedge$ seethru) else $d_\text{med}$, \quad
             $\text{behind\_fail} \gets (x_d>d_\text{surf}+2 h_v)$ \Comment{Equation~\ref{eq:behind-gate}}
      \If{$\text{check\_thin}$ \textbf{and} not $\text{behind\_fail}$ \textbf{and} $\boldsymbol{\rho}_x\ge\rho_\text{thin\,thresh}$}
         \State $t_x^i \gets \textsc{ThinStructureGS}(x,i)$ \Comment{call Algorithm~\ref{alg:gs-thin}}
         \If{$t_x^{\,i}\ge\tau_\text{thin}$}
            \State $W_x \mathrel{+}= t_x^{\,i}$, \quad $Z_x \mathrel{+}= 1$, \quad $\text{vmask} \mathrel{|}= (1\!\ll\!i)$
         \EndIf
      \EndIf
   \EndFor
   \State $\tP_x \gets W_x/Z_x$ if $Z_x>\varepsilon$ else $0$ \Comment{Equation~\ref{eq:compact-fusion}}
   \State $\vv_x \gets \text{vmask}$ if $\tP_x\ge\tau_\text{cons}$ else $0$ \Comment{Equation~\ref{eq:vmask}}
\EndFor
\State \Return $(\tP, \vv)$
\end{algorithmic}
\end{algorithm}

\begin{algorithm}
\caption{Thin-Structure Segmentation for 2DGS via Two-Counter Footprint Coverage}
\label{alg:gs-thin}
\begin{algorithmic}
\Require voxel $x$ with center $\vx_c$, edge $h_v$, voxel depth $x_d$, view $i$ with refined mask $\widetilde{\mM}_i$, depth maps $(\{\mD^{(s)}_\text{min}\}_{s=1}^S, \{\mD^{(s)}_\text{med}\}_{s=1}^S, \{\mD^{(s)}_\text{max}\}_{s=1}^S, \{\mN^{(s)}_\text{samp}\}_{s=1}^S)$, camera $(\mK_i,\mT_i)$, integral images $\tI_n,\tI_{n^2}$, hyperparameters $(\gamma,\beta,\rho_\text{cover},n_\text{abs},\rho_\text{occ}, \varepsilon)$

\State Form the 8 voxel corners $\{\vx_j^i = \vx_c + \tfrac{h_v}{2}\boldsymbol{\epsilon}_j\}_{j=1}^{8}$, $\boldsymbol{\epsilon}_j\in\{-1,+1\}^3$
\State $(u_j^x, v_j^x) \gets \pi_i(\vx_j^i)$ Project corners.  
\State Construct the axis-aligned bounding box $\sB_x^i = [u_\text{min},u_\text{max}] \times [v_\text{min},v_\text{max}]$ in pixel coordinates, clipped to the image
\State area $A_\sB = (u_\text{max} - u_\text{min}) \times (v_\text{max} - v_\text{min})$
\State $n_\text{min}\gets\max(n_\text{abs},\lceil\rho_\text{occ}A_\sB\rceil)$ \Comment{Equation~\ref{eq:thin-nmin}}
\State $(\bar n_\text{samp},\sigma_n)\gets$ four-corner query of $(\tI_n,\tI_{n^2})_i$ over $\sB_x^i$
\State $n_\text{depth}\gets 0$, \quad $n_{m\wedge d}\gets 0$, \quad $m_\text{max}\gets 0$, \quad $d_\text{min\_to\_surf}^2\gets+\infty$
\For{each pixel $(x_p,y_p)\in\sB_x^i$}
   \State read $\widetilde{\mM}_i(x_p,y_p)$
   \State $d_\text{min}, d_\text{med}, d_\text{max}, n_\text{samp} \gets (\mD^{(i)}_\text{min}, \mD^{(i)}_\text{med}, \mD^{(i)}_\text{max}, \mN^{(i)}_\text{samp})_{x_p,y_p}$
   \If{$d_\text{med}<\varepsilon$} \textbf{continue}  \EndIf
   \State $E\gets\text{max}(n_\text{samp}-\bar n_\text{samp},0)$,\quad $\eta\gets 1/(1+\beta E)$,\quad
          $\tau_\text{depth}^\text{thin}\gets\gamma h_v\eta$
   \State $\text{layered}\gets(n_\text{samp}>\bar n_\text{samp}+\sigma_n)$
   \State $\text{valid\_d}\gets$ asymmetric per-pixel gate \Comment{Equation~\ref{eq:asym-gate}}
   \If{$\text{valid\_d}$}
      \State $n_\text{depth}\mathrel{+}=1$
      \If{$\widetilde{\mM}_i(x_p,y_p)>0.5$}
         \State $n_{m\wedge d}\mathrel{+}=1$,\quad
                $m_\text{max}\gets\text{max}(m_\text{max},\widetilde{\mM}_i(x_p,y_p))$
         \State unproject $(x_p,y_p)$ to $\vp_w^{(x_p,y_p)}$ using $d_\text{min}$ if layered else $d_\text{med}$
         \State $d_\text{min\_to\_surf}^2\gets\text{min}(d_\text{min\_to\_surf}^2,\|\vp_w^{(x_p,y_p)}-\vx_c\|^2)$ \Comment{Equation~\ref{eq:thin-d2min}}
      \EndIf
   \EndIf
\EndFor
\If{$n_\text{depth}<n_\text{min}$} \Return $0$ \EndIf \Comment{degeneracy floor}
\State $p_\text{covered}\gets n_{m\wedge d}/n_\text{depth}$ \Comment{Equation~\ref{eq:gs-coverage}}
\If{$p_\text{covered}\ge\rho_\text{cover}$ \textbf{and} $d_\text{min-to-surf}^2\le(1.5\,h_v)^2$}
   \State \Return $m_\text{max}$ \Comment{Equation~\ref{eq:thin-tthin-gs}}
\Else
   \State \Return $0$
\EndIf
\end{algorithmic}
\end{algorithm}

\begin{algorithm}
\caption{Voxel-to-Surfel Mapping for 2DGS}
\label{alg:vox2surfel}
\begin{algorithmic}
\Require trained surfel set $\sG=\{g_i=(\boldsymbol{\mu}_i,(s_{i,1},s_{i,2}),\vq_i,\alpha_i)\}_{i=1}^{M}$;\ scene center $\vc_\text{scene}$, AABB half-extent $b$, voxel edge $h_v$, segmentation bitfield $\tB^\text{seg}$(packed from $\tP$), per-voxel view-coverage bitmask $\vv$, refined SAM masks $\{\widetilde{\mM}_s\}_{s=1}^{S}$ over the $S$ retained views, thresholds $(\tau_a,\,\tau_M,\,\tau_\kappa)$

\Statex \emph{// Pass 1: Bitfield lookup (permissive starting set)}
\For{$i\in\{i:\lnot\text{NaN}(\boldsymbol{\mu}_i)\wedge\|\boldsymbol{\mu}_i-\vc_\text{scene}\|_\infty<b\}$ \textbf{in parallel}}
   \State $m_i\gets\texttt{morton3D}\!\bigl(\lfloor(\boldsymbol{\mu}_i-\vc_\text{scene}+b)/h_v\rfloor\bigr)$ \Comment{Voxel index of $g_i$}
   \State $\vf_i^{(1)}\gets\tB^\text{seg}_{m_i}$ \Comment{Equation~\ref{eq:pass1}}
\EndFor

\Statex \emph{// Pass 2: Needle-floater culling}
\For{each $i$ \textbf{in parallel}}
   \State $a_i\gets\max(s_{i,1},s_{i,2})\,/\,(\min(s_{i,1},s_{i,2})+\epsilon)$ \Comment{aspect ratio}
   \State $\vf_i^{(2)}\gets\vf_i^{(1)}\wedge(a_i\le\tau_a)$ \Comment{Equation~\ref{eq:pass-needle}}
\EndFor

\Statex \emph{// Pass 3 Footprint coverage analysis (no rendering, no alpha, no depth ordering, no color)}
\State initialize $\kappa_i\gets 0$ for all $i$
\For{each view $s=1,\dots,S$}
   \Statex \quad (i) Voxel view mask pruning. \emph{Process surfel $g_i$ in view $s$ only when its associated voxel was marked for that view.}
   \State $\va_s\gets\bigl\{\,i\,:\,\vf_i^{(2)}{=}1\ \wedge\ s\in\vv_{m_i}\,\bigr\}$ \Comment{semantic view restriction}

   \Statex \quad (ii) Surfel projection and $1\sigma$ bounding-box construction.
   \For{each $i\in\va_s$ \textbf{in parallel}}
       \State Recover the two world-space tangent semi-axes $(\vt_i^{(0)},\vt_i^{(1)})$ from $(\vq_i,s_{i,1},s_{i,2})$
       \State Project the center $\boldsymbol{\mu}_i$ and the four tangent tips $\boldsymbol{\mu}_i\pm\vt_i^{(0,1)}$ into view $s$ to obtain the screen-space surfel centre $(c_{i,x},c_{i,y})$ and the projected tangent vectors $(\boldsymbol{\alpha}_i^{(0)},\boldsymbol{\alpha}_i^{(1)})$
       \State \textbf{1-$\sigma$ bbox}\ $\sB_{i,s}^{1\sigma}\!\gets\![c_{i,x}{\pm}e_x]\!\times\![c_{i,y}{\pm}e_y]$,\ $e_x{=}\text{max}(|\alpha_{i,x}^{(0)}|,|\alpha_{i,x}^{(1)}|)$, $e_y$ analogous \Comment{tighter than the rasterizer's $3\sigma$ bbox, excludes the $1\sigma\!\to\!3\sigma$}
   \EndFor

   \Statex \quad (iii)Tile assignment and tile-key sort.
   \State Enumerate, for $i\in\va_s$, the tiles overlapped by $\sB_{i,s}^{1\sigma}$, emitting the set of surfel-tile pairs $\{(t,i)\}$
   \State Sort the pairs by tile id \Comment{Equation~\ref{eq:keys-no-depth}}
   \State Identify, for each tile $t$, the contiguous range of surfels assigned to it

   \Statex \quad \textit{(iv)~Tile-parallel footprint gather, $1\sigma$ bbox traversal $+$ $3\sigma$ Cramer acceptance.}
   \State Initialize per-surfel counters $\text{total\_fp}_i^{(s)}\gets 0$, $\text{mask\_fp}_i^{(s)}\gets 0$ for $i\in\va_s$
   \For{each tile $t$ \textbf{in parallel}}
       \For{each pixel $(x_p,y_p)$ inside tile $t$}
           \For{each surfel $i$ assigned to tile $t$ with $(x_p,y_p)\in\sB_{i,s}^{1\sigma}$} \Comment{Step A: $1\sigma$ bbox traversal}
                \State Evaluate the standard 3$\sigma$ ellipse test, if cleared $\text{total\_fp}_i^{(s)}\,\mathrel{+}=\,1$, \If{$\widetilde{\mM}_s(x_p,y_p)\ge\tau_M$} $\text{mask\_fp}_i^{(s)}\,\mathrel{+}=\,1$ \Comment{If clears the ellipse test} \Comment{Equation~\ref{eq:cov-atomics}} \EndIf
            \EndFor
       \EndFor \EndFor

   \Statex \quad (v) Per-view coverage and max-over-views aggregation.
   \For{each $i\in\va_s$ \textbf{in parallel}}
       \State $\kappa_i^{(s)}\gets\text{mask\_fp}_i^{(s)}\,/\,\max(\text{total\_fp}_i^{(s)},\,1)$, $\kappa_i\gets\max\!\bigl(\kappa_i,\,\kappa_i^{(s)}\bigr)$ \Comment{Equation~\ref{eq:kappa}, one clean view suffices averaging would dilute partial occlusions}
   \EndFor
\EndFor
\Statex \emph{// Final indicator: threshold on max-view coverage}
\For{each $i$ \textbf{in parallel}}
   \State $\vf_i\gets\vf_i^{(2)}\wedge(\kappa_i\ge\tau_\kappa)$ \Comment{Equation~\ref{eq:final-F} $\vf_i\equiv\vf_i^{(3)}$}
\EndFor
\State \Return $\vf$
\end{algorithmic}
\end{algorithm}

% ================================================================
\section{View Selection and View Generation}
\label{app:view-selection}
This section provides a detailed description of the view-generation strategies used in our framework. For LLFF scenes, which are forward-facing, a small set of anchor views is selected manually because most of the object geometry is visible from a narrow cone of viewpoints, so a few clicks suffice. For Mip-NeRF~$360^\circ$ scenes, rely exclusively on automatically generated \emph{Fibonacci views}. These are obtained by sampling $N$ uniformly distributed directions on a sphere using Fibonacci sampling and converting each direction into a camera-to-world (c2w) matrix. The synthesized views are then ranked using our geometric view-scoring strategy and the top-$K$ are surfaced as anchors. The following subsections describe the Fibonacci sampling procedure, the geometric ranking algorithm, and the centroid-aligned refinement views used during progressive segmentation. The same LLFF-manual, Mip-NeRF-Fibonacci division is used by both the NeRF and the 2DGS instantiations.
% ----------------------------------------------------------------
\subsection{Fibonacci Sphere Sampling}
\label{app:fibonacci}
To generate uniformly distributed viewpoints for unbounded Mip-NeRF $360^\circ$ scenes, we adopt the Fibonacci lattice sampling method~\citep{gonzalez2010measurement}. This approach distributes $N$ points on a unit sphere with minimal angular clustering and excellent low-discrepancy properties.

Let the golden ratio be
$
\phi = \frac{1 + \sqrt{5}}{2}, \qquad \phi^{-1} \approx 0.618.
$
The azimuth of the $i$-th sample is chosen using the golden-angle increment:
$
\theta_i = 2\pi i \phi^{-1},
$
which produces an angular separation of approximately $137.5^\circ$ between consecutive points. The corresponding elevation is computed with an equal-area parameterization:
$
\phi_i = \arccos\!\left(1 - \frac{2i}{N}\right).
\label{eq:fib}
$

\begin{figure}
    \centering
    \includegraphics[width=0.6\linewidth]{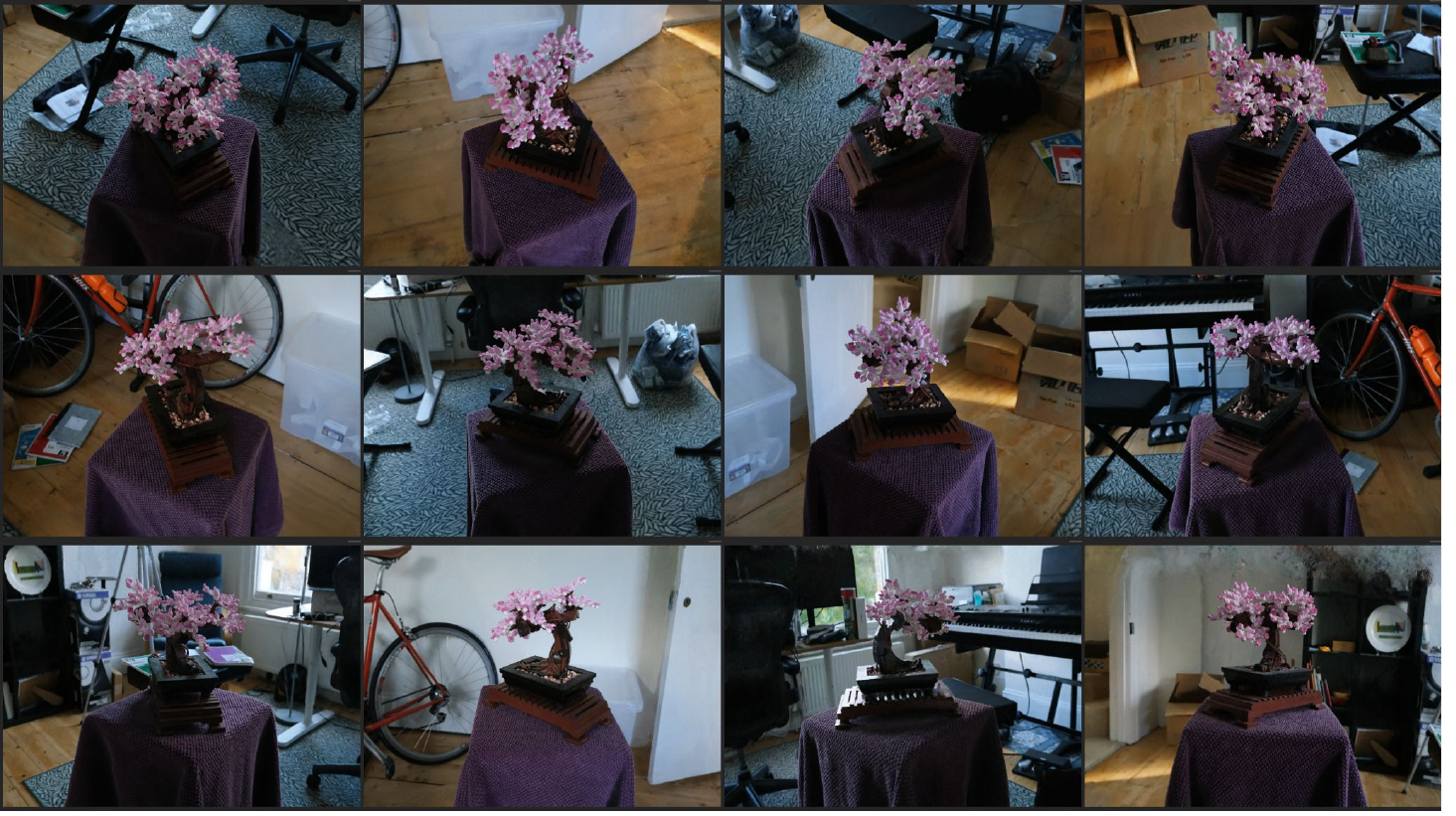}
    \caption{Fibonacci-based sampled views.}
    \label{fig:fibonacci_views}
\end{figure}

Each $(\theta_i, \phi_i)$ pair defines a direction vector on the sphere, which we convert into a camera-to-world (c2w) matrix to synthesize a novel view of the scene. All posed training images are processed once to generate these $N$ \emph{Fibonacci views}, which form the candidate pool for geometric ranking. In our experiments, $N{=}12$ provides sufficient angular diversity for Mip-NeRF~$360^\circ$, from which we retain the top-$K$ ($K{=}5$) ranked views for user annotation as shown in Figure~\ref{fig:fibonacci_views}.

\paragraph{2DGS instantiation.}
The Fibonacci procedure is representation-agnostic, it only requires camera-to-world matrices and a renderer to synthesize the candidate images, and it is used for all Mip-NeRF~$360^\circ$ scenes in both instantiations. For the 2DGS instantiation we reuse the same azimuth and elevation construction (Equation~\ref{eq:fib}), then render each candidate through the 2DGS rasterizer (Section~\ref{sec:method-gs-fusion}). The 2DGS case differs in the \emph{radius} at which each anchor is placed. Because surfels are reliably reconstructed only within the shell spanned by the training cameras, a fixed global radius can place anchors in unsupported regions and introduce floaters and holes. We therefore make the radius adaptive per Fibonacci direction (Section~\ref{sec:method-anchor}), keeping the camera within the valid directional shell. The pool size $N{=}12$ and the retained top-$K{=}5$ are unchanged from the NeRF case, so the bound on per-scene user effort holds across both backbones.

% ----------------------------------------------------------------
\subsection{Geometric View Ranking Algorithm}
\label{app:geometric-ranking}

To identify the most informative viewpoints, we rank the Fibonacci-sampled views using a composite geometric score. This scoring mechanism prioritizes views that maximize spatial diversity and align with canonical scene axes. We compute three components for each candidate view $v_i$ to determine its final rank.

\textbf{Diversity Score ($D_i$).} We calculate the mean pairwise cosine distance between the viewing direction of view $v_i$ and all other candidates. This penalizes clustering, ensuring the selected views are distributed uniformly around the scene.

\textbf{Cardinal Coverage ($C_i$).} We favor views that align with standard canonical perspectives (front, back, left, right, top, bottom). We compute the inverse distance between the view vector and the nearest global coordinate axis.

\textbf{Pitch Extrema ($P_i$).} We normalize the absolute pitch angle of the camera to reward high-latitude viewpoints, helping reveal geometry often occluded in purely horizontal trajectories (e.g., object tops).

The final score for view $i$ is a weighted sum:
\begin{equation}
    I(i) = 0.4\,D_i + 0.3\,C_i + 0.3\,P_i,
\end{equation}
, where $D_i$, $C_i$, and $P_i$ denote normalized diversity, coverage, and pitch-extremity scores, respectively.

\begin{figure}
    \centering
    \includegraphics[width=0.6\linewidth]{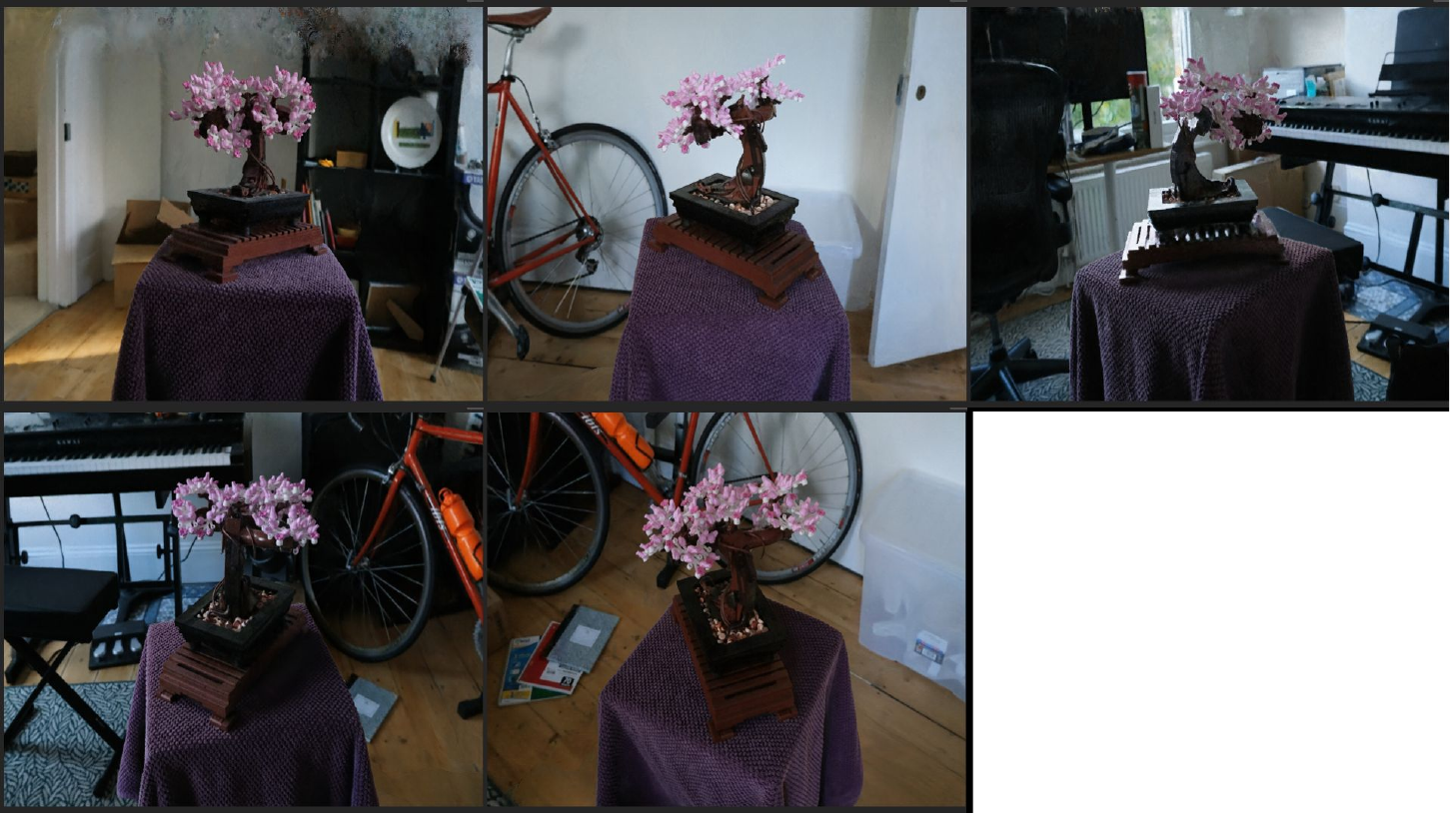}
    \caption{Top-5 views via geometric view ranking.}
    \label{fig:geometric_ranking}
\end{figure}

The ranking procedure is summarized in Algorithm~\ref{alg:view_ranking}, and an example result is shown in Figure~\ref{fig:geometric_ranking}.

\begin{algorithm}
\caption{Geometric View Ranking}
\label{alg:view_ranking}
\begin{algorithmic}
\Require Candidate trajectory $\mathcal{T} = \{v_i\}_{i=1}^N$, weights $w_d, w_c, w_p$
\State Define canonical axes $A = \{\pm \mathbf{x}, \pm \mathbf{y}, \pm \mathbf{z}\}$
\For{each view $v_i \in \mathcal{T}$}
    \State \textbf{Diversity:} Mean cosine distance to other views
    \Statex \quad $D_i = \frac{1}{N-1} \sum_{j \neq i} (1 - \mathbf{d}_i \cdot \mathbf{d}_j)$
    \State \textbf{Cardinality:} Inverse distance to nearest axis
    \Statex \quad $C_i = \max_{a \in A} (1 / (\|\mathbf{d}_i - \mathbf{a}\| + \epsilon))$
    \State \textbf{Pitch:} Normalized absolute pitch
    \Statex \quad $P_i = |\theta_i| / 90^{\circ}$
\EndFor
\State Normalize scores $D, C, P$ to $[0,1]$
\State Compute final score: $I_i = w_d D_i + w_c C_i + w_p P_i$
\State \Return Top-$K$ views with highest $I_i$
\end{algorithmic}
\end{algorithm}

\paragraph{2DGS instantiation.}
The ranking score depends only on the viewing direction $\mathbf{d}_i$ and the elevation angle, both of which are read directly from the c2w matrix of each candidate camera. The score is therefore independent of the underlying representation. We apply Algorithm~\ref{alg:view_ranking} unchanged to the 2DGS Fibonacci pool and surface the same top-$K{=}5$ as anchors.
% ----------------------------------------------------------------
\subsection{Centroid View Generation}
\label{app:centroid-view}
\begin{figure}
    \centering
    \includegraphics[width=0.6\linewidth]{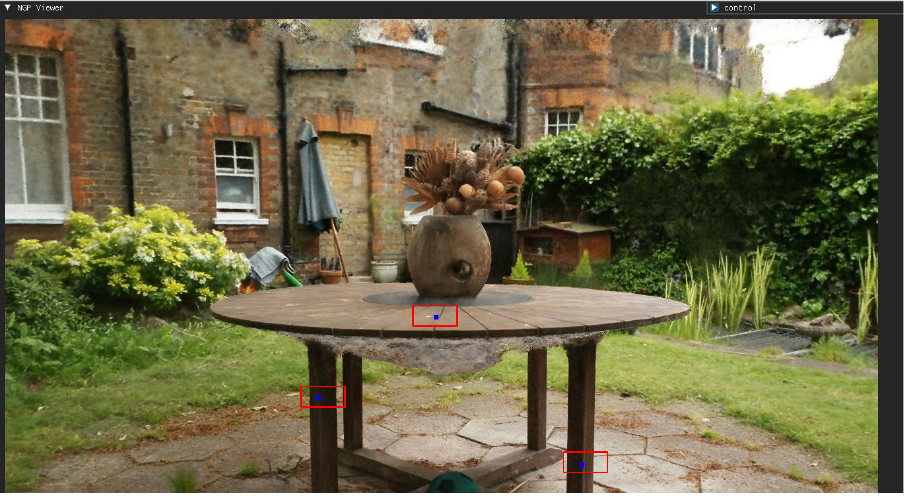}
    \caption{Anchor view with user prompts.}
    \label{fig:anchor_view}
\end{figure}

For each selected anchor view shown in Figure~\ref{fig:anchor_view}, the user provides a point prompt on the object. We generate a localized \emph{centroid view} by back-projecting this 2D point into 3D space and advancing the camera along the view vector. This operation effectively zooms in on the target region, reducing background clutter and increasing the sampling resolution for the subsequent segmentation step as shown in Algorithm~\ref{alg:centroid_view}.

\paragraph{NeRF instantiation}
The $3$D look-at point is obtained by back-projecting the $2$D prompt using the $\mD_\text{mw}$ depth along the ray, rather than the expected-depth from volume rendering integral of~\citet{liu2024sanerf}. This yields a sharper estimate of the true surface crossing and is computed in $O(1)$ inside the existing ray-marching loop. The camera is then advanced toward the look-at point by a fixed zoom fraction (typically $0.67$ for LLFF and $0.47$ for
Mip-NeRF~360$^\circ$) though they are interactively adjustable. Because the trained NeRF is bounded by its AABB, every ray is automatically clipped to a finite, artifact-free depth range, so a fixed push never drives the camera into an undefined region.

\begin{algorithm}
\caption{Centroid View Generation (NeRF instantiation)}
\label{alg:centroid_view}
\begin{algorithmic}[1]
\Require Anchor view camera $C_\text{anchor}$, 2D point prompt $u \in \mathbb{R}^2$, zoom fraction $\lambda \in (0,1)$
\State \textbf{Unprojection:}
\Statex \quad Retrieve depth $d = \text{DepthMap}(u)$
\Statex \quad Compute 3D point $x_\text{world} = \text{BackProject}(u, d, C_\text{anchor})$
\State \textbf{Camera Update:}
\Statex \quad Get current position $p = \text{Pos}(C_\text{anchor})$
\Statex \quad New position $p' = p + \lambda (x_\text{world} - p)$
\Statex \quad New look-at direction $f' = \text{normalize}(x_\text{world} - p')$
\State Form new view matrix $C_\text{new}$ looking at $x_\text{world}$ from $p'$
\State \Return $C_\text{new}$
\end{algorithmic}
\end{algorithm}

\begin{figure}[H]
    \centering
    \includegraphics[width=0.9\linewidth]{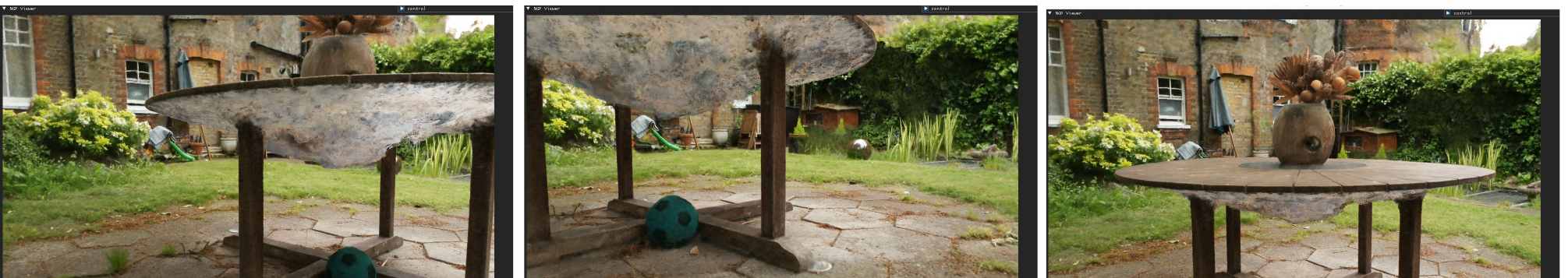}
    \caption{Centroid zoom views based on user prompts on the anchor view in Figure~\ref{fig:anchor_view}.}
    \label{fig:centroid_zoom_views}
\end{figure}

These centroid views, shown in Figure~\ref{fig:centroid_zoom_views}, serve as the input to SAM. The resulting high-resolution $2$D masks are then fused via the voxel-aggregation kernel described in Section~\ref{app:nerf-fusion}.

\paragraph{2DGS instantiation: physical-radius neighborhood and occlusion-aware refinement.}
The centroid procedure is unchanged in spirit; the representation-specific element is the unprojection depth $d$ at the prompt pixel $u$, read from the 2DGS rasterizer's median-depth output ($\mD_\text{med}$) rather than the NeRF expected-depth channel. Because 2DGS scenes are unbounded, the fixed-fraction push of Algorithm~\ref{alg:centroid_view} does not transfer; each 2DGS centroid view is instead generated in two strategies (Algorithm~\ref{alg:centroid-gs}): a neighborhood-driven radius estimate, followed by an occlusion-aware radius refinement.

\subparagraph{Strategy 1: Local Gaussian neighborhood and initial radius.}
Let $z=\mD_\text{med}(\vu)$ be the median depth at the clicked pixel $\vu$ and $vx_w=\texttt{BackProject}(\vu,z)$ the unprojected $3$D point. We first define a \emph{physical search radius} as the world-space extent that projects to a fixed patch fraction $p_f$ of the image width,
\begin{equation}
R_\text{phys}=p_f\,z\,\tan(\mathrm{FOV}_x/2), \text{where, }\;\; p_f=0.25,
\label{eq:Rphys}
\end{equation}
and gather the local neighborhood of opaque surfels inside this sphere whose opacity is represented by $\alpha$,
\begin{equation}
\mathcal{N} = \bigl\{i : \|\mu_i-\displaystyle \vx_w\| < R_\text{phys}\ \wedge\
\alpha_i > \alpha_\text{min}\bigr\}, \text{where, } \mu_i \text{ is surfel center}\in \mathcal{G},\; \alpha_\text{min}=0.1.
\label{eq:neigh}
\end{equation}
If $|\mathcal{N}|<5$ the prompt landed in a sparsely populated region (thin structure, far background, or partial occlusion). We set the centroid radius $r_\text{init}=z\,p_f$ and let Strategy $2$ refine it. For dense neighborhoods, we estimate the lateral spread \emph{perpendicular to the view direction} so the radius is invariant to depth-aligned elongation. With unit view direction $\hat{\mathbf{v}}$ and
$\mathbf{r}_i=\mu_i-\mathbf{x}_w$, the in-plane offset and its robust extent are
\begin{equation}
\mathbf{r}_i^{\perp} = \mathbf{r}_i - (\mathbf{r}_i\!\cdot\!\hat{\mathbf{v}})\hat{\mathbf{v}},
\qquad
R_\text{lat} = \mathcal{P}_{90}\bigl(\{\|\mathbf{r}_i^{\perp}\|\}_{i\in\mathcal{N}}\bigr),
\label{eq:Rlat}
\end{equation}
where the $90$th percentile rejects stray boundary surfels. We then place the orbital camera so that the neighborhood of lateral half-width $R_\text{lat}$
subtends a target fraction $f_\text{fill}$ of the frame. We set the centroid radius:
\begin{equation}
r_\text{init} = \frac{R_\text{lat}}{f_\text{fill}\,\tan(\mathrm{FOV}_x/2)},
\qquad f_\text{fill}=0.5,
\label{eq:Rinit}
\end{equation}
so that larger clusters push the camera back and tighter clusters draw it in.

Strategy $1$ also returns a $2$D visualization box on the \emph{anchor} view, so the user can immediately verify that the click captured the intended object extent
without waiting for the centroid render.

\begin{figure}[t]
\centering
\begin{subfigure}[b]{0.48\linewidth}
\centering
\includegraphics[width=\linewidth]{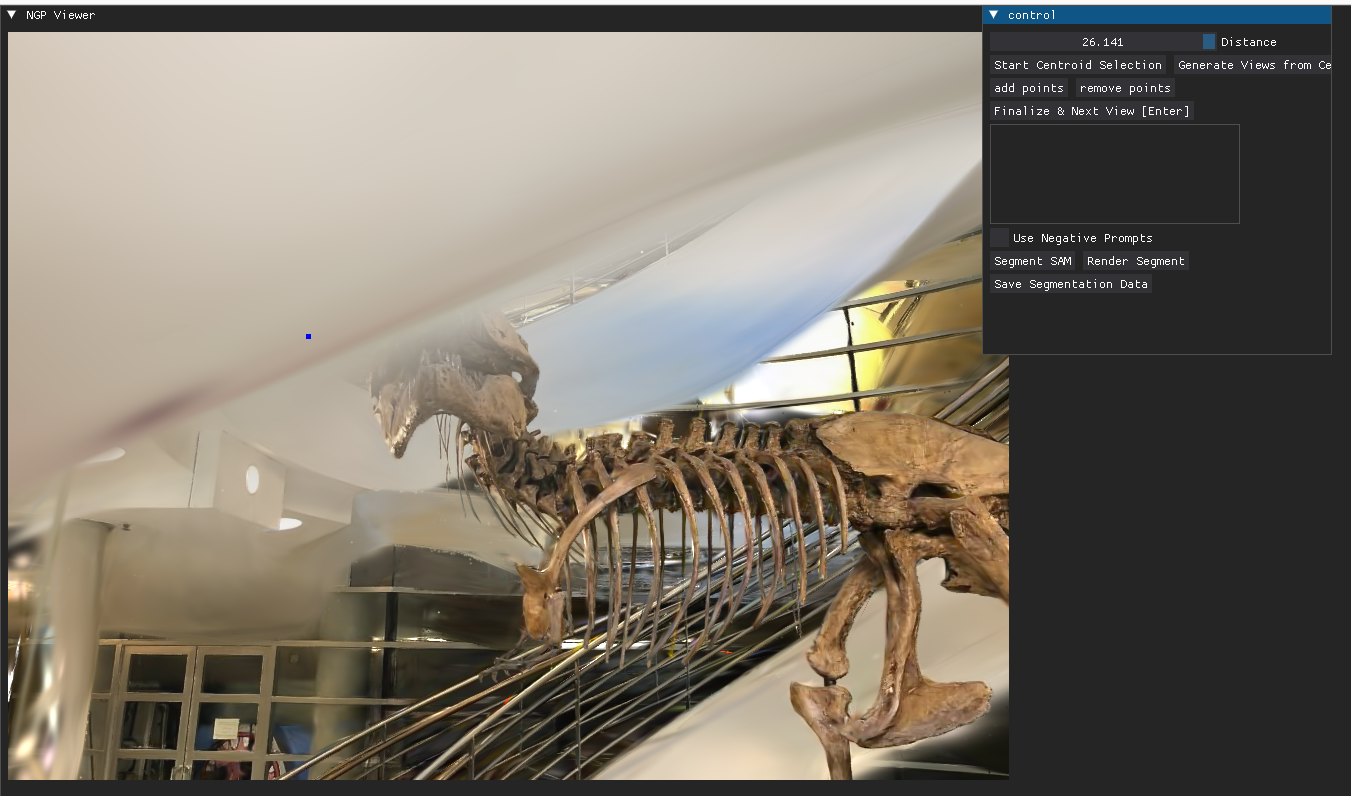}
\caption{Raw depth-driven radius ($r_\text{init}{=}26.14$): occluders intrude on the line of sight.}
\label{fig:strategy12-without}
\end{subfigure}\hfill
\begin{subfigure}[b]{0.48\linewidth}
\centering
\includegraphics[width=\linewidth]{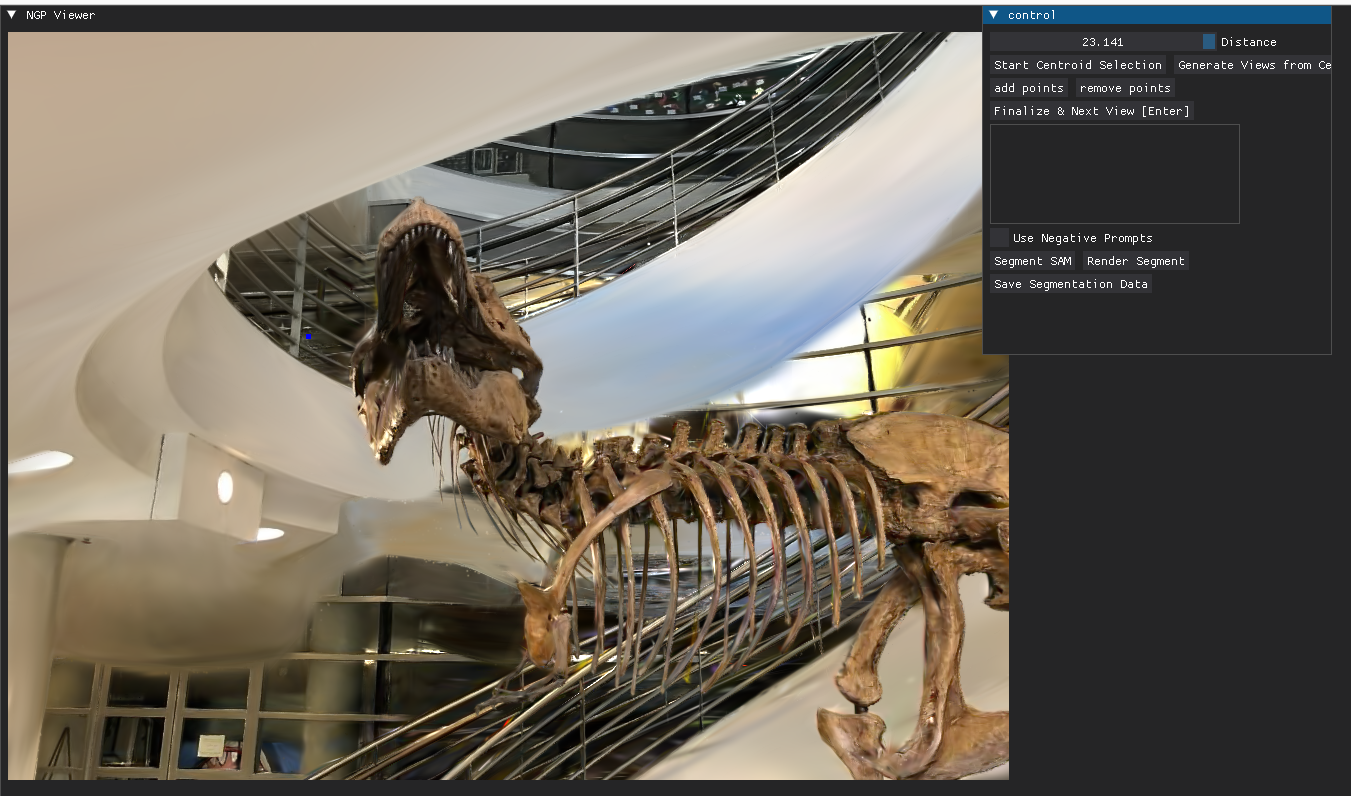}
\caption{Occlusion-aware radius ($r_k{=}23.14$): unobstructed view of the target.}
\label{fig:strategy12-with}
\end{subfigure}
\caption{Effect of the centroid-view refinement (Algorithm~\ref{alg:centroid-gs}). The dynamic occlusion test pulls the orbital radius in until the target is no
longer dominated by foreground occluders.}
\label{fig:strategy12}
\end{figure}

\subparagraph{Strategy 2: Occlusion-aware radius refinement.}
A persistent failure mode is that $r_\text{init}$ can place the centroid camera \emph{inside} a foreground floater, so the rendered view is dominated by occluders rather than the target as shown in Figure~\ref{fig:strategy12}. We probe a sequence of $K_r$ decreasing radii $\{r_k\}$ stepping linearly from $r_\text{init}$ down to $0.3\,r_\text{init}$, rasterize each candidate at low resolution so the render costs $\sim 10$ ms, and accept the \emph{largest} radius whose central depth patch is not dominated by occluders. The key design choice is a \emph{dynamic} occlusion threshold tied to the actual target geometry rather than to the camera radius: projecting the cached neighborhood $\{\mu_i\}_{i\in\mathcal{N}}$ into the probed camera and taking the nearest positive camera-$Z$ value $z_\text{near}$,
\begin{equation}
\tau_\text{occ}(r_k) = 0.85\,z_\text{near},
\label{eq:tauocc}
\end{equation}
(falling back to $0.7\,r_k$ only when the projection is empty), the $0.85$ factor leaves a $15\%$ safety band so grazing target surfels are not misread as
occluders. Over a central patch $\mathcal{P}$ of fractional size $f_\text{fill}$,
the occluded fraction is
\begin{equation}
\mathcal{O}(r_k) = \frac{|\{(x,y)\in\mathcal{P}:\,D_\text{med}(x,y)<\tau_\text{occ}(r_k)\}|}{|\mathcal{P}|},
\label{eq:occfrac}
\end{equation}
and we accept the first $r_k$ (in decreasing order) with
$\mathcal{O}(r_k)<\tau_\mathcal{O}=0.15$, falling back to the most zoomed-in radius if none qualifies. Scanning from large to small deliberately yields the
most \emph{generous} unobstructed framing, preserving the largest plausible context around the click. Figure~\ref{fig:strategy12-with} shows the effect of the refined radius, which pulls the camera to a vantage that looks past the occluders onto the target.

\begin{algorithm}[t]
\caption{2DGS Centroid-View Generation (per click)}
\label{alg:centroid-gs}
\begin{algorithmic}[1]
\Require anchor camera $\mC_\text{anchor}$, user prompt $\vu$, Gaussians $\mathcal{G}$, $\texttt{render}(w2c,w',h')$, $p_f{=}0.25$, $f_\text{fill}{=}0.5$, $\alpha_\text{min}{=}0.1$, probes $K_r{=}5$, $\tau_O{=}0.15$
\State $z\leftarrow \mD_\text{med}(\vu)$;\quad
       $\vx_w\leftarrow\texttt{BackProject}(\vu,z,\mC_\text{anchor})$
\State \emph{// Strategy 1: physical neighborhood}
\State $R_\text{phys}\leftarrow p_f\,z\,\tan(\mathrm{FOV}_x/2)$
\State $\mathcal{N}\leftarrow\{i:\|\mu_i-\vx_w\|<R_\text{phys}\wedge\alpha_i>\alpha_\text{min}\}$
\Comment{Equation~\ref{eq:neigh}}
\If{$|\mathcal{N}|<5$}\Comment{sparse fallback}
   \State $r_\text{init}\leftarrow z\,p_f$;\quad $B_\text{anchor}\leftarrow\varnothing$
\Else
   \State $\mathbf{r}_i^\perp\leftarrow\mathbf{r}_i-(\mathbf{r}_i\!\cdot\!\hat{\mathbf{v}})\hat{\mathbf{v}}$;\quad
          $R_\text{lat}\leftarrow\mathcal{P}_{90}(\{\|\mathbf{r}_i^\perp\|\})$
   \State $r_\text{init}\leftarrow R_\text{lat}/(f_\text{fill}\tan(\mathrm{FOV}_x/2))$
          \Comment{Equation~\ref{eq:Rinit}}
   \State $\sB_\text{anchor}\leftarrow$ projected bounding box of $\{\mu_i\}_{i\in\mathcal{N}}$ on anchor view
\EndIf
\State \emph{// Strategy 2: occlusion-aware refinement}
\State $(w',h')\leftarrow(\text{max}(\sigma w, 64), \text{max}(\sigma h, 64))$
\Comment{Low resolution image rendering}
\State $\mP\leftarrow$ central patch of size $f_\text{fill}$
\State $\{r_k\}\leftarrow$ linspace$(r_\text{init},0.3\,r_\text{init},K_r)$;\quad
       $r_\star\leftarrow r_{K_r-1}$
\For{$k=0,\dots,K_r-1$}\Comment{largest$\to$smallest}
   \State $w2c_k\leftarrow\texttt{BuildPose}(\mC_\text{anchor},\vx_w,r_k)$
   \State $\mD\leftarrow\texttt{render}(w2c_k,\sigma w,\sigma h)[\,\cdot,\cdot,\mD_\text{med}]$
     \State $\sD_\mP \leftarrow \{\mD_\text{med}(x,y)\in\mP \;|\; \mD_\text{med}(x,y)>0.01\}$
     \If{$|\sD_\mP|=0$} 
        \State \textbf{continue} 
     \EndIf
     \State $\mathbf{p}_i^{cam} \leftarrow \mR_{w2c}\mu_i+\vt_{w2c}, \quad \forall i\in\mathcal{N}$ 
     \State $\sZ \leftarrow \{(\vp_i^{cam})_z \;|\; (\vp_i^{cam})_z>0\}$

   \If{$|\sZ|>0$}
        \State $z_\text{near} \leftarrow \min(\sZ)$ \quad
        \State $\tau_\text{occ}\leftarrow 0.85\,z_\text{near}$
        \Comment{Equation.~\ref{eq:tauocc}}
    \Else
        \State $\tau_\text{occ} \leftarrow 0.7\,r_k$
    \EndIf
    
    \State $\sO(r_k)\leftarrow$ occluded fraction over central patch
    \State $\sO(r_k) \leftarrow \dfrac{ |\{d\in\sD_\mP:d<\tau_\text{occ}\}| }{ |\sD_\mP| }$ \Comment{Equation.~\ref{eq:occfrac}}
   \If{$\sO(r_k)<\tau_O$}
       \State $r_\star\leftarrow r_k$; \textbf{break}\Comment{largest unoccluded radius}
   \EndIf
\EndFor
\State \Return centroid camera $C_\star= \texttt{BuildPose} (\mC_\text{anchor}, \vx_w, r_\star)$, $\sB_\text{anchor}$

\end{algorithmic}
\end{algorithm}

% ================================================================
\section{Additional Mathematical Formulation}
\label{app:math-pseudocode}
This section provides the analytical form of the depth-gradient factor $\pi_\text{depth}$, the only component of the fusion kernel that was deferred from the main paper.

% ----------------------------------------------------------------
\subsection{Depth gradient factor \texorpdfstring{$\pi_{\text{depth}}$}{pi}}
\label{app:depth-gradient}

The depth-gradient factor adapts the distance tolerance at object boundaries. It relaxes the tolerance in flat areas and tightens the tolerance at boundary regions where there is a sharp depth discontinuity.

For each pixel $(x_p, y_p)$, we estimate a local depth gradient using central differences on the refined depth range $(\mD_{\text{min}}(x_p, y_p), \mD_{\text{max}}(x_p, y_p))$. Let,
$
d(x,y) = \frac{\mD_{\text{min}}(x_p,y_p) + \mD_{\text{max}}(x_p,y_p)}{2}
$
be the center depth. The horizontal and vertical finite differences are:

$
g_x = d(x_p+1, y_p) - d(x_p - 1, y_p),
$

$
g_y = d(x_p, y_p + 1) - d(x_p, y_p - 1).
$

We compute the squared gradient magnitude:

$
G = g_x^2 + g_y^2,
$

and define the depth-gradient factor as an exponential decay:
$
g_{\text{depth}}(x,y) = \exp\!\bigl(-k\,G\bigr),\qquad k>0.
$
Flat regions yield $G \approx 0$ and $g_{\text{depth}}\approx 1$, while sharp discontinuities produce a large $G$ and a small decay factor.

\paragraph{2DGS instantiation.}
The 2DGS fusion kernel reuses the same pixel-wise depth-gradient factor, $\pi_\text{depth}$, as the NeRF kernel. The depth source changes, central differences are taken on the $(\mD_\text{min}, \mD_\text{max})$ buffers produced by the 2DGS rasterizer rather than by NeRF ray marching.

% ================================================================
\section{Additional GS Experiments}
\label{app:gs-additional}
\label{sec:exp-sensitivity}
We provide the render-resolution and threshold-sensitivity analysis for the Gaussian-Splatting methods, which the main paper (Section~\ref{sec:exp-quant}) summarizes.

\paragraph{Background.}                                                                                                          
``Resolution'' here refers to the \emph{image render resolution} at which the 2D mask is rasterized. DivAS-GS derives its occupancy structure adaptively from the scene's Gaussian statistics mentioned at ~\ref{para:occupancy_gs} and exposes no manually tuned voxel resolution. SAGA and SA3D-GS form a mask by assigning each Gaussian a scalar affinity to the query feature (SAGA) or an optimized mask score (SA3D-GS), respectively, and keeping the primitives whose values exceed a threshold $\tau$. The foreground is therefore a hard, \emph{per-primitive} decision in which a single $\tau$ must both reject background primitives and retain thin foreground primitives. Argmax-based methods (Gaussian Grouping, SANeRF-HQ) instead assign a label to every primitive or pixel and are not subject to this trade-off. We sweep render resolution and $\tau$ to characterize the threshold-based methods. DivAS-GS, which never thresholds a per-primitive scalar, is reported at its single canonical operating point.

% ----------------------------------------------------------------
\begin{table}[t]
\centering
\caption{Resolution and threshold sensitivity across methods. We report \emph{per-dataset mean} mIoU/IoU for each method. ``Full Resolution'' renders the mask \emph{directly} at native GT resolution from the scene representation. Per dataset column, the best value is \textbf{bold} and the second-best is
\underline{underlined}.}
\label{tab:resolution-sensitivity}
\small
\begin{tabular}{llcc}
\toprule
Method & Configuration & Mip-NeRF~360$^\circ$ & LLFF \\
\midrule
DivAS-GS (ours) & Full Resolution, $\tau{=}0.1$ \emph{(primary)} & 0.9476 & \underline{0.9230} \\
DivAS-GS (ours) & Training Resolution, $\tau{=}0.1$              & 0.9448 & \textbf{0.9238} \\
\midrule
SAGA            & Full Resolution, $\tau{=}0.5$ \emph{(primary)} & \underline{0.9533} & 0.8739 \\
SAGA            & Training Resolution, $\tau{=}0.5$              & \textbf{0.9594} & 0.9151 \\
SAGA            & Full Resolution, $\tau{=}0.1$                  & 0.9377 & 0.8925 \\
\midrule
SA3D-GS         & Full Resolution, $\tau{=}0.1$ \emph{(primary)} & 0.6512 & 0.7502 \\
SA3D-GS         & Training Resolution, $\tau{=}0.1$              & 0.6489 & 0.7481 \\
\bottomrule
\end{tabular}
\end{table}

\paragraph{DivAS-GS is resolution-stable at a single threshold.}
\label{para:saga_holes}
Table~\ref{tab:resolution-sensitivity} shows that DivAS-GS IoU changes by at most $0.30$ points between full and training resolution, with the sign inconsistent across benchmarks (full resolution higher on Mip-NeRF 360$^\circ$ by $0.30$ pt while training resolution higher on LLFF by $0.08$ pt). This near-invariance holds under a fixed $\tau{=}0.1$ without any resolution-specific tuning, a property we attribute to the architectural decoupling described below.
For SAGA, $\tau$ must serve two competing roles: (i) rejecting background and (ii) retaining thin foreground, whose covering primitives carry affinities that fall between the two clusters, i.e., foreground/background. On LLFF at $\tau{=}0.5$, the training-resolution score is high ($0.9151$), but the full-resolution score is $0.8739$; lowering $\tau$ to $0.1$ improves full resolution only partially ($0.8925$) by re-admitting culled boundary gaussians, so no single $\tau$ is at once complete and clean. The resulting gaps(holes) in thin geometry are not a 2D sampling effect. They are visible at both full and training resolution (Section~\ref{sec:exp-qualitative}), and reducing $\tau$ recovers them only partially while degrading other regions. Consequently, the configuration that maximizes SAGA's IoU (training resolution, $\tau{=}0.5$) does not correspond to a structurally clean mask. The score and its mask topology disagree.

\paragraph{DivAS-GS decouples geometric separation from binarization.}
DivAS-GS does not threshold a per-primitive scalar. Foreground membership is decided  \emph{upstream} by multi-view voxel-occupancy consensus. While mapping back the voxel to surfel, each surfel multi-view occupancy consensus that falls below the threshold is removed from the foreground set before any mask threshold is applied. The mask threshold $\tau{=}0.1$ then operates exclusively as a binarizer on an already-clean surfel set, free to remain low without re-admitting background. Because membership rests on geometric and cross-view agreement rather than on a single primitive's appearance, thin structures that are occupied and consistently observed are retained at both resolutions. This is why DivAS-GS achieves $0.9230$ LLFF IoU at full resolution, surpassing SAGA's best result across all resolution threshold settings ($0.9151$). Raising $\tau$ from $0.1$ to $0.5$ on DivAS-GS only over-culls thin-structure surfels (reproducing holes), confirming that $0.1$ is the natural operating point for binarization and that background rejection has already been handled. Unlike SAGA, SA3D-GS shows essentially no resolution sensitivity ($\Delta{<}0.3$ pt across all twelve scenes), confirming that the floater artifacts and missing object parts induced by the unbounded mask-score objective are equally visible at both resolutions, leaving a consistent quality floor of $0.65$ on Mip-NeRF~360$^\circ$ and $0.75$ on LLFF regardless of how the mask is measured.

% ================================================================
\section{NeRF Instantiation: Qualitative Results}
\label{app:nerf-additional}

We compare the DivAS (NeRF) instantiation against SA3D and SANeRF-HQ on representative LLFF and Mip-NeRF~$360^\circ$ scenes.

\begin{figure}[H]
    \centering
    \includegraphics[width=0.9\linewidth]{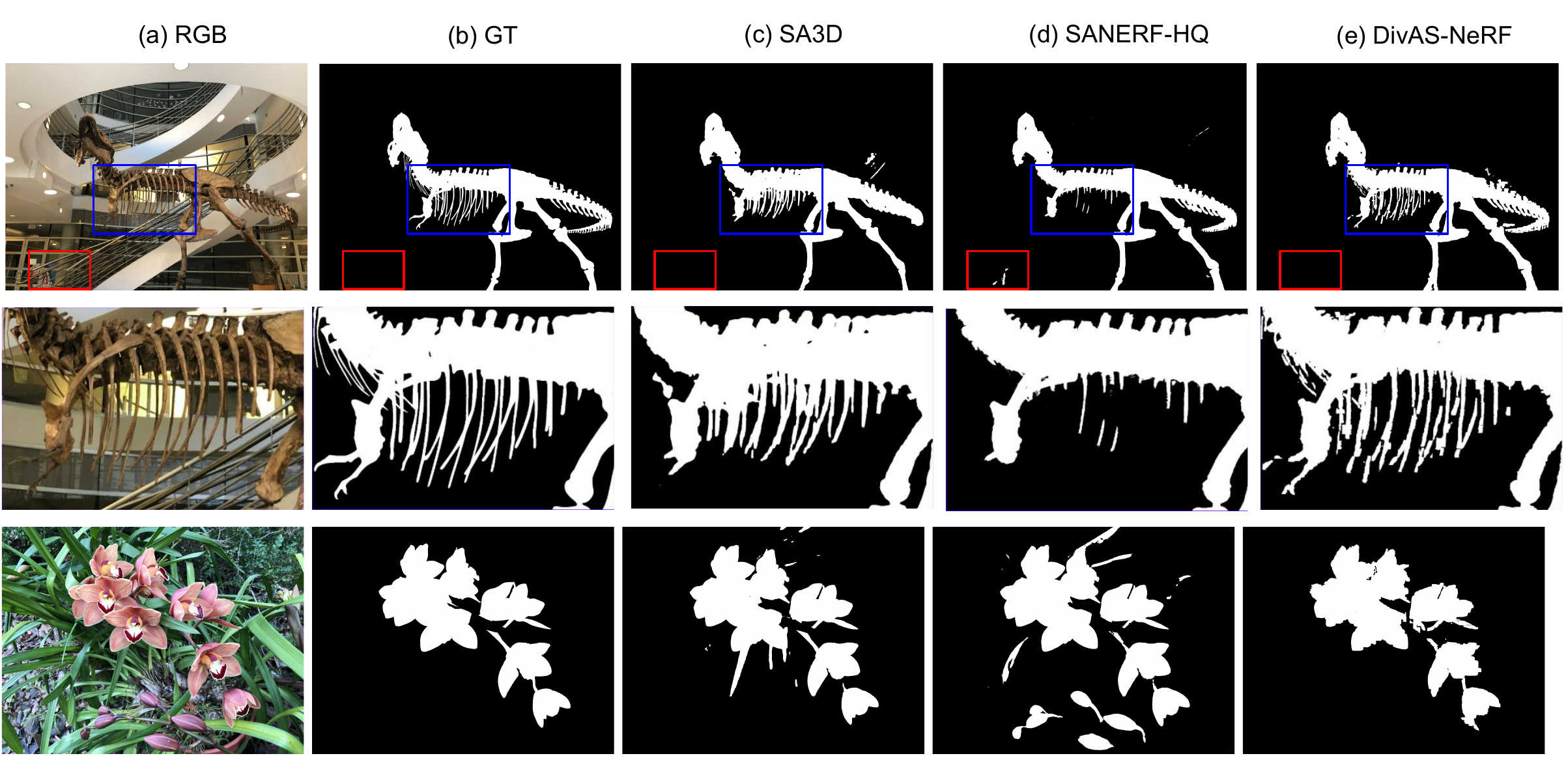}
    \caption{Results of different methods for the NeRF instantiation on LLFF.}
    \label{fig:qualitative_llff}
\end{figure}

As shown in Figure~\ref{fig:qualitative_llff}, we visualize segmentation results for the \textit{Trex} object across different methods in the NeRF backbone. SA3D captures fine structures like ribs but misses global parts such as the tail, and slightly leaks into the background (row $2$). SANeRF-HQ recovers the complete shape but fails on thin features like the ribs and hands. It also exhibits over-segmentation by incorrectly marking background regions with similar appearance, highlighted in the red box (row $1$). This behavior aligns with its tendency toward false positives observed in the \textit{Orchids} scene. Our method preserves both global structure and delicate parts, such as ribs and hands, while avoiding false activations and over-segmentation.

\begin{figure}[H]
    \centering
    \includegraphics[width=0.8\linewidth]{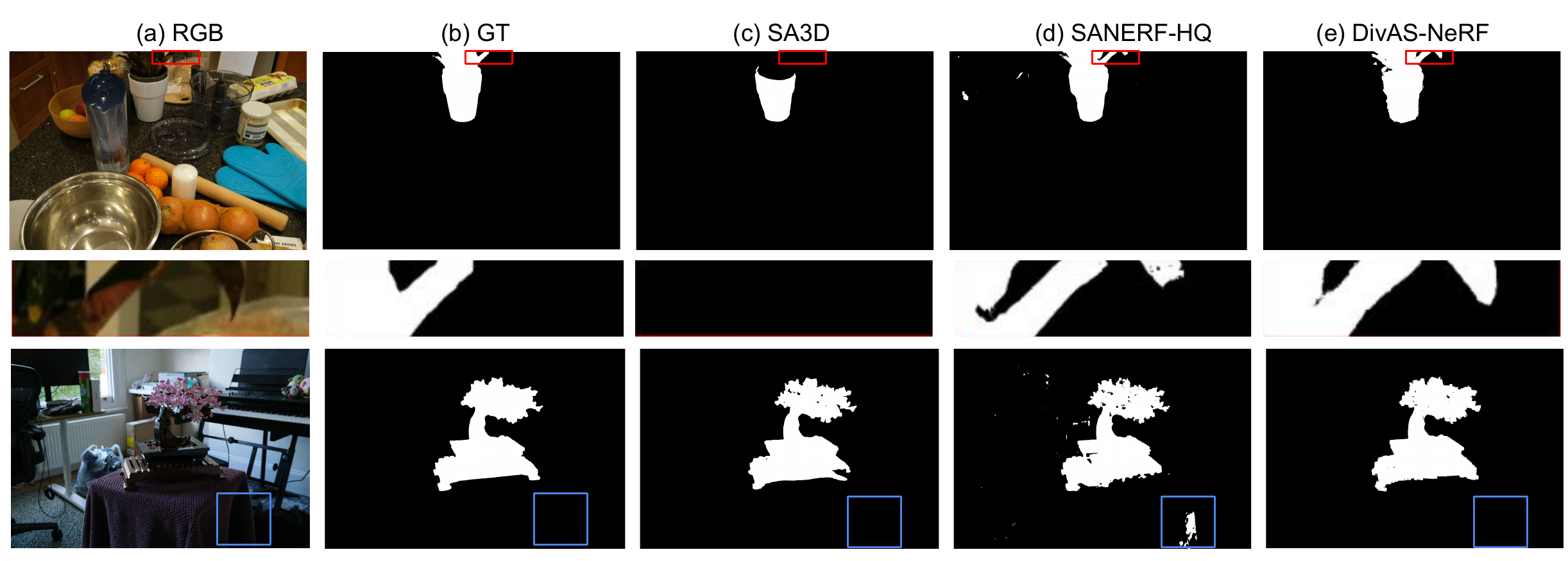}
    \caption{Results of different methods for the NeRF instantiation on Mip-NeRF~$360^\circ$.}
    \label{fig:qualitative_mip}
\end{figure}

In Figure~\ref{fig:qualitative_mip}, we compare methods on the \textit{Counter (flowerpot)} and \textit{Bonsai} scenes from Mip-NeRF~$360^\circ$ in the NeRF backbone. In the first row, the red box marks a region where the pseudo ground truth misses part of the flowerpot's leaf. The zoomed-in crops (row $2$) show that our method better preserves object boundaries and maintains multi-view geometric consistency. SA3D misses occluded plant regions due to its single-view prompting strategy. SANeRF-HQ does not match the complete leaf structure shown in row $2$ compared to our method. In the third row \textit{Bonsai}, SANeRF-HQ again shows over-segmentation, assigning distant background areas as part of the target object (blue box). Our method produces sharper, more localized masks that are consistent with the true object geometry.

% ================================================================
\section{Additional Ablations}
\label{app:additional-ablations}

The ablations in Section~\ref{sec:exp-ablation} cover the sensitivity analysis of the hyperparameters used in NeRF and the corresponding counterpart in 2DGS instantiation. In contrast to the NeRF case, most of the 2DGS defaults are scene-specific. The reason is structural: the voxel size is adaptive per scene, and the surfel primitives are anisotropic and randomly oriented, so a single global value cannot be generalized across scenes. We expose these as user-tunable sliders. The defaults shipped work well across the scenes; the user can expect approximately $\pm 1$-$1.5$ percentage points of IoU around the reported results in Section~\ref{sec:exp-quant} without tuning.

% ----------------------------------------------------------------
\subsection{Cumulative-weight threshold \texorpdfstring{$\tau_\text{cw}$}{tau\_cw} (NeRF)}
\label{app:cumulative-weight-nerf}

\begin{figure}[t]
    \centering
    \includegraphics[width=0.95\linewidth]{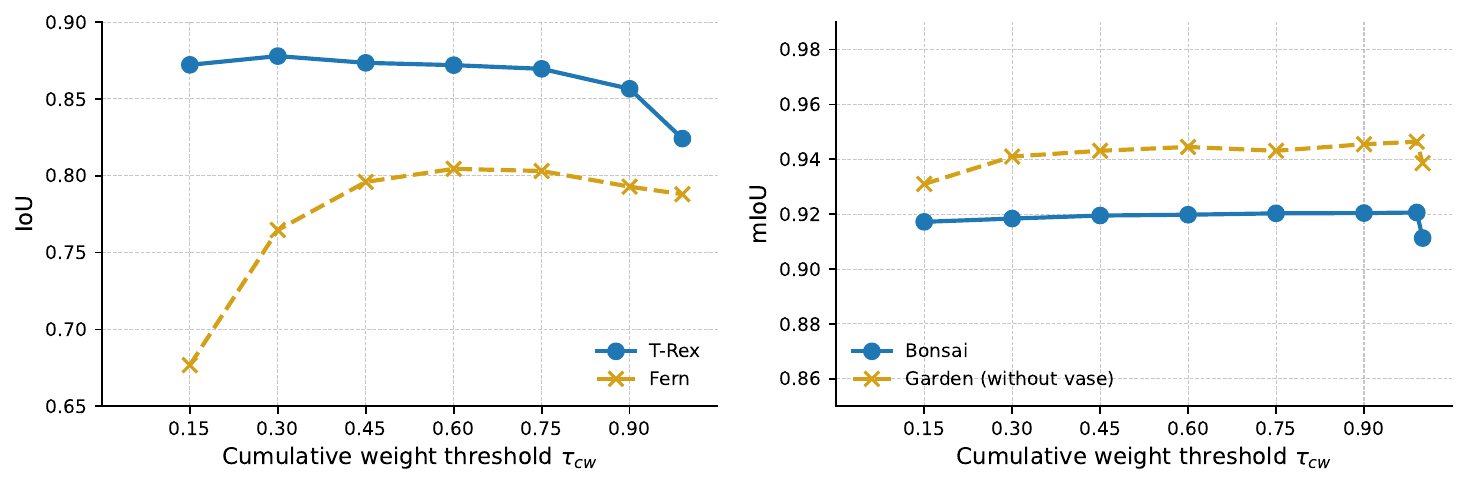}
    \caption{IoU vs.\ cumulative-weight target $\tau_\text{cw}$ for the NeRF instantiation on representative LLFF and Mip-NeRF~$360^\circ$ scenes.}
    \label{fig:ablation_cumulative_weight_nerf}
\end{figure}

The cumulative-weight cutoff $\tau_\text{cw}$ sets the upper bound of the depth range $\mD_\text{max}$ that is fed to the fusion kernel. We sweep it over $\{0.15$-$0.9\}$ on the four representative scenes of Section~\ref{sec:exp-ablation}, namely \textit{Trex} and \textit{Fern} from LLFF and \textit{Bonsai} and \textit{Garden (without vase)} from Mip-NeRF~$360^\circ$. As shown in Figure~\ref{fig:ablation_cumulative_weight_nerf}, the IoU remains stable for $\tau_\text{cw}\!\in[0.3, 0.75]$ and drops sharply at higher values ($\ge 0.9$). At such thresholds, the valid depth range widens until background voxels become activated, reducing segmentation precision. The IoU of Mip-NeRF~$360^\circ$ scenes (\textit{Bonsai}, \textit{Garden}) is nearly insensitive to $\tau_\text{cw}$ because hierarchical sampling places the bulk of density samples near the surface, so the cumulative weight converges quickly regardless of the cutoff. LLFF scenes show more variation because uniform sampling along the ray requires a higher $\tau_\text{cw}$ to exhaust the foreground density and can include background at higher values. Although $\tau_\text{cw}{=}0.6$ gives a marginally higher IoU on \textit{Trex}, qualitative evaluation on \textit{Fern} favors $0.75$ for smoother and more complete masks. We fix $\tau_\text{cw}{=}0.75$ as the NeRF default.

% ----------------------------------------------------------------
\subsection{Base tolerance multiplier \texorpdfstring{$\gamma$}{gamma}}
\label{app:gamma}

\begin{figure}
    \centering
    \includegraphics[width=0.6\linewidth]{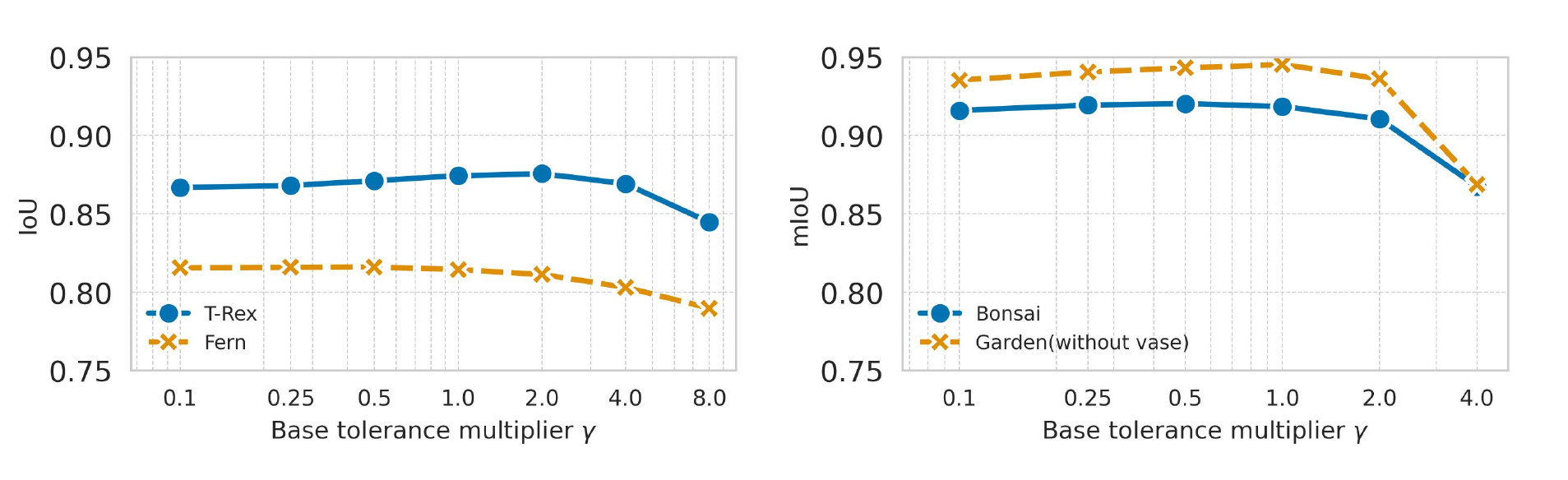}
    \caption{IoU/mIoU vs.\ base tolerance multiplier $\gamma$.}
    \label{fig:base_tolerance_multiplier_plot}
\end{figure}

\begin{figure}
    \centering
    \includegraphics[width=\linewidth]{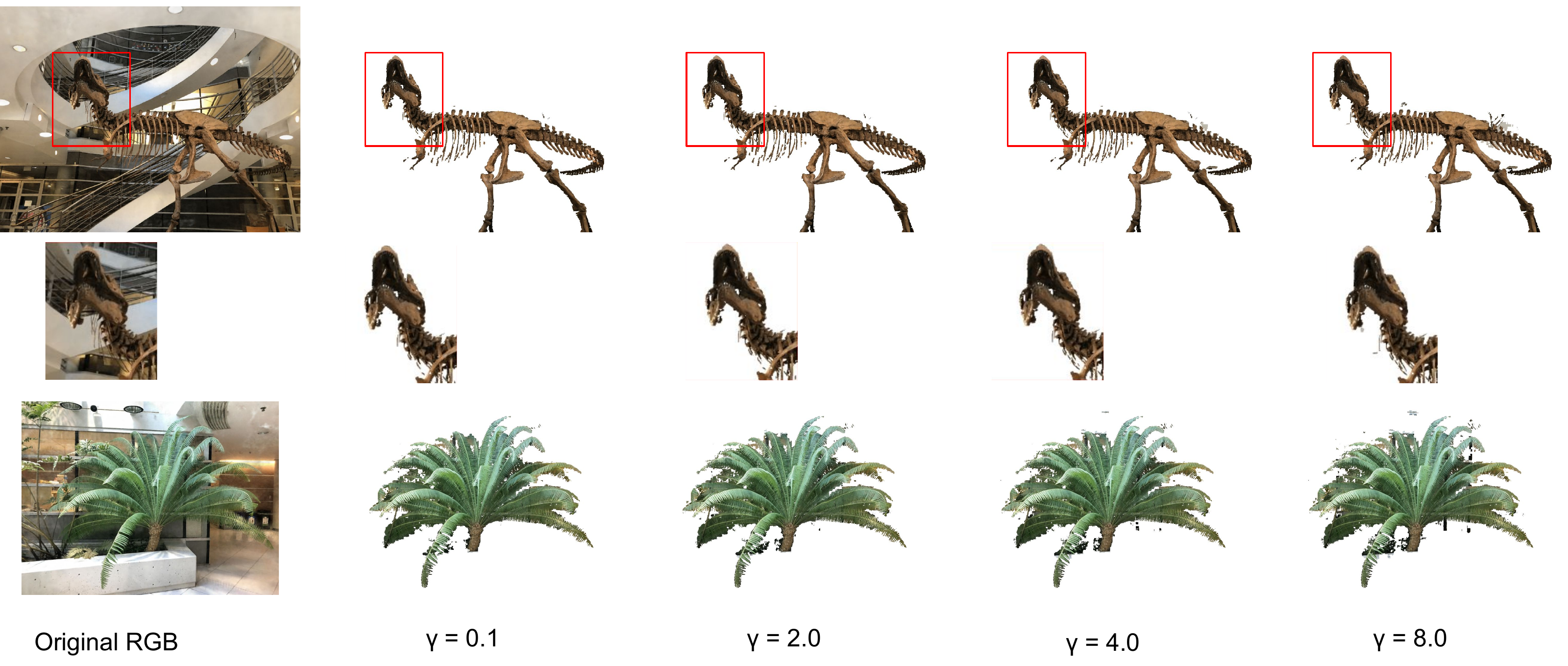}
    \caption{Qualitative ablations for LLFF scenes on the base tolerance multiplier $\gamma$.}
    \label{fig:base_tolerance_multiplier_llff}
\end{figure}

\begin{figure}
    \centering
    \includegraphics[width=\linewidth]{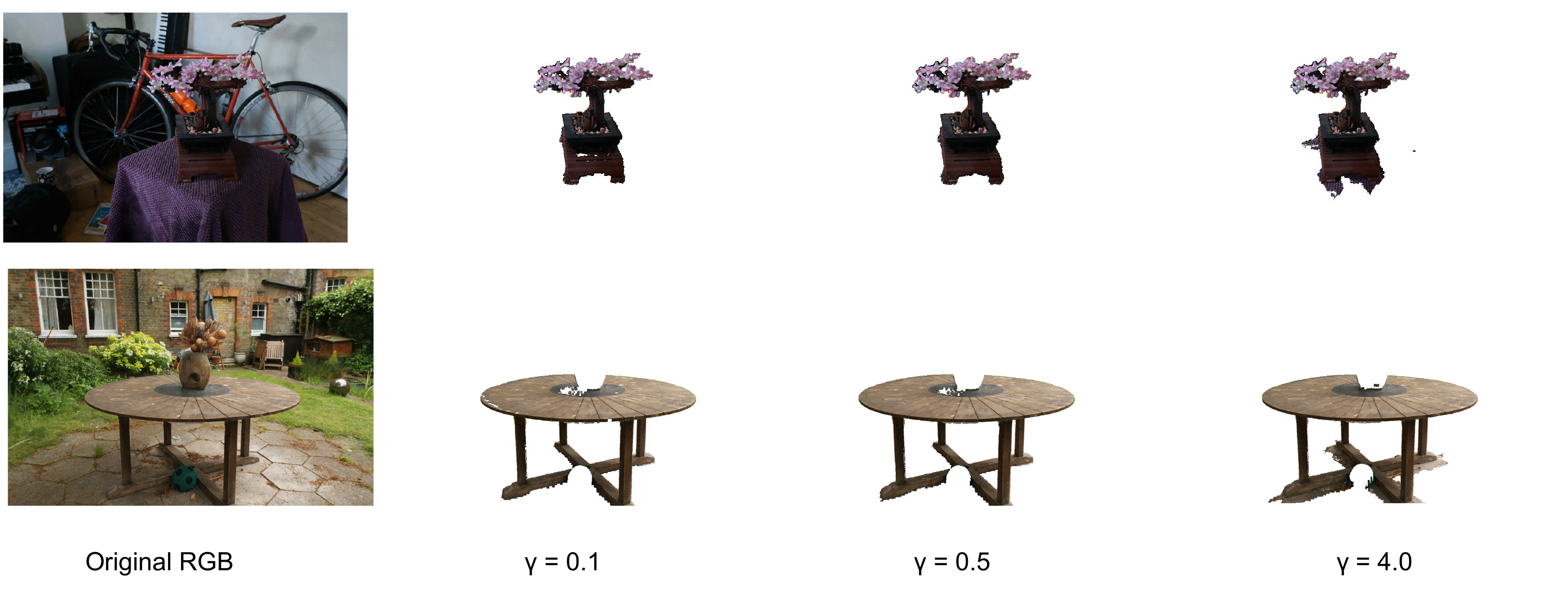}
    \caption{Qualitative ablations for Mip-NeRF $360^\circ$ scenes on the base tolerance multiplier $\gamma$.}
    \label{fig:base_tolerance_multiplier_mip}
\end{figure}

Figure~\ref{fig:base_tolerance_multiplier_plot} reports the effect of the base tolerance multiplier $\gamma$ on IoU for LLFF and mIoU for Mip-NeRF~$360^\circ$ scenes. Across LLFF (Trex and Fern), IoU remains stable for $\gamma \in [0.1, 2.0]$ and shows a mild drop at larger values. However, the qualitative results in Figure~\ref{fig:base_tolerance_multiplier_llff} reveal an important behavior that IoU alone does not capture.

For the Trex scene, $\gamma{=}2.0$ produces a slightly higher IoU than $\gamma{=}4.0$, but it systematically fails to recover thin structures around the jaw, ribs, spine, and near the head boundary of the object. The zoomed-in crops (row $2$) highlight how $\gamma{=}4.0$ restores the head skeleton while still suppressing background noise. Because thin-structure completeness is essential for our interactive segmentation setting, we adopt $\gamma{=}4.0$ for LLFF scenes.

For Mip-NeRF~$360^\circ$ (Bonsai and Garden), the trend differs. mIoU is nearly flat in the interval $\gamma \in [0.1, 2.0]$, with a slight peak at $\gamma{=}0.5$, and then sharply degrades at $\gamma{=}4.0$ (Figure~\ref{fig:base_tolerance_multiplier_plot}). The qualitative comparisons in Figure~\ref{fig:base_tolerance_multiplier_mip} support this, $\gamma{=}0.1$ under-segments thin boundaries, whereas $\gamma{=}4.0$ introduces strong background leakage. $\gamma{=}0.5$ provides the best trade-off, preserving edge detail without introducing artifacts.

Overall, while IoU curves are relatively flat across a wide range of $\gamma$, the qualitative sensitivity to thin structures and background leakage motivates our scene-dependent defaults ($\gamma{=}4.0$ for LLFF, $\gamma{=}0.5$ for Mip-NeRF~$360^\circ$).

\paragraph{2DGS instantiation.}
The 2DGS analog is scene-specific. This is because of the reason stated in \ref{app:additional-ablations}. We therefore expose this parameter as a user-tunable slider with a per-scene default.

% ----------------------------------------------------------------
\subsection{Low-sensitivity kernel-parameter sweeps}
\label{app:compact-sweeps}

The remaining NeRF fusion-kernel parameters change the score only marginally across their full sweeps. To keep the presentation compact, we stack the LLFF and Mip-NeRF~$360^\circ$ measurements side by side and group the four parameters into two paragraphs rather than separate subsections.

\paragraph{Per-sample bonus $\beta$ and maximum tolerance bonus $b_{\max}$.}
\begin{table}[t]
\centering
\small
\caption{Effect of the per-sample bonus $\beta$ and the maximum tolerance bonus $b_{\max}$ on IoU (LLFF: Trex, Fern) and mIoU (Mip-NeRF~$360^\circ$: Bonsai, Garden), NeRF instantiation.}
\label{tab:beta_bmax}
\resizebox{\linewidth}{!}{%
\begin{tabular}{ccc|ccc||ccc|ccc}
\toprule
\multicolumn{6}{c||}{$\beta$} & \multicolumn{6}{c}{$b_{\max}$} \\
\cmidrule(lr){1-6}\cmidrule(lr){7-12}
\multicolumn{3}{c|}{LLFF} & \multicolumn{3}{c||}{Mip-NeRF~$360^\circ$} & \multicolumn{3}{c|}{LLFF} & \multicolumn{3}{c}{Mip-NeRF~$360^\circ$} \\
$\beta$ & Trex & Fern & $\beta$ & Bonsai & Garden & $b_{\max}$ & Trex & Fern & $b_{\max}$ & Bonsai & Garden \\
\midrule
0.01 & 0.877 & 0.807 & 0.01 & 0.919 & 0.935 & 1  & 0.876 & 0.808 & 0.10 & 0.918 & 0.933 \\
0.05 & 0.876 & 0.812 & 0.02 & 0.919 & 0.938 & 5  & 0.870 & 0.807 & 0.25 & 0.919 & 0.937 \\
0.10 & 0.875 & 0.810 & 0.05 & 0.920 & 0.943 & 10 & 0.869 & 0.803 & 0.50 & 0.920 & 0.939 \\
0.20 & 0.869 & 0.803 & 0.10 & 0.920 & 0.943 & 20 & 0.869 & 0.802 & 1    & 0.920 & 0.943 \\
0.30 & 0.865 & 0.800 & 0.20 & 0.919 & 0.943 & 40 & 0.869 & 0.802 & 2    & 0.920 & 0.943 \\
0.40 & 0.860 & 0.799 & 0.40 & 0.919 & 0.944 &    &       &       & 4    & 0.920 & 0.943 \\
\bottomrule
\end{tabular}}
\end{table}
Table~\ref{tab:beta_bmax} reports the effect of the per-sample bonus $\beta$, which controls the depth tolerance based on the contribution of the number of samples involved in ray marching for a pixel. Across all scenes, the IoU variation with respect to $\beta$ is small. Extremely low values, such as $\beta{=}0.01$, yield slightly higher IoU in Trex and Fern, but the gains are numerically marginal and do not accurately reflect the actual segmentation quality. Qualitatively, very small $\beta$ behaves similarly to a low $\gamma$ value (Section~\ref{app:gamma}), leading to underestimation of thin structures and missing fine skeletal parts even when IoU appears high. Larger values ($\beta \ge 0.2$) instead provide a more stable tolerance growth and consistently recover fine structures without introducing background leakage, so we adopt $\beta{=}0.2$ as a balanced default for the NeRF instantiation. The same table reports the maximum accumulated tolerance bonus $b_{\max}$, which caps how much $\beta$ can relax the geometric-consistency threshold. The IoU again changes only marginally as $b_{\max}$ varies over a wide range. Extremely small values (e.g., $b_{\max}{=}1$) slightly increase IoU on LLFF for the same thin-structure reason discussed for $\beta$, while larger values saturate quickly and behave indistinguishably beyond $b_{\max}{\ge}10$. We therefore adopt $b_{\max}{=}10$ for LLFF and $b_{\max}{=}1$ for Mip-NeRF~$360^\circ$.

\paragraph{2DGS instantiation.}
The role of the per-sample bonus is inverted in the 2DGS kernel and is therefore not directly comparable to the NeRF parameter swept in Table~\ref{tab:beta_bmax}. In NeRF, a high $n_\text{samp}$ on a pixel indicates dense ray sampling and the bonus \emph{relaxes} the tolerance to admit the additional valid samples. In 2DGS, a high $n_\text{samp}$ within the footprint indicates many overlapping surfels along the ray, which signals that the rasterizer-reported depth is noisy, so the kernel reuses the same per-sample statistic to \emph{tighten} rather than relax the tolerance.

\paragraph{Depth range factor $\lambda_\text{range}$ and density threshold $\rho_\text{thresh}$.}
\label{app:rho-thresh}
\begin{table}[t]
\centering
\small
\caption{Effect of the depth range factor $\lambda_\text{range}$ and the density threshold $\rho_\text{thresh}$ on IoU (LLFF: Trex, Fern) and mIoU (Mip-NeRF~$360^\circ$: Bonsai, Garden), NeRF instantiation.}
\label{tab:lambda_rho}
\resizebox{\linewidth}{!}{%
\begin{tabular}{ccc|ccc||c|cc|cc}
\toprule
\multicolumn{6}{c||}{$\lambda_\text{range}$} & \multicolumn{5}{c}{$\rho_\text{thresh}$} \\
\cmidrule(lr){1-6}\cmidrule(lr){7-11}
\multicolumn{3}{c|}{LLFF} & \multicolumn{3}{c||}{Mip-NeRF~$360^\circ$} & & \multicolumn{2}{c|}{LLFF} & \multicolumn{2}{c}{Mip-NeRF~$360^\circ$} \\
$\lambda_\text{range}$ & Trex & Fern & $\lambda_\text{range}$ & Bonsai & Garden & $\rho_\text{thresh}$ & Trex & Fern & Bonsai & Garden \\
\midrule
0.01 & 0.869 & 0.803 & 0.1 & 0.921 & 0.942 & 1  & 0.852 & 0.767 & 0.920 & 0.946 \\
0.02 & 0.869 & 0.804 & 0.5 & 0.920 & 0.943 & 5  & 0.866 & 0.797 & 0.920 & 0.943 \\
0.05 & 0.869 & 0.803 & 1   & 0.920 & 0.943 & 10 & 0.869 & 0.803 & 0.920 & 0.939 \\
0.10 & 0.866 & 0.800 & 2   & 0.920 & 0.943 & 20 & 0.871 & 0.787 & 0.920 & 0.933 \\
0.20 & 0.863 & 0.797 & 5   & 0.920 & 0.943 & 50 & 0.871 & 0.637 & 0.921 & 0.925 \\
0.40 & 0.862 & 0.797 & 10  & 0.920 & 0.943 &    &       &       &       &       \\
0.80 & 0.862 & 0.797 &     &       &       &    &       &       &       &       \\
\bottomrule
\end{tabular}}
\end{table}

\begin{figure}[t]
    \centering
    \includegraphics[width=0.9\linewidth]{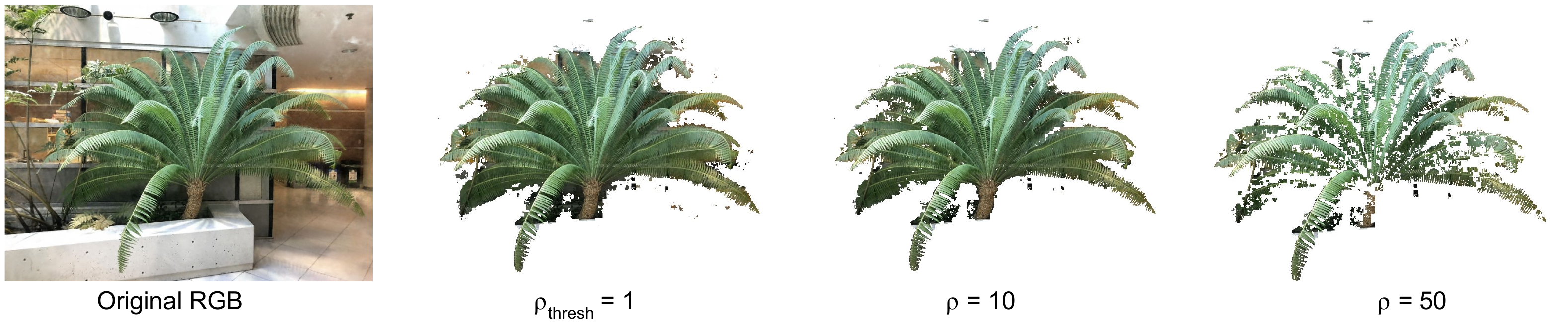}
    \caption{Qualitative ablation for Fern on the density threshold $\rho_\text{thresh}$ (NeRF instantiation).}
    \label{fig:density_threshold}
\end{figure}

$\lambda_\text{range}$ regulates the depth tolerance used in the geometric-consistency check. Its effect is most visible near foreground-background boundaries, where depth discontinuities may occur. In LLFF, uniform ray sampling produces larger depth gaps, requiring a strict tolerance ($\lambda_\text{range}{=}0.01$-$0.05$). In contrast, Mip-NeRF~$360^\circ$ uses hierarchical sampling, concentrating samples near surfaces and permitting a more relaxed setting ($\lambda_\text{range}{=}5.0$). As Table~\ref{tab:lambda_rho} shows, IoU varies only marginally with $\lambda_\text{range}$, indicating that the parameter plays a stabilizing but low-impact role. The density threshold $\rho_\text{thresh}$ controls which voxels are considered valid by pruning low-density background regions. Since NeRF scenes typically occupy only $1$-$3\%$ of the volume, a balanced threshold is essential. Low values ($\rho{=}1$) admit background leakage, while high values (e.g., $\rho{=}50$) begin to suppress thin or semi-transparent object parts, as the Fern drop to $0.637$ at $\rho{=}50$ in Table~\ref{tab:lambda_rho} and the qualitative comparison in Figure~\ref{fig:density_threshold} make clear. LLFF scenes therefore benefit from a moderately strict $\rho_\text{thresh}{=}10$. Mip-NeRF~$360^\circ$ employs a higher-resolution voxel grid ($256^3$) together with hierarchical sampling, producing a sharper, more localized density field, so the effect of $\rho_\text{thresh}$ is visually subtle across a wide range of values. This aligns with common NeRF practice, for example, Instant-NGP~\citep{muller2022instantngp} uses $\rho{=}10$ for a coarser $128^3$ grid, where higher grid resolutions naturally require lower pruning thresholds. We adopt $\rho_\text{thresh}{=}5$ as the default for Mip-NeRF~$360^\circ$, which best preserves bonsai and garden structures while suppressing background artifacts.

\paragraph{2DGS instantiation.}
The 2DGS kernel disables $\lambda_\text{range}$. The reason is that the 2DGS rasterizer reports a single planar median-depth value $d_\text{med}$ around which all surfels tightly pack on the surface of the hollow object, whereas NeRF learns a wide band $[d_\text{min}, d_\text{max}]$ that includes the object's internal body. The spatial tolerance is therefore driven entirely by the voxel size $h_v$ multiplied by $\gamma$ and the inverted bonus, and no range-derived slack is needed. The interpretation of voxel occupancy also differs substantially from the NeRF case. Because the 2DGS occupancy grid is constructed by scattering surfel opacity into Morton-indexed voxels, each occupied voxel often corresponds directly to a small set of surface surfels rather than to a volumetric density. We accumulate the opacity contributions of all surfels mapped to a voxel and observe that the median non-empty voxel occupancy is approximately $1.0$ across the evaluated scenes, so we adopt a permissive density floor of $\rho_\text{thresh}{=}0.5$, corresponding to half of the typical occupied-voxel support. Unlike the NeRF case, where density thresholding primarily removes volumetric background regions, overly aggressive pruning in 2DGS can permanently eliminate surface surfels because no neighboring occupied voxels exist behind the visible shell to recover the geometry. For this reason, both the thick and thin paths share the same density floor, which is fixed across all scenes and datasets.

% ----------------------------------------------------------------
\subsection{Thin percent cover threshold \texorpdfstring{$\rho_{\text{percent\,cover\,thresh}}$}{rho\_cover}}
\label{app:rho-cover}

\begin{table}[H]
\centering
\begin{tabular}{c c c}
\toprule
$\mathbf{\rho_\text{cover}}$ & Trex & Fern \\
\midrule
0.05 & 0.862 & 0.792 \\
0.10 & 0.868 & 0.797 \\
0.20 & 0.869 & 0.803 \\
0.40 & 0.858 & 0.811 \\
0.50 & 0.852 & 0.813 \\
\bottomrule
\end{tabular}
\caption{Ablation on the percentage cover threshold $\rho_\text{percent\,cover\,thresh}$ (shorthand $\rho_\text{cover}$) on IoU for LLFF scenes (NeRF instantiation).}
\label{tab:rho_cover_llff}
\end{table}

\begin{table}[H]
\centering
\begin{tabular}{c c c}
\toprule
$\mathbf{\rho_\text{cover}}$ & Bonsai & Garden (without vase) \\
\midrule
0.1 & 0.920 & 0.944 \\
0.2 & 0.921 & 0.944 \\
0.3 & 0.920 & 0.943 \\
0.4 & 0.920 & 0.943 \\
0.5 & 0.918 & 0.944 \\
\bottomrule
\end{tabular}
\caption{Ablation on the percentage cover threshold $\rho_\text{percent\,cover\,thresh}$ (shorthand $\rho_\text{cover}$) on mIoU for Mip-NeRF~$360^\circ$ scenes (NeRF instantiation).}
\label{tab:rho_cover_mip}
\end{table}

\begin{figure}
    \centering
    \includegraphics[width=0.9\linewidth]{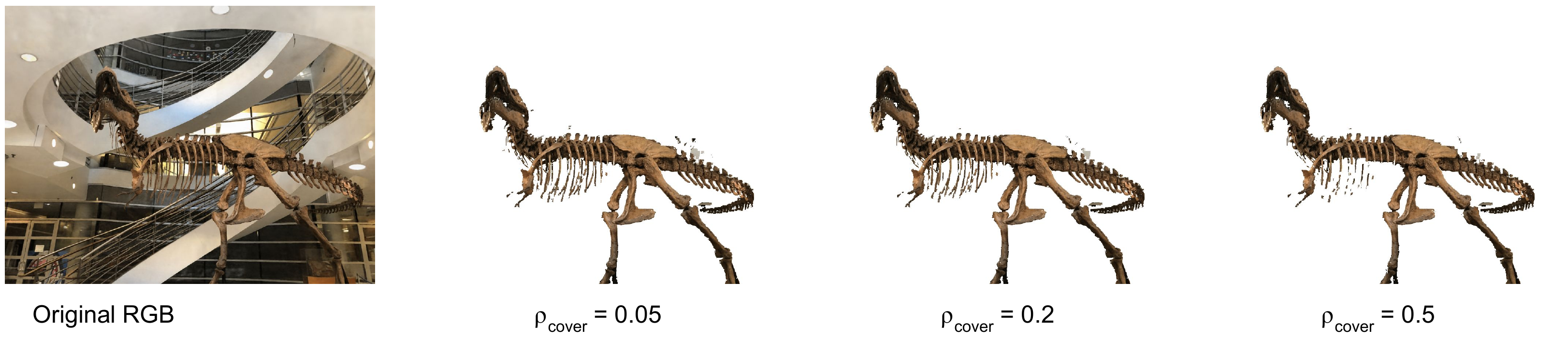}
    \caption{Qualitative ablation for Trex on the thin percent cover threshold (shorthand $\rho_\text{cover}$).}
    \label{fig:thin_percent_cover}
\end{figure}

Tables~\ref{tab:rho_cover_llff} and~\ref{tab:rho_cover_mip} report the effect of the thin percent-cover threshold $\rho_\text{percent\,cover\,thresh}$ (denoted $\rho_\text{cover}$). This parameter determines how many masked pixels must project into a voxel before it is considered a valid thin-structure candidate.

In LLFF scenes, each voxel aggregates contributions from multiple refined SAM mask pixels. While the thick-structure path only checks alignment at the voxel center, many boundary voxels receive partial coverage due to imperfect pixel-voxel alignment or thin geometric structures. The thin-structure path uses $\rho_\text{cover}$ to detect such cases. Higher thresholds favor conservative labeling and prevent leakage, whereas lower thresholds may trigger over-segmentation.

As seen in Table~\ref{tab:rho_cover_llff}, LLFF is moderately sensitive to this parameter. Qualitative examples in Figure~\ref{fig:thin_percent_cover} illustrate the trade-off, small values (e.g., $\rho_\text{cover}{=}0.05$) introduce noticeable background leakage, while large values (e.g., $\rho_\text{cover}{=}0.50$) miss thin structures. We therefore adopt $\rho_\text{cover}{=}0.20$ for LLFF.

In contrast, Mip-NeRF~$360^\circ$ is substantially more robust (Table~\ref{tab:rho_cover_mip}). The $256^3$ voxel grid provides twice the resolution of LLFF, reducing ambiguity in partial coverage. A slightly larger threshold ($\rho_\text{cover}{=}0.30$) works well in practice and avoids unnecessary voxel marking while keeping computation minimal.

\paragraph{2DGS instantiation.}
The 2DGS analog is scene-specific. This is because of the reason stated in \ref{app:additional-ablations}. It plays the same role as accepting a thin-structure voxel only when a sufficient fraction of its projected footprint is consistent with the foreground mask and depth.

% ----------------------------------------------------------------
\subsection{Thin density threshold \texorpdfstring{$\rho_{\text{thin\_thresh}}$}{rho\_thin}}
\label{app:rho-thin}

\begin{table}
\centering
\begin{tabular}{c c c}
\toprule
$\mathbf{\rho_\text{thin}}$ & Trex & Fern \\
\midrule
5 & 0.840 & 0.779 \\
10 & 0.842 & 0.783 \\
30 & 0.860 & 0.799 \\
50 & 0.869 & 0.803 \\
100 & 0.868 & 0.808 \\
\bottomrule
\end{tabular}
\caption{Ablation on the thin-structure density threshold $\rho_\text{thin\_thresh}$ (shorthand $\rho_\text{thin}$) on IoU for LLFF scenes (NeRF instantiation).}
\label{tab:rho_thin_llff}
\end{table}

\begin{table}
\centering
\begin{tabular}{c c c}
\toprule
$\mathbf{\rho_\text{thin}}$ & Bonsai & Garden (without vase)  \\
\midrule
10 & 0.920 & 0.945 \\
30 & 0.920 & 0.943 \\
50 & 0.921 & 0.943 \\
100 & 0.921 & 0.943 \\
\bottomrule
\end{tabular}
\caption{Ablation on the thin-structure density threshold $\rho_\text{thin\,thresh}$ (shorthand $\rho_\text{thin}$) on mIoU for Mip-NeRF~$360^\circ$ scenes (NeRF instantiation).}
\label{tab:rho_thin_mip}
\end{table}

\begin{figure}
    \centering
    \includegraphics[width=0.9\linewidth]{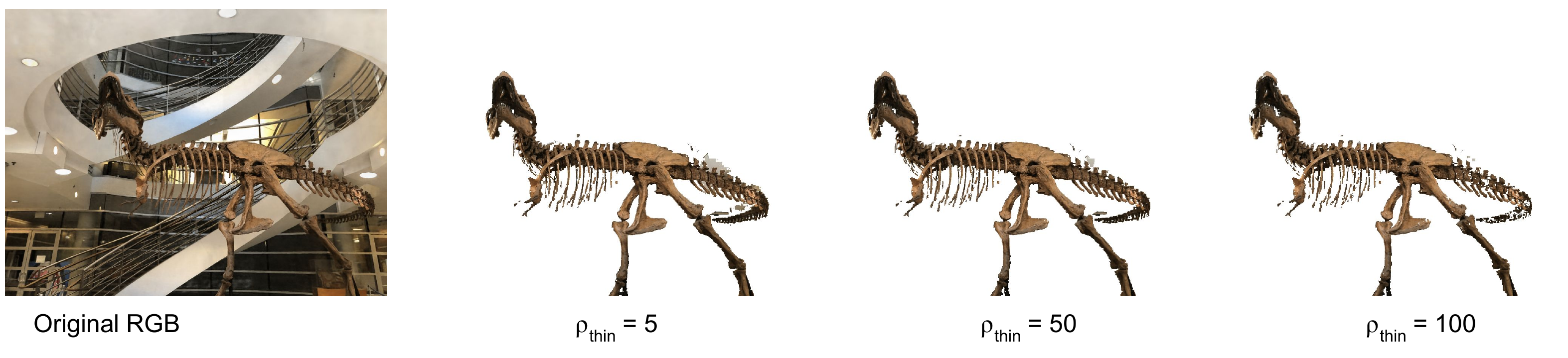}
    \caption{Qualitative ablation for Trex on the thin density threshold $\rho_\text{thin\,thresh}$ (NeRF instantiation).}
    \label{fig:thin_density_threshold}
\end{figure}

The thin-structure density threshold $\rho_\text{thin\_thresh}$ decides when the algorithm activates the thin-structure path. This path uses voxel-footprint projection and recovers details that the thick path often misses. Unlike the standard density threshold, which mainly removes background voxels, $\rho_\text{thin\_thresh}$ must be higher. It should fire only when NeRF predicts strong density, which signals likely object occupancy. Thin regions, such as the Trex rib bones, show this clearly. They have high NeRF density but very small spatial support. The thick path ignores them, but the thin path can recover them. As Figure~\ref{fig:thin_density_threshold} shows, low values (e.g., $\rho_\text{thin}{=}5$) cause background leakage, while very large values (e.g., $100$) start removing fine structures. The IoU results in Tables~\ref{tab:rho_thin_llff} and~\ref{tab:rho_thin_mip} show the same trend. For LLFF, we set $\rho_\text{thin\,thresh}{=}50$. This value preserves thin structures while avoiding extra projection work on background voxels. For Mip-NeRF~$360^\circ$, the finer $256^3$ grid gives sharper densities, so a slightly lower value works well there, and we use $\rho_\text{thin\_thresh}{=}30$.

\paragraph{2DGS instantiation.}
As argued in the 2DGS paragraph of Section~\ref{app:rho-thresh}, the thick and thin paths of the 2DGS kernel share the same density floor, $\rho_\text{thresh}{=}0.5$. We therefore do not maintain a separate $\rho_\text{thin\_thresh}$ parameter for the 2DGS instantiation. The eligibility check for the thin path inside Algorithm~\ref{alg:gs-thin} uses the same floor as the thick path.

% ----------------------------------------------------------------
\subsection{2DGS-specific kernel hyperparameters}
\label{app:gs-specific}

Two 2DGS hyperparameters that have no NeRF counterparts and are scene-specific are exposed as sliders in the UI.

\paragraph{See-through gap $\gamma_\text{ST}$ (slider \texttt{see\_through\_gap}).}
$\gamma_\text{ST}$ controls when the thick path is allowed to enter the thin fallback for a voxel whose center-aligned ray hits empty space between the foreground and the next opaque surface. The acceptance condition is given by Equation~\ref{eq:seethrough}, anchored to the camera-to-target distance rather than to $h_v$ so that the gate stays scale-invariant across the per-scene adaptive voxel size. Large values open the gate aggressively and admit far-side leakage, while very small values close the gate even on legitimate thin features at oblique angles.

\paragraph{Surfel coverage threshold (slider \texttt{surfel\_coverage}).}
The final mask threshold in Pass~3 of the voxel-to-surfel mapping (Algorithm~\ref{alg:vox2surfel}, Section~\ref{sec:method-vox2surfel}) decides which surfels are kept after the tile-sorted footprint coverage accumulation. Default $0.5$. Smaller values yield slightly broader 3D masks, at the expense of boundary surfels with low coverage, leading to over-segmentation. Larger values produce a sharper boundary at the cost of dropping low-coverage thin surfels, leading to holes in the mask.

% ----------------------------------------------------------------
% ================================================================
\section{Complexity and Runtime Analysis}
\label{app:complexity}

The end-to-end NeRF-instantiation runtime against SANeRF-HQ (a consistent $1.7$-$2.1\times$ speedup) is reported in the main paper, Section~\ref{sec:exp-quant} (Table~\ref{tab:simple_timing}). This section provides the finer-grained propagation-time analysis behind that result.

\paragraph{Propagation-time comparison with SA3D.}
We compare DivAS (NeRF instantiation) against the optimization-based baseline SA3D~\citep{cen2023segment}. The total propagation cost is reported in Table~\ref{tab:appendix-propagation-cmp-nerf}. The SA3D column is the cumulative cost of the mask inverse-rendering pass summed over all training iterations of the SA3D optimization, whereas the DivAS column $T_\text{pipe}^\text{tot}$ is the cumulative cost of forward render, depth-weighted SAM mask refinement, and CUDA fusion-kernel summed over all centroid views. The speedup column is the ratio of the cumulative cost of SA3D vs DivAS. We additionally report two finer-grained DivAS kernel timings, the cumulative fusion-kernel time $T_\text{kern}^\text{tot}$ and the per-call mean $\bar T_\text{kern}$, since these dominate the cost of any single user-driven update. The peak GPU memory consumption is reported only for DivAS. The DivAS peak VRAM is included only to characterize the absolute footprint of our method, not as a competitive claim against SA3D in NeRF instantiation.

As shown in Table~\ref{tab:appendix-propagation-cmp-nerf}, DivAS is $3.60\times$ to $33.10\times$ faster than SA3D at the per-method propagation level, with a mean speedup of $18.07\times$. The gap widens substantially with dataset size, where the speedup is $\sim 3.6$-$8.3\times$ on the bounded LLFF scenes but jumps to $\sim 27.3$-$33.1\times$ on the unbounded Mip-NeRF~$360^\circ$ scenes, which have many more training cameras and a denser scene representation. The reason is structural, SA3D performs a global mask inverse-rendering optimization whose per-iteration cost scales with the number of training cameras and with the trained NeRF field, so its propagation time grows roughly linearly with the training-set size $N_\text{train}$. DivAS instead processes centroid views which are dependent on the complexity of the object instead of the training set size. The DivAS propagation budget is therefore largely independent of $N_\text{train}$, so larger or denser captures, the regime that inflates SA3D's optimization cost the most, do not slow DivAS down by the same factor. On the absolute kernel cost, $\bar T_\text{kern}$ stays below $65$\, ms per call across all four scenes, and the cumulative $T_\text{kern}^\text{tot}$ stays below $240$\,ms even on Garden, the bulk of the DivAS propagation wall-clock is therefore centroid-view rendering and SAM mask refinement, not the fusion kernel itself. Peak GPU memory is bounded by $\sim 7.95$\,GB on the four scenes, well within the limits of consumer hardware.

\begin{table}
\centering
\caption{Propagation-time comparison between SA3D and DivAS (NeRF instantiation), Peak VRAM consumption by DivAS on four representative scenes. SA3D and $T_\text{pipe}^\text{tot}$ are in seconds, kernel timings are in milliseconds, and lower is better for the time columns. The Mean row contains arithmetic means across the four scenes for all columns except Peak VRAM, which reports the maximum per-scene peak.}
\label{tab:appendix-propagation-cmp-nerf}
\small
\begin{tabular}{lrrrrrr}
\toprule
Scene & SA3D (s) $\downarrow$ & $T_\text{pipe}^\text{tot}$ (s) $\downarrow$ & $T_\text{kern}^\text{tot}$ (ms) $\downarrow$ & $\bar T_\text{kern}$ (ms) $\downarrow$ & Peak VRAM (GB) & Speedup $\uparrow$ \\
\midrule
\multicolumn{7}{l}{\textit{LLFF Dataset}} \\
Fern              &  19.934 & \textbf{5.535}  &  79   & 26.0 & 7.81 &  3.60$\times$ \\
Trex              &  45.065 & \textbf{5.457}  & 185   & 62.0 & 6.86 &  8.26$\times$ \\
\multicolumn{7}{l}{\textit{Mip-NeRF~$360^\circ$ Dataset}} \\
Bonsai            & 172.314 & \textbf{5.207}  & 142   & 28.3 & 7.94 & 33.10$\times$ \\
Garden (no vase)  & 282.419 & \textbf{10.349} & 237   & 47.4 & 7.95 & 27.30$\times$ \\
\midrule
\textbf{Mean}     & 129.933 & \textbf{6.637} & \textbf{160.8} & 40.9 & 7.95 & \textbf{18.07$\times$} \\
\bottomrule
\end{tabular}
\end{table}

\paragraph{Complexity Analysis (NeRF instantiation).}
The fusion kernel scales linearly with the number of active voxels. Each voxel performs a fixed number of projections and depth checks across $|\mathbf{M}|$ depth-refined SAM masks, giving $\text{O}(|\tM|)$ cost for the thick-structure path. For thin-structure voxels (approximately $40\%$ of active voxels), we additionally scan a small $2$D footprint of $\text{O}(F)$ pixels ($F \approx 200$-$300$). Thus, the per-voxel cost becomes $\text{O}(|\tM|)$ or $\text{O}(|\tM|\!\times\!F)$ depending on the local structure type. Since voxel updates are fully parallelized on the GPU and most of the $3$D volume is removed by NeRF density pruning, the effective runtime scales as $\mathcal{O}(V_\text{eff})$, where $V_\text{eff}$ denotes density-valid voxels.

\paragraph{2DGS instantiation.}
The 2DGS fusion kernel preserves the linear-in-active-voxels scaling but operates on a Morton-scattered single-cascade grid constructed once from the trained surfel opacities (Section~\ref{sec:method-gs-fusion}). The thick path is $\text{O}(|\tM|)$ per voxel as in the NeRF case. The voxel footprint in thin path is determined entirely by the projected voxel extent. Footprint statistics over $\mN_\text{samp}$ are still amortized through integral images so that the local mean and variance of $\mN_\text{samp}$ inside this bounding box are read in $\text{O}(1)$, but the per-pixel inner loop itself walks every pixel in $F$. The per-voxel thin-path cost is therefore $\text{O}(|\tM|\!\times\!F)$ with the same shape as the NeRF analysis above, only the $F$ regime differs. Empirically (per-call logs on \textit{Garden}, the largest Mip-NeRF~$360^\circ$ scene we evaluate), $F$ ranges from $\sim 1.3\text{k}$ to $\sim 6.7\text{k}$ pixels with a mean of $\sim 2.8\text{k}$, well above the $F\!\approx\!200\text{-}300$ regime of the NeRF kernel because the 2DGS voxel size $h_v$ scales with the median surfel footprint rather than with a fixed grid resolution and the voxel grid is single cascade compared to multi-cascade in NeRF. The voxel-to-surfel mapping in Algorithm~\ref{alg:vox2surfel} adds three monotone passes whose dominant cost is the tile-sorted preprocess$\rightarrow$sort$\rightarrow$gather pipeline of Pass$3$, again linear in the total number of surfels touched.

\paragraph{Metrics.}
We report five efficiency quantities in Table~\ref{tab:appendix-divas-gs-efficiency}. All wall-clock columns labelled ``Train'' and ``Interact.''\ are reported in \textbf{mm:ss} (minutes:seconds), and all GPU-time columns ($T_{\text{pipe}}^{\text{tot}}$, $\bar T_{\text{pipe}}$, $\bar T_{\text{kern}}$) are reported in \textbf{ms} (milliseconds). The pipeline and kernel timings are GPU-side, the time the GPU actually spends inside the rendering, depth-weighting, fusion-kernel, and voxel-to-surfel stages, and exclude user-driven idle periods. (i) The \textbf{scene-construction time} ``Train'' is the one-shot 2DGS~\citep{huang20242d} scene training time. (ii) The \textbf{interactive wall time} ``Interact.'' is the wall-clock time of the DivAS-GS method. (iii) The cumulative \textbf{pipeline time} $T_{\text{pipe}}^{\text{tot}}$ is the sum of render, depth-weighting, fusion-kernel, and voxel-to-surfel mapping GPU costs over all \emph{centroid} views. This time is from the first user click to the last accepted view, and timing starts when the user requests centroid view generation, at which point all anchor views have already been processed. (iv) The per-call \textbf{kernel time} $\bar T_{\text{kern}}=T_{\text{kern}}^{\text{tot}}/K_{\text{seg}}$ is the per-call mean of fusion kernel plus voxel-to-surfel mapping, with $K_{\text{seg}}$ is count of those centroid views for which a SAM mask was actually fused (views the user accepted without adding a prompt are excluded from this average since the kernel does not run on them). (v) The \textbf{per-view pipeline mean} $\bar T_{\text{pipe}}$ uses $K_{\text{total}}$, the total number of centroid views shown, in the denominator, so the number is directly comparable to the per-view mask-inverse-rendering cost of SA3D-style baselines.

\begin{table}
\centering
\caption{DivAS-GS GPU timing analysis of different components on a representative subset of four scenes (two LLFF, two Mip-NeRF~$360^\circ$).}
\label{tab:appendix-divas-gs-efficiency}
\begin{tabular}{lrrrrr}
\toprule
 & Train & Interact. & $T_{\text{pipe}}^{\text{tot}}$ & $\bar T_{\text{pipe}}$ & $\bar T_{\text{kern}}$ \\
Scene & (mm:ss) & (mm:ss) & (ms) & (ms) & (ms) \\
\midrule
\multicolumn{6}{l}{\textit{LLFF Dataset}} \\
Fern              & 18:52 & 04:27 &  367 & 41 & 21 \\
Trex              & 13:15 & 06:26 &  304 & 38 & 27 \\
\multicolumn{6}{l}{\textit{Mip-NeRF~$360^\circ$ Dataset}} \\
Bonsai            & 12:33 & 05:02 &  528 & 31 & 28 \\
Garden (no vase)  & 27:31 & 09:24 & 1381 & 73 & 67 \\
\bottomrule
\end{tabular}
\end{table}

\paragraph{Compute-time observation.}
Across the four representative scenes in Table~\ref{tab:appendix-divas-gs-efficiency}, the cumulative GPU pipeline cost $T_{\text{pipe}}^{\text{tot}}$ did not cross $1.5$\,s for an entire interactive segmentation run, and the per-call kernel time $\bar T_{\text{kern}}$ stays under $70$~ms on every scene. The bulk of the ``Interact.''\ wall-clock budget is therefore not GPU compute but user interaction time, the user clicking on object regions, inspecting the back-projected 3D mask, and accepting or refining the next centroid view. This is in sharp contrast to optimization-based baselines, whose total runtime is dominated by an upfront, non-interactive offline optimization pass that the user does not have control. By collapsing the per-update cost to a sub-second pipeline call, DivAS-GS keeps the user in the loop and provides immediate, fine-grained feedback at every view rather than after a multi-minute compute stage.

\paragraph{Propagation-time comparison with SA3D-GS.}
We compare DivAS-GS against the optimization-based baseline SA3D-GS~\citep{cen2023segment}. The total propagation cost is reported in Table~\ref{tab:appendix-propagation-cmp}. The SA3D-GS column is the cumulative cost of forward render, loss evaluation through the gaussian primitives, and backward summed over all training iterations of the SA3D-GS optimization, whereas the DivAS-GS column is the cumulative cost of forward render, depth-weight mask refinement, fusion kernel, and voxel-to-surfel mapping summed over all centroid views, and the speedup column is the ratio SA3D-GS / DivAS-GS. As shown in Table~\ref{tab:appendix-propagation-cmp}, DivAS-GS is $2.62\times$ to $13.66\times$ faster than SA3D-GS at the per-method propagation level, with a mean speedup of $7.49\times$. The gap widens dramatically with scene dataset size, the speedup is $\sim 2.6$-$4.8\times$ on the bounded LLFF scenes but jumps to $\sim 8.9$-$13.7\times$ on the unbounded Mip-NeRF~$360^\circ$ scenes. The reason is structural, SA3D-GS performs a global optimization whose per-iteration cost scales with the number of training cameras and the number of trainable Gaussians, so the propagation time grows linearly with the dataset size $N_\text{train}$. DivAS-GS instead processes ~$3-4$ centroid views per anchor, and the per-call cost is dominated by the active-voxel count, which depends only on the foreground complexity of the scene. This makes the DivAS-GS propagation budget independent of $N_\text{train}$, so larger or denser captures do not slow it down compared to the SA3D-GS optimization-based method.

\begin{table}
\centering
\caption{Total propagation time (ms) summed over the full interactive run on the four representative scenes. Lower is better.}
\label{tab:appendix-propagation-cmp}
\begin{tabular}{lrrr}
\toprule
Scene & SA3D-GS (ms) $\downarrow$ & DivAS-GS (ms) $\downarrow$ & Speedup \\
\midrule
\multicolumn{4}{l}{\textit{LLFF Dataset}} \\
Fern             &   959.989 & \textbf{366.515}  &  2.62$\times$ \\
Trex             &  1450.179 & \textbf{303.772}  &  4.77$\times$ \\
\multicolumn{4}{l}{\textit{Mip-NeRF~$360^\circ$ Dataset}} \\
Bonsai           &  7213.945 & \textbf{528.080}  & 13.66$\times$ \\
Garden (no vase) & 12307.233 & \textbf{1381.418} &  8.91$\times$ \\
\midrule
\textbf{Mean}    & 5482.836 & \textbf{644.946} & \textbf{7.49$\times$} \\
\bottomrule
\end{tabular}
\end{table}

% ================================================================
\section{Limitations and Future Work}
\label{app:limitations}

DivAS operates on top of an already-reconstructed NeRF or 2DGS scene and therefore inherits the geometric envelope of that reconstruction. Where the underlying backbone yields sparse, noisy, or otherwise low-confidence geometry, for example, under wide-baseline captures or strongly under-sampled regions, the depth-weighted voting can mis-aggregate evidence, and the reconstruction error propagates into the segmentation mask. A second, related case is foreground-background depth ambiguity, when the target object lies very close to a textured background surface in the centroid view, the depth-derived weighting becomes less discriminative, which could lead to background bleeding. The automatic Fibonacci anchor scheduler is geometry-aware but not illumination-aware, so if a proposed anchor observes the object under poor lighting, hard shadows, or weak texture contrast, SAM may return false negatives that propagate into the fused 3D mask, in our experience this is usually mitigated by picking an alternative anchor through the set of fibonacci views or nudging the centroid-view camera, though it remains an automation gap rather than a hard failure. The centroid zoomed views reduce false negatives and mitigate false positives of the SAM model, but not in all cases, so overall, the segmentation quality of the model is upper-bounded by the SAM model. Transparent and highly reflective materials i.e., non-Lambertian surfaces remain challenging in this setting which is limitation of both NeRF and 2DGS representations which often reconstruct such regions with ambiguous geometry and not able to reconstruct well (e.g., glass in the middle of the table in the Garden scene), and DivAS may then segment only the parts of the surface for which the underlying geometry is locally well-defined, even when the corresponding 2D SAM masks look reasonable. Non-inward-facing trajectories were not evaluated in this work and remain open.

Several directions are open for future work. Extending the automatic view scheduler beyond the inward-facing viewing-sphere assumption would broaden the automated regime to irregular capture trajectories. Making the anchor scheduler illumination-aware so that the GUI proposes lit, well-textured views by default would reduce the residual SAM false-negative rate without additional user effort. Finally, extending the framework to dynamic scenes and more unconstrained capture protocols is a natural next step given that the shared interaction-and-fusion recipe already transfers across representations through representation-specific adapters.
% \section{Appendix}

\end{document}